%% file: arxiv_new_version.tex
\documentclass[11pt]{article}

\usepackage[preprint]{acl}

\usepackage{times}
\usepackage{latexsym}
\usepackage{booktabs}
\usepackage{array} 
\usepackage{ragged2e}
\usepackage{arydshln}
\usepackage{multirow}
\usepackage{placeins}
\usepackage{xspace}

\usepackage{tcolorbox}

\usepackage{rotating}

\usepackage[T1]{fontenc}

\usepackage[utf8]{inputenc}
\usepackage{amssymb}

\usepackage{microtype}

\usepackage{inconsolata}
\usepackage{pifont}
\usepackage{textcomp}

\usepackage{graphicx}

\title{When Visuals Aren't the Problem: Evaluating Vision-Language Models on Misleading Data Visualizations}

\newcommand{\gtlogo}{\raisebox{3.4pt}{\includegraphics[scale=0.04]{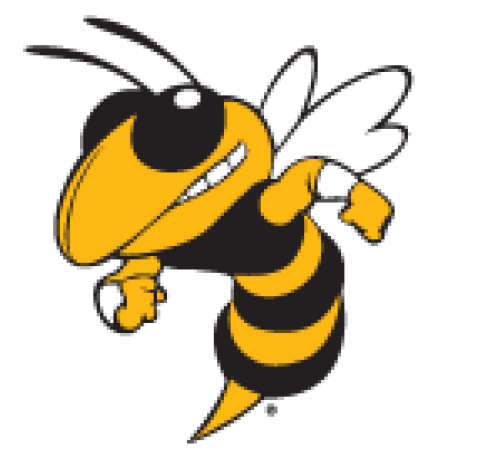}}}
\newcommand{\hrvlogo}{\raisebox{3.4pt}{\includegraphics[scale=0.05]{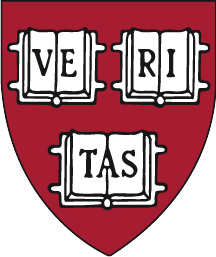}}}
\newcommand{\bitslogo}{\raisebox{3.4pt}{\includegraphics[scale=0.02]{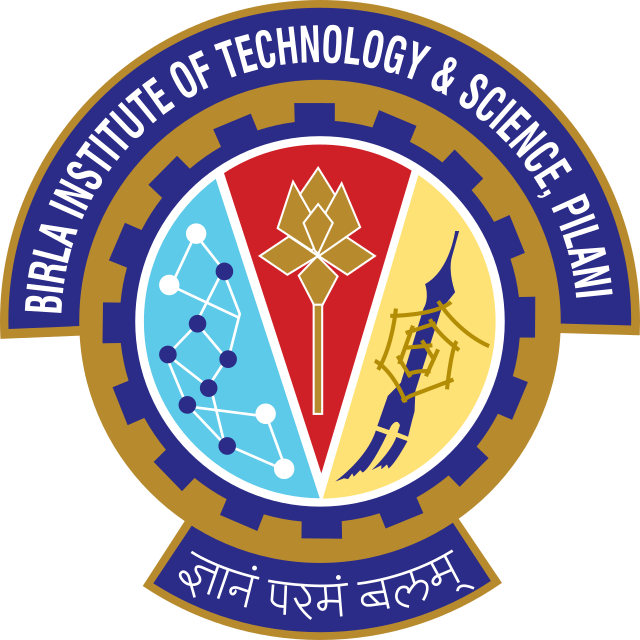}}}

\definecolor{myColour}{HTML}{003366}

\makeatletter
\def\thanks#1{%
  \protected@xdef\@thanks{%
    \@thanks
    \protect\footnotetext{\noindent#1}%
  }%
}
\makeatother
\author{
\bf \hypersetup{linkcolor=black} Harsh Nishant Lalai \bitslogo \footnotemark[1] \hspace{0.4in}
Raj Sanjay Shah \gtlogo \footnotemark[1] \hspace{0.4in}
Hanspeter Pfister \hrvlogo \\
\bf \hypersetup{linkcolor=black} Sashank Varma \gtlogo \hspace{0.4in}
Grace Guo \hrvlogo \\
Birla Institute of Technology and Science, Pilani \bitslogo \\
 Georgia Institute of Technology \gtlogo  \hspace{0.4in} Harvard University \hrvlogo\\
\thanks{* Equal contribution. Emails: lalaiharsh26@gmail.com, rajsanjayshah@gatech.edu, gguo31@g.harvard.edu. Code and dataset available at 
\href{https://github.com/Harsh-Lalai/Evaluating-Vision-Language-Models-on-Misleading-Data-Visualizations}{\textit{GitHub}} 
and 
\href{https://huggingface.co/datasets/MaybeMessi/MisVisBench}{\textit{HuggingFace}} respectively.}
}

\newcommand{\gpt}{\raisebox{0pt}{\includegraphics[scale=0.05,trim={0 20 0 12}]{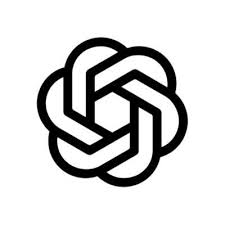}}}
\newcommand{\gemini}{\raisebox{0pt}{\includegraphics[scale=0.017]{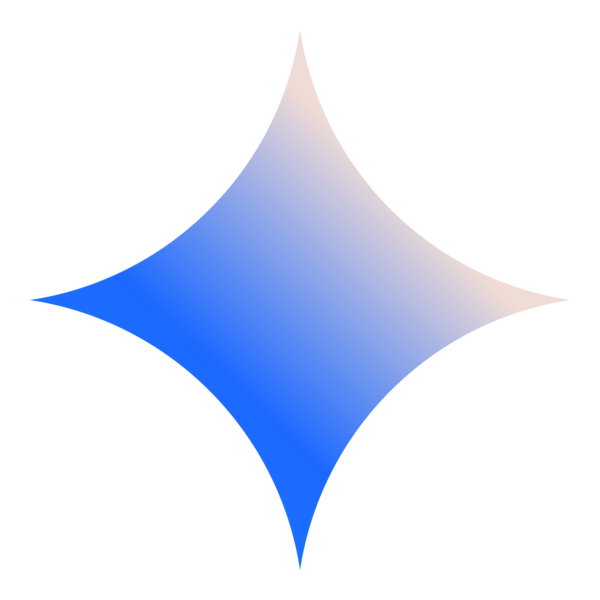}}}
\newcommand{\qwen}{\raisebox{0pt}{\includegraphics[scale=0.005,trim={0 20 0 12}]{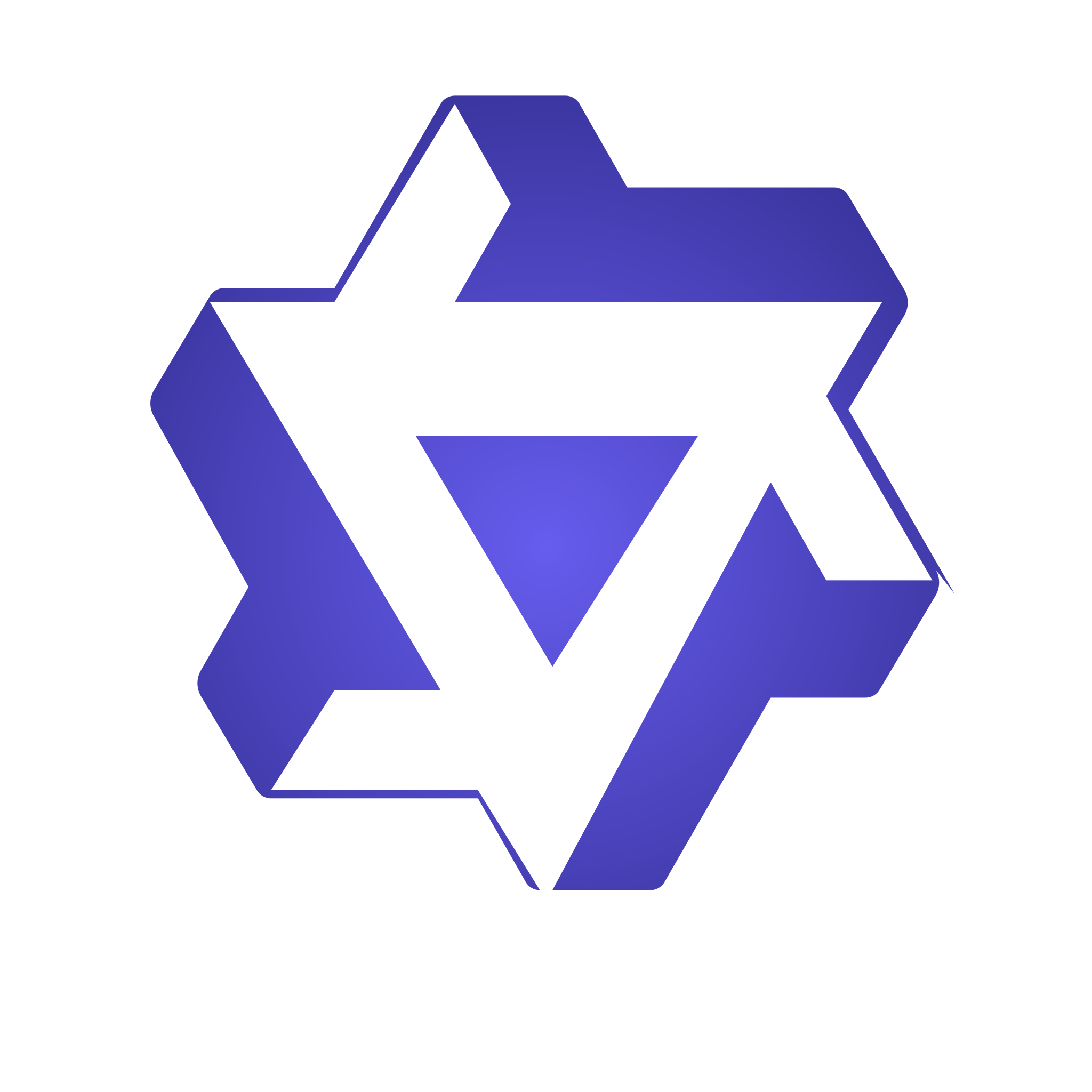}}}

\newcommand{\geminitwofivef}{\gemini \textsubscript{2.5-F}\xspace}
\newcommand{\geminitwofivep}{\gemini \textsubscript{2.5-P}\xspace}
\newcommand{\geminithreezerop}{\gemini \textsubscript{3.0-P}\xspace}
\newcommand{\geminithreeonep}{\gemini \textsubscript{3.1-P}\xspace}

\newcommand{\gptfive}{\gpt \textsubscript{5}\xspace}
\newcommand{\gptfivemini}{\gpt \textsubscript{5-Mini}\xspace}

\newcommand{\qwentwofive}{\qwen \textsubscript{2.5}\xspace}
\newcommand{\qwentwofivechartqa}{\qwen \textsubscript{2.5-ChartQA}\xspace}
\newcommand{\qwenthree}{\qwen \textsubscript{3}\xspace}

\newcommand{\cmark}{\textcolor{green!80!black}{\ding{51}}}
\newcommand{\xmark}{\textcolor{red}{\ding{55}}}
\newcommand{\bcircle}{\textcolor{blue}{\ding{109}}}

\begin{document}
\maketitle
\begin{abstract}

Visualizations help communicate data insights, but deceptive data representations can distort their interpretation and propagate misinformation. While recent Vision Language Models (VLMs) perform well on many chart understanding tasks, their ability to detect misleading visualizations, especially when deception arises from subtle reasoning errors in captions, remains poorly understood. 
Here, we evaluate VLMs on misleading visualization-caption pairs grounded in a fine-grained taxonomy of reasoning errors (e.g., Cherry-picking, Causal inference) and visualization design errors (e.g., Truncated axis, Dual axis, inappropriate encodings).
To this end, we develop a benchmark that combines real-world visualization with human-authored, curated misleading captions designed to elicit specific reasoning and visualization error types, enabling controlled analysis across error categories and modalities of misleadingness.
Evaluating many commercial and open-source VLMs, we find that models detect visual design errors substantially more reliably than reasoning-based misinformation, and frequently misclassify non-misleading visualizations as deceptive.
Overall, our work fills a gap between coarse detection of misleading content and the attribution of the specific reasoning or visualization errors that give rise to it.
\vspace{-5pt}
\end{abstract}

\section{Introduction}
\vspace{-5pt}

\begin{figure*}[htp]
\centering
\includegraphics[trim={0cm 12pt 0cm 12pt}, width=\textwidth]{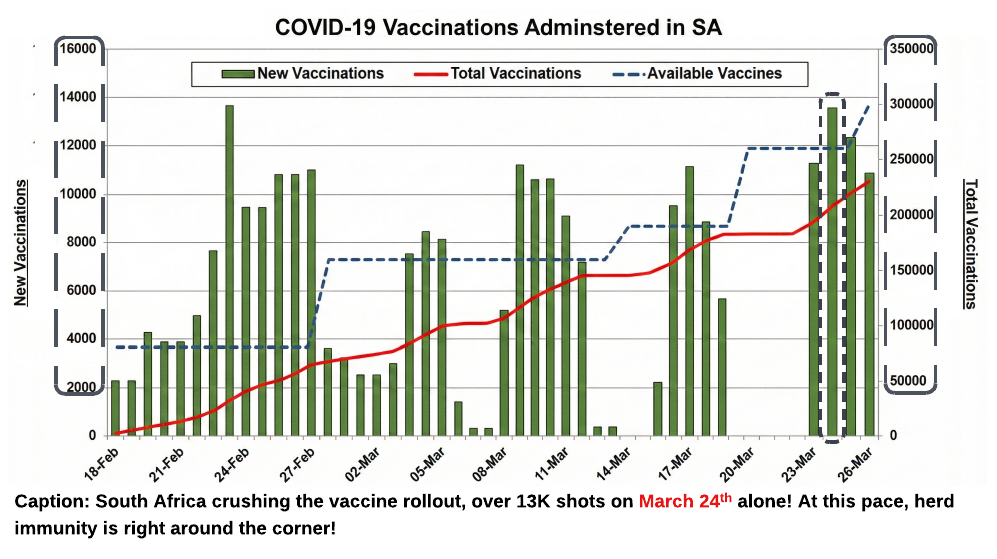}
\caption{Example of a misleading chart-caption pair with both visual design and reasoning errors. The chart contains a \textbf{dual-axis} visualization design error, which may be confusing because viewers must mentally map each axis to its corresponding visual representation (bar or line). The caption also introduces a reasoning error by extrapolating a \textbf{cherry-picked} short-term increase to a broader \textbf{causal claim}. Together, these factors can distort interpretation without altering the underlying data.}
\label{fig:fig1}
\vspace{-15pt}
\end{figure*}

Visualizations are often used to communicate data-driven insights and to convey complex information effectively. When paired with well-crafted captions, visualizations can improve understanding and decision-making across domains ranging from journalism \cite{weber2012data, fu2023more} to scientific research \cite{mogull2015current, duarte2022role}. However, the same communicative power that makes visualizations impactful makes them susceptible to misrepresentations. Misleading captions and deceptive data representations can distort interpretation, propagate misinformation, and break public trust in data communication \cite{pandey2015deceptive,parks2021lie,akhtar2024chartcheck}.

\noindent\textbf{Defining misleading visualizations:}
We adopt the definition provided by \citet{richards2013deceptive} and \citet{pandey2015deceptive} as \emph{``a graphical depiction of information, designed with or without an intent to deceive, that may create a belief about the message and/or its components, which varies from the actual message.''} Importantly, under this definition, a mislead does not require malicious intent; even well-intentioned visualizations can mislead through ambiguous framing, selective emphasis, or incorrect interpretation.

Prior work on visualization misinformation has largely focused on flawed visual design, from early notions of graphical integrity \cite{tufte1983visual} to subsequent taxonomies and descriptions of design errors \cite{ pandey2015deceptive, correll2017black, lo2022misinformed}.
However, recent studies have shown that misleading real-world visualizations often arise not only from flawed visual encodings (e.g., truncated axes or dual axes), but also from subtle \emph{reasoning errors} in how captions describe or infer meaning from the data \cite{lisnic2023misleading, lan2024came}.
Such errors can appear even when the visualization itself is plausible or professionally produced. 
Figure~\ref{fig:fig1} shows a real-world example of a chart with both visualization design and reasoning errors.
The design includes a misleading dual-axis, while the caption cherry-picks a short-term spike in vaccinations to convey a distorted yet seemingly credible message.

Recent Vision-Language Models show strong performance on many chart understanding and multimodal reasoning tasks \cite{masry2022chartqa,islam2024large}. This progress raises a natural question: \emph{can VLMs detect misleading visualizations and accurately attribute them to specific documented reasoning and visualization design error types?} While existing benchmarks primarily focus on fact verification or chart-based Q\&A, they provide limited insight into how models handle reasoning-based misinformation embedded in visualization-caption pairs. In contrast, we study whether models can attribute misleadingness to specific error types and disentangle whether it arises from the caption, the visualization, or both.

For this, we introduce a benchmark comprising real-world visualizations with human-authored and curated misleading captions designed to elicit specific error types, enabling controlled analysis across error types and modalities (caption, visualization, or both). We assess a range of frontier commercial and open models and provide a diagnostic analysis of where current systems succeed and fail across error types, including their tendency to over-flag non-misleading examples. Lastly, we discuss learnings towards the real-world deployment of such systems.

\input{related_works_raj}
\input{Problem_new_version}

\section{Findings}
\label{section:findings}
\vspace{-5pt}

\subsection*{Finding 1: VLMs struggle to reliably detect and classify misinformation errors in real-world visualization-caption pairs.}

\begin{table}[t]
\centering
\resizebox{\columnwidth}{!}{
\begin{tabular}{lccccccccc}
\toprule
\textbf{Metric} 
& \geminithreeonep
& \geminithreezerop
& \geminitwofivep 
& \geminitwofivef 
& \gptfive 
& \gptfivemini 
& \qwenthree 
& \qwentwofive 
& \qwentwofivechartqa \\
\midrule
\textbf{F1} & 0.56 & \textbf{0.57} &  0.59  & 0.52 & 0.56 & 0.53 & 0.47 & 0.27 & 0.24 \\
\textbf{PM} & 0.85 & \textbf{0.89} &  0.84  & 0.75 & 0.86 & 0.82 & 0.78 & 0.81 & 0.77 \\
\textbf{EM} & 0.21 & \textbf{0.23} &  0.06  & 0.02 & 0.12 & 0.06 & 0.03 & 0.16 & 0.16 \\
\bottomrule
\end{tabular}
}
\caption{The performance of various VLMs on our dataset. We report the combined weighted F1, Partial Match, and Exact Match scores. }
\label{tab:overall_combined_whole_benchmark_transposed}
\vspace{-14pt}
\end{table}

Across all evaluated models, misinformation error detection remains challenging, with no system achieving strong performance on the full benchmark.
Even the highest-performing VLMs achieve only mid-range weighted F1 scores (Best model: \geminithreezerop, 0.57), indicating limited ability to identify and attribute misinformation in real-world visualization-caption pairs (Table \ref{tab:overall_combined_whole_benchmark_transposed}). Interestingly, \geminithreeonep performs comparably to \geminithreezerop, suggesting that this benchmark remains challenging even for the newer model releases.
Additionally, this limitation persists even for models explicitly fine-tuned for chart understanding: \qwentwofivechartqa performs comparably to general-purpose versions of the Qwen family and substantially below other frontier systems. This suggests that relatively strong performance on existing chart-centric benchmarks primarily serves other goals, for example, value extraction and factual question answering, and that these capabilities may not \emph{consistently transfer to reasoning about misleading framing, selective interpretation, or multimodal misinformation.} Overall, the uniformly low F1 scores indicate that detecting and attributing the full spectrum of reasoning and visualization errors remains a difficult task for current state-of-the-art VLMs.

This difficulty is also echoed in the distinct gap between Partial Match and Exact Match scores across all models. While PM scores are relatively high, showing that models frequently identify some form of misleadingness, EM scores remain extremely low, showing that models \emph{rarely recover the complete and correct set of errors present in a given example.} This disparity indicates that current VLMs may operate at the level of coarse detection rather than precise attribution: they may often sense that a visualization-caption pair is problematic, but fail to fully enumerate or correctly describe the underlying reasoning and visualization errors. As a result, partial detection can overstate models' actual ability to perform fine-grained multimodal error attribution, as partial overlap may mask missing, spurious, or mislocalized error predictions.

\subsection*{Finding 2: VLMs are systematically better at detecting visual deception than reasoning-based deception, even when the latter is isolated.}

\begin{table}[t]
\centering
\resizebox{\columnwidth}{!}{
\begin{tabular}{lccc:ccc}
\toprule
 & \multicolumn{3}{c:}{\textbf{Reasoning Errors}} 
 & \multicolumn{3}{c}{\textbf{Visualization Errors}} \\
\cmidrule(lr){2-4}
\cmidrule(lr){5-7}
\textbf{Model} & F1 & PM & EM
               & F1 & PM & EM \\
\midrule
\geminithreeonep & 0.52 & 0.64 & 0.44 & 0.60 & 0.60 & 0.46 \\ 
\geminithreezerop & 0.52 & 0.66 & 0.45 & 0.63 & 0.69 & 0.51 \\
\geminitwofivep         & 0.56 & 0.58 & 0.25 & 0.62 & 0.58 & 0.24 \\
\geminitwofivef          & 0.46 & 0.40 & 0.08 & 0.58 & 0.56 & 0.27 \\
\gptfive           & 0.53 & 0.63 & 0.31 & 0.59 & 0.65 & 0.39 \\
\gptfivemini       & 0.47 & 0.48 & 0.15 & 0.59 & 0.67 & 0.38 \\
\qwenthree         & 0.46 & 0.59 & 0.31 & 0.47 & 0.42 & 0.08 \\
\qwentwofive        & 0.26 & 0.61 & 0.47 & 0.29 & 0.50 & 0.38 \\
\qwentwofivechartqa & 0.22 & 0.56 & 0.42 & 0.26 & 0.47 & 0.38 \\
\bottomrule
\end{tabular}
}
\caption{Performance of the VLMs on reasoning and visualization error classification on the whole dataset. We report each score separately for reasoning and visualization errors. Models consistently achieve higher scores on visualization error detection than reasoning errors, suggesting greater difficulty in identifying and reasoning about misinformation embedded in captions. }
\vspace{-13pt}
\label{tab:overall_reasoning_visual_whole_benchmark}
\end{table}

All the models achieve lower weighted F1 scores on reasoning-error classification than on visualization-error classification on the whole benchmark (Table \ref{tab:overall_reasoning_visual_whole_benchmark}), \textit{highlighting the greater difficulty of reasoning-based misleads}. Importantly, no model reverses this trend, indicating a systematic asymmetry rather than model-specific variation. For instance, closed-source models (\gpt and \gemini series) exhibit a 6 to 12 point gap in F1 scores between the two tasks, and this gap becomes smaller for the open-sourced models (\qwenthree, \qwentwofive, \qwentwofivechartqa); however, even these fail to detect caption-based reasoning errors better than visualization errors.

\begin{table*}[!t]
\centering
\resizebox{\linewidth}{!}{
\begin{tabular}{lccccccc}
\toprule
\multicolumn{8}{c}{\textbf{Reasoning Errors}} \\
\midrule
\textbf{Model}
& \textbf{Cherry}
& \textbf{Causal}
& \textbf{Setting an Arbi-}
& \textbf{Statistical}
& \textbf{Incorrect Reading}
& \textbf{Issues with}
& \textbf{Misrepresentat-} \\
& \textbf{Picking}
& \textbf{Inference}
& \textbf{trary Threshold}
& \textbf{Nuance}
& \textbf{of the Chart}
& \textbf{Data Validity}
& \textbf{ion of Studies} \\

\midrule

\geminithreeonep & 0.43 & 0.74 & 0.55 & 0.13 & 0.06 & 0.28 & 0.22 \\
\geminithreezerop & 0.48 & 0.71 & 0.51 & 0.11 & 0.04 & 0.26 & 0.24 \\
\geminitwofivep & 0.57 & 0.67 & 0.62 & 0.12 & 0.10 & 0.04 & 0.15 \\
\geminitwofivef & 0.43 & 0.59 & 0.51 & 0.10 & 0.03 & 0.08 & 0.28 \\
\gptfive & 0.53 & 0.68 & 0.56 & 0.13 & 0.04 & 0.13 & 0.22 \\
\gptfivemini & 0.49 & 0.59 & 0.50 & 0.10 & 0.03 & 0.08 & 0.26 \\
\qwenthree & 0.45 & 0.60 & 0.49 & 0.18 & 0.02 & 0.05 & 0.15 \\
\qwentwofive & 0.46 & 0.20 & 0.19 & 0.12 & 0.02 & 0.07 & 0.04 \\
\qwentwofivechartqa & 0.41 & 0.17 & 0.16 & 0.03 & 0.03 & 0.00 & 0.04 \\

\midrule
\multicolumn{8}{c}{\textbf{Visualization Errors}} \\
\midrule

\textbf{Model}
& \textbf{Truncated}
& \textbf{Dual}
& \textbf{Values as}
& \textbf{Inverted}
& \textbf{Uneven}
& \textbf{Unclear}
& \textbf{Inappropriate} \\
& \textbf{Axis}
& \textbf{Axis}
& \textbf{Area/ Volume}
& \textbf{Axis}
& \textbf{Binning}
& \textbf{Encoding}
& \textbf{Encoding} \\
\midrule

\geminithreeonep & 0.67 & 0.89 & 0.65 & 0.47 & 0.08 & 0.38 & 0.24 \\
\geminithreezerop & 0.66 & 0.89 & 0.71 & 0.49 & 0.12 & 0.40 & 0.24 \\
\geminitwofivep & 0.66 & 0.92 & 0.66 & 0.44 & 0.13 & 0.33 & 0.16 \\
\geminitwofivef & 0.58 & 0.86 & 0.68 & 0.44 & 0.08 & 0.34 & 0.16 \\
\gptfive & 0.54 & 0.89 & 0.73 & 0.15 & 0.11 & 0.36 & 0.20 \\
\gptfivemini & 0.60 & 0.89 & 0.66 & 0.36 & 0.16 & 0.37 & 0.18 \\
\qwenthree & 0.17 & 0.81 & 0.59 & 0.11 & 0.12 & 0.28 & 0.12 \\
\qwentwofive & 0.16 & 0.75 & 0.14 & 0.09 & 0.00 & 0.14 & 0.07 \\
\qwentwofivechartqa & 0.12 & 0.68 & 0.16 & 0.04 & 0.00 & 0.12 & 0.07 \\

\bottomrule
\end{tabular}
}

\caption{Per-error F1 scores for reasoning and visualization error classification on the full dataset. Models perform relatively well on some visually distinctive errors (e.g., Dual Axis and Values as Area) and some linguistic reasoning errors (e.g., Causal Inference), but struggle on errors requiring statistical interpretation or careful chart reading.}
\label{tab:per_error_f1_whole_benchmark}
\vspace{-13pt}
\end{table*}

\begin{figure}[ht]
    \centering
        \centering
        \includegraphics[trim={15pt 15pt 15pt 15pt}, width=\linewidth]{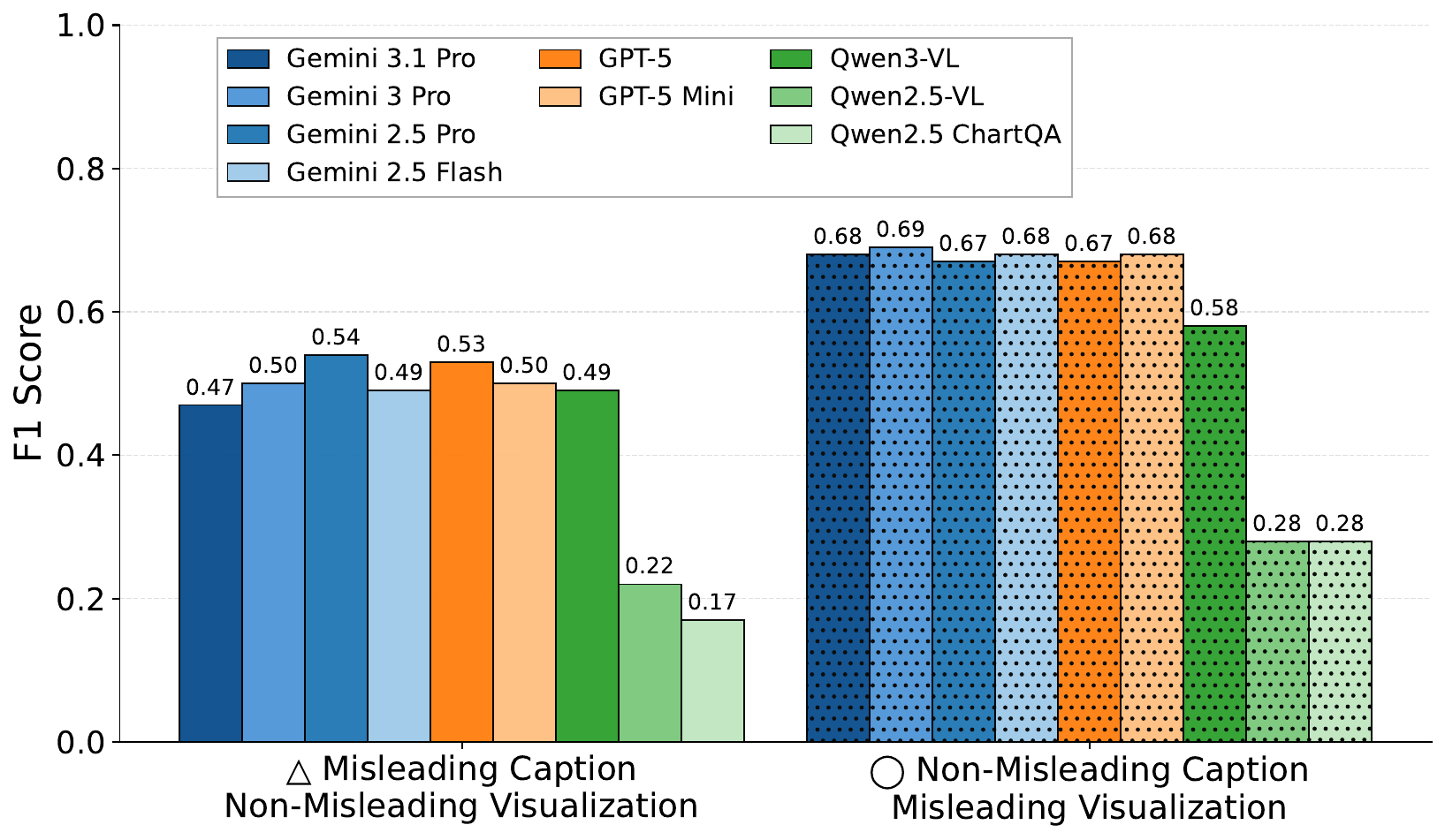}
        
            \caption{Combined weighted F1 scores for VLMs on benchmark subsets containing only one modality of misinformation.}
        \label{fig:reasoning_vs_viz_bar_chart}
        \vspace{-13pt}
\end{figure}

This asymmetry persists under controlled conditions that isolate a single source of misleadingness. When evaluated on samples containing misleading visualizations with non-misleading captions $\bigcirc$, models achieve substantially higher performance than on samples containing misleading captions paired with otherwise standard visualizations $\triangle$ (Figure \ref{fig:reasoning_vs_viz_bar_chart}). Because these subsets remove cross-modal confounds, the resulting performance gap provides direct evidence that reasoning-based deception, independent of visual distortions, poses a greater challenge for current VLMs. 
\emph{Notably, this pattern contrasts with a broad body of prior work showing that models often perform better on text-based reasoning than on image-based understanding \cite{sim2025can, park2025generalizing}. Our results suggest that this advantage does not straightforwardly extend to settings in which textual claims must be evaluated against visual evidence rather than in isolation.}

\subsection*{Finding 3: VLMs succeed on surface-level error patterns but struggle with epistemic and context-dependent reasoning errors.}

F1, PM, and EM scores vary substantially across error categories (see Table \ref{tab:per_error_f1_whole_benchmark}). 
Errors with salient visual structures or recurring linguistic templates are detected with relatively higher accuracy. In contrast, errors that require epistemic judgment, statistical reasoning, or careful alignment between captions and underlying data remain challenging.

Among reasoning errors, models perform best on categories such as causal inference (F1: 0.59 - 0.74), the use of arbitrary thresholds (F1: 0.50 - 0.56), and cherry-picking (F1: 0.43 - 0.53). These errors often involve \emph{recognizable textual cues}, such as explicit cause-and-effect language, selectively framed time windows, or highlighted cutoffs, that can be easily identified without engagement with the underlying data-generating process. 
Similarly, several visualization errors with \emph{visually distinctive patterns}, such as dual axes, truncated axes, and value-as-area encodings, achieve comparatively higher detection performance. These error types share consistent perceptual or structural signatures that appear amenable to pattern-based recognition.

In contrast, VLMs struggle markedly with errors that require contextual or epistemic reasoning. Reasoning categories such as incorrect reading of the chart (F1: 0.02 - 0.10), failure to account for statistical nuance (F1: 0.10 - 0.18), issues with data validity (F1: 0.05 - 0.28), and misrepresentation of scientific studies show \emph{uniformly} low performance. 
Correct identification often requires aligning textual claims with visual trends and reasoning about omitted baselines, uncertainty, or the plausibility of scientific assertions.
Notably, categories such as Failure to Account for Statistical Nuance and Unclear Encoding also exhibit high false positive rates, indicating that models frequently over-predict these context-dependent errors (See Appendix Table \ref{tab:per_error_fpr_whole_benchmark}).

A similar pattern is observed within visualization error detection. While visually salient distortions are often recognized (e.g., dual-axis), more subtle design flaws, such as uneven binning, inappropriate encodings, or inverted axes, remain difficult for most models. Detecting these errors requires precise spatial comparison or knowledge of visualization design principles, which current VLMs do not consistently demonstrate. As a result, even classic visualization pitfalls evade detection when they lack strong visual regularities.

\subsection*{Finding 4: VLMs frequently over-flag non-misleading visualizations-caption pairs ($\varnothing$) as misleading, indicating a false positive bias.}

\begin{figure}[!h]
    \centering
        \centering
        \includegraphics[trim={15pt 15pt 15pt 15pt}, width=\linewidth]{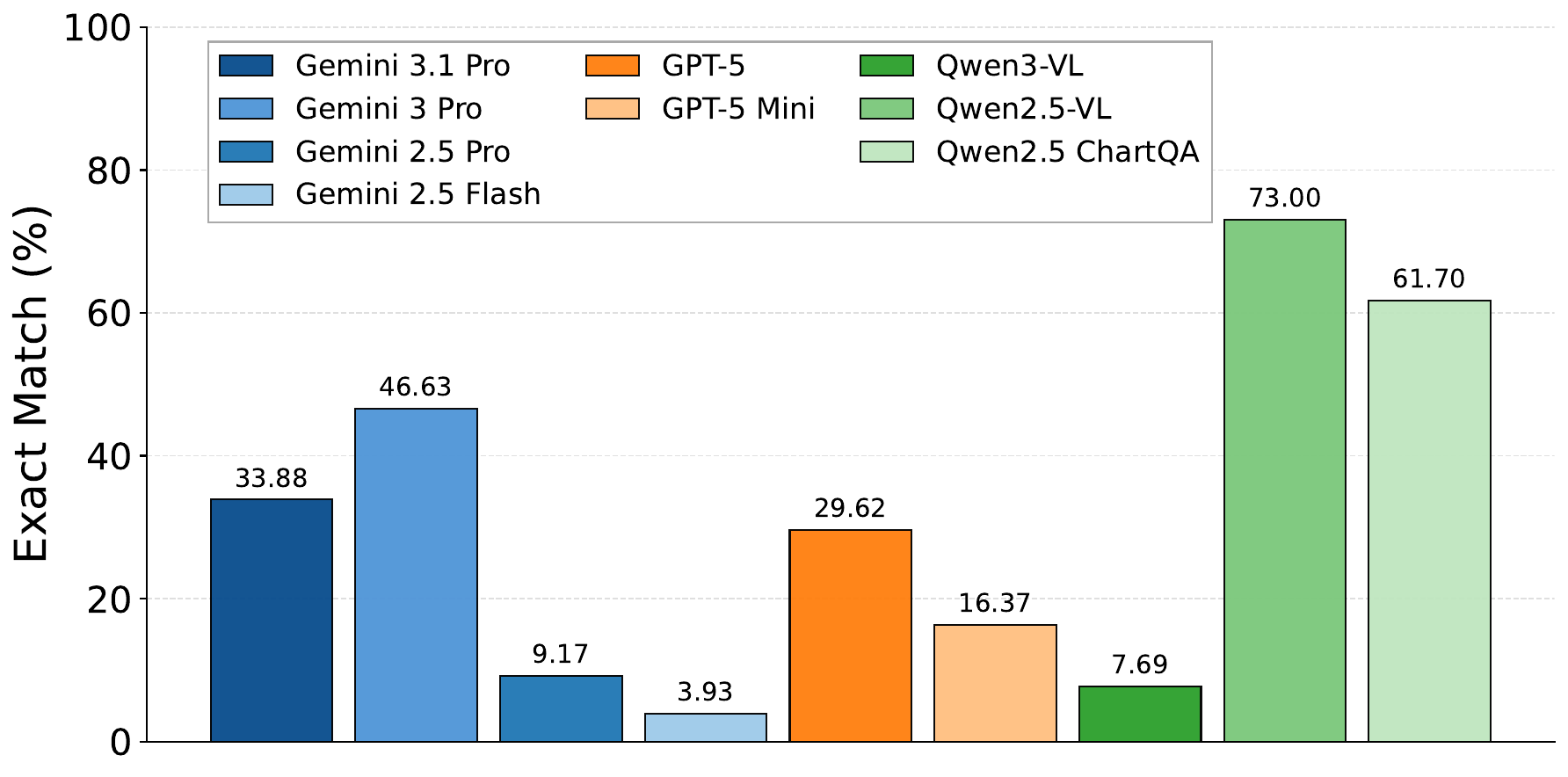}
        \caption{EM scores on the \textit{Non-Misleading Caption, Non-Misleading Visualization (case $\varnothing$}) subset of the benchmark. Most VLMs incorrectly flag clean examples as containing at least one error.}
        \label{fig:clean_data_bar_chart}
        \vspace{-14pt}
\end{figure}

In addition to struggling with accurate error attribution, many VLMs frequently misclassify non-misleading visualization-caption pairs as containing one or more errors (refer to Figure \ref{fig:clean_data_bar_chart}). 
On the subset containing no reasoning or visualization errors (case $\varnothing$), several models achieve low Exact Match rates, indicating frequent false positives even when both modalities are clean. 
This suggests that models often label inputs as misleading even without explicit evidence.

This over-flagging pattern suggests a calibration issue: in the absence of clear error signals, models tend to default toward predicting the presence of an error. In many realistic deployment settings, such as social media platforms, most of the data visualizations are expected to be non-misleading. In such contexts, a tendency to over-flag benign content makes deployment problematic.

Lastly, it is important to note that this tendency is not uniform across models. Some open-source models (\qwentwofivechartqa, \qwentwofive) achieve higher exact-match accuracy on clean examples, but this improved calibration comes at the cost of lower detection rates on misleading cases. 
Detailed false positive rates across error categories are reported in Appendix \ref{app:grid-wise_results}.

\section{Discussion}

Our findings highlight several key considerations for real-world deployment of VLMs in misinformation detection systems. 
A likely deployment scenario for such systems is the monitoring of visualization-caption pairs on social media platforms, where charts are frequently used to support claims in public discourse. 
In these environments, the majority of visualizations are expected to be benign, with misleading cases constituting a relatively small fraction of overall content. Under this base-rate assumption, effective deployment requires not only the ability to flag genuinely misleading visualizations but also the capacity to reliably recognize error-free cases and to provide accurate explanations when intervention occurs.

\textbf{Effective deployment requires both accurate detection and justification.} 
Prior work in visualization research has emphasized that misleadingness is rarely binary, noting that ``the notion that a visualization is either deceptive or not elides the subtlety of many [misinformation] techniques'' \cite{correll2017black}. Our results align with this perspective: although several VLMs achieve moderately high Partial Match scores, their consistently low Exact Match performance indicates that models often detect misleadingness without correctly attributing underlying reasoning or visualization errors. In deployment settings, such partial detection is insufficient. Flagging content as misleading without accurately identifying \emph{why} it is misleading risks producing incorrect or uninformative justifications, limiting the system's utility for diagnosis, explanation, or downstream moderation.

This limitation is particularly consequential in domains such as health communication, political discourse, and financial reporting, where mischaracterizing the basis of a misleading claim can propagate incorrect conclusions or undermine trust.
Accurately distinguishing between different misinformation techniques is not only important for detection, but also for building tools that raise awareness, support human judgment, and mitigate the effects of deceptive framing \cite{correll2017black, chen2021vizlinter}.
As such, while VLMs show \emph{some} potential for detecting misleading visualizations, their deployment as standalone diagnostic systems would require substantial improvements in attribution accuracy and explanation reliability.

\textbf{Reasoning-based errors pose a unique challenge.} Compared to visual distortions, reasoning errors require contextual understanding and alignment between the caption and the visual evidence. Our results show that models consistently underperform on these categories, despite the community view that models handle text more effectively than visual inputs.
Furthermore, visualization researchers have posited that reasoning errors in deceptive visualizations are so persuasive and persistent because they ``generally do not contain formal logical fallacies, as the conclusion always logically follows
from the presented premises'' \cite{lisnic2023misleading}. In such cases, simple fact-checking or surface-level verification is insufficient: effective detection requires identifying how claims are derived from the data, not merely whether the data itself is accurate.

\textbf{Over-flagging further complicates deployment.}
Beyond missed or incomplete detections, our results show that many VLMs frequently misclassify non-misleading visualization-caption pairs as deceptive. This false-positive bias suggests a calibration issue in which models default toward flagging content under uncertainty, rather than reliably recognizing error-free cases. In deployment settings dominated by benign visualizations, this tendency can substantially reduce practical utility by massively flagging samples for human review.

\emph{In the current state of VLMs, we recommend against deploying current VLMs as standalone systems for detecting or moderating misleading data visualizations, and instead limit their use to carefully designed human-in-the-loop settings until substantial improvements are achieved in error attribution, reasoning robustness, and calibration.}

\section{Conclusion}

In this work, we evaluated whether current vision-language models can detect misinformation in visualization-caption pairs by jointly modeling visual design errors and caption-level reasoning errors. Unlike prior benchmarks that largely assume truthful visualizations, our benchmark explicitly includes reasoning errors in captions, which is an important but often neglected dimension of real-world misinformation. Across a diverse set of models, we find that while VLMs are relatively effective at identifying certain perceptual distortions, they struggle substantially with subtle design guideline violations and reasoning errors that require contextual interpretation, statistical nuance, or the evaluation of claims against the visual evidence. These results suggest that strong performance on chart understanding benchmarks does not directly translate into robust multimodal misinformation detection. Our benchmark and findings highlight concrete limitations of current VLMs and provide a foundation for developing models and evaluations that better emulate how misinformation manifests in real-world visual communication.

\section*{Limitations}
Limitations to our work are as follows: (1) No task-specific fine-tuning. We evaluate models using their default inference configurations and do not explore whether task-specific fine-tuning or instruction tuning on our benchmark could improve performance. While this choice reflects realistic out-of-the-box deployment scenarios, fine-tuning may meaningfully alter both detection accuracy and calibration behavior. (2) Model coverage is not exhaustive. Although we evaluate a diverse set of proprietary and open-source VLMs based on our resource constraints, our open-source analysis primarily focuses on the Qwen family. Future work could extend this evaluation to a broader range of community models to better assess generalizability across architectures and training paradigms. (3) Static visualizations only. Our benchmark focuses on static chart-caption pairs and excludes interactive or animated visualizations, which are increasingly common in online settings. Detecting misleadingness in such formats may pose additional challenges not captured here. (4) No external knowledge or verification tools. Unlike current agentic systems, models are evaluated without access to external sources, such as fact-checking databases or domain-specific knowledge bases. As a result, performance on reasoning errors involving data validity or scientific misrepresentation may underestimate what could be achieved with retrieval-augmented or tool-assisted systems. (5) No downstream user impact analysis. Our evaluation focuses on model performance and error attribution accuracy, and does not examine how model outputs influence human judgment, trust, or decision-making. Understanding how partial detections or incorrect explanations affect users is an important direction for future work. (6) Inherited dataset characteristics. A subset of our benchmark reuses samples from \citet{lisnic2023misleading}. As with any such dataset, this portion inherits the characteristics and potential limitations of the original resource, while the remainder of our benchmark is constructed from additional sources.

\section*{Ethical considerations}

This work relies on a benchmark constructed from publicly available visualization-caption pairs sourced from online platforms such as X (formerly Twitter) and Reddit. 

\textbf{Data sourcing and platform compliance.}
All visualizations in the benchmark are derived from publicly available posts on X and Reddit. Note: \citet{lisnic2023misleading} provided curated sets of visualizations from \emph{X} in the form of tweet ID and their own annotations. To adhere to platform policies and content-sharing requirements, we do not redistribute raw social media content. Instead, we release only tweet IDs and Reddit post IDs, allowing data to be extracted (rehydrated) in accordance with the respective platforms' terms of service. We do not include private, deleted, or access-restricted content, nor do we collect or infer personally identifiable or sensitive user information. 

\textbf{Annotation and caption construction.}
To enable controlled analysis of misinformation mechanisms, some captions are human-authored (by the project team) or curated to introduce specific reasoning errors grounded in prior visualization research. These captions are intended to model common misleading practices rather than to endorse the claims they express. Dataset documentation distinguishes between original and researcher-authored content.

\textbf{Risks of misuse and over-interpretation.}
Because the benchmark labels specific reasoning and visualization errors, models trained or evaluated on it could be misused for unsupervised moderation or for generating misleading content.

Overall, our dataset is intended to support diagnostic evaluation and responsible research on multimodal misinformation detection, and should be used with an awareness of its scope and limitations.

\bibliography{custom}
\newpage
\appendix
\section{Appendices}
\label{sec:appendices}

\subsection{Prompts used in the paper}
\label{app:prompts}

The prompts used for our tasks were iteratively refined in consultation with visualization experts on the author team. These experts reviewed early versions of the prompt and provided feedback on clarity and specificity. 
In each round, experts reviewed the task instructions, the definitions of reasoning and visualization error categories, and representative examples. Their review protocol focused on three aspects. 
First, they validated the error definitions to ensure that categories were theoretically grounded. Second, they improved instruction clarity by revising ambiguous phrasing. 
Third, they tested the prompts on representative sample cases across all conditions and reviewed model outputs to identify potential confusions.
Based on this feedback, we arrived at the final versions listed below.

\begin{tcolorbox}[title=Reasoning Error Classification Prompt,
                  colback=gray!5,
                  colframe=black,
                  boxrule=0.5pt,
                  arc=2pt]
\small
You will be provided with a visualization, its accompanying caption,
and descriptions of reasoning errors. These reasoning errors represent
ways in which people use captions to spread misinformation. \\

Your task is to carefully examine the image and its accompanying caption.
Then, based on the information and the descriptions of reasoning errors,
determine which kinds of misinformation, if any, are being propagated. \\

If none of the reasoning errors apply, classify the reasoning error as
"None." \\

Please classify which reasoning errors are present and explain your
reasoning. If more than one classification applies, include all applicable
classifications in a list. Even if only one classification applies, the
"classification" field must still be a list. \\

Only provide output in the following JSON format: \\

\{ \\

"reason": "[Explanation]", \\

"classification": ["Cherry-picking/Causal inference/Setting an arbitrary
threshold/Failure to account for statistical nuance/Incorrect reading of
chart/Issues with data validity/Misrepresentation of scientific studies/None"] \\

\} \\

Image: \{image\} \\

Accompanying Text: \{caption\} \\

Error Descriptions: \{reasoning\_error\_descriptions\} \\

\end{tcolorbox}

\begin{tcolorbox}[title=Visualization Error Classification Prompt,
                  colback=gray!5,
                  colframe=black,
                  boxrule=0.5pt,
                  arc=2pt]

\small
You will be provided with a visualization, its accompanying caption, and descriptions of the visualization error. These visualization errors represent ways in which people use visualization to spread misinformation. \\

Your task is to carefully examine the image and its accompanying caption. Then, based on the information and the descriptions of visualization errors, determine which kinds of misinformation, if any, are being propagated here. \\

If none of the visualization errors apply, you may classify the visualization error as "None." \\

Please classify which visualization errors are present and explain your reasoning. If more than one classification applies, include all applicable classifications in a list. Even if there is only one classification, the "classification" field must still be a list. \\

Only provide output in the following JSON format: \\

\{ \\

"reason": "[Explanation]", \\

"classification": ["Truncated axis/Dual axis/Value as area or volume/Inverted axis/Uneven binning/Unclear encoding/Inappropriate encoding/None"] \\

\} \\

Image: \{image\} \\

Accompanying Text: \{caption\} \\

Error Descriptions: \{visualizaton\_error\_descriptions\}

\end{tcolorbox}

\subsection{Error Descriptions}
\label{app:error_descriptions}

We present detailed descriptions of the reasoning (Table \ref{tab:reasoning_error_descriptions}) and visualization errors (Table \ref{tab:visualization_error_descriptions}) used in our benchmark. 
The descriptions are adapted from \citet{lisnic2023misleading}, and further refined through feedback from visualization graduate students.

\begin{table*}[!h]
\centering
\resizebox{0.9\linewidth}{!}{
\renewcommand{\arraystretch}{1}
\begin{tabular}{p{3cm} p{14.5cm}}
\toprule
\textbf{Error Type} & \textbf{Description} \\
\midrule
Setting an Arbitrary Threshold

\emph{\small Shorthand: Arb. Thr.} & Setting a benchmark or threshold that lacks a solid factual basis or recognition by standard authorities. The arbitrary threshold is used to judge or compare data, leading to potentially misleading conclusions because it appears meaningful but isn't supported by official criteria.

Key Characteristics:

- Unjustified Benchmarks: The threshold is chosen without logical reasoning, often to support a specific argument.

- Selective Highlighting: Data aligning with the threshold is emphasized, ignoring the broader context.

- Visual Manipulation: Annotations or labels make the threshold seem more significant than it is.

- Lack of Context: The threshold is presented without proper context or comparison to recognized standards.

 \\
\midrule
Cherry-picking

\emph{\small Shorthand: Cher. Pick.}  & Selectively presenting data points that support a specific argument while ignoring those that don't. This can create a biased and misleading representation of the data. 

Key Characteristics:

- Selective Data Points: Only data that supports the argument is shown.

- Ignoring Context: Broader data context that might contradict the argument is omitted.

- Overemphasis: Overemphasizing certain data points to sway opinion.

\\
\midrule
Causal Inference 

\emph{\small Shorthand: Caus. Infer.} & Assuming a cause-and-effect relationship between two variables based on their correlation, without sufficient evidence to support such a claim.

Key Characteristics:

- Correlation Assumed as Causation: Assuming that because two variables are correlated, one must cause the other.

- Lack of Evidence: No rigorous evidence to support the causal link.

- Ignoring Other Factors: Failing to consider other variables that might influence the outcome.

\\
\midrule
Issues with Data Validity 

\emph{\small Shorthand: Data Val.}  & Questioning the accuracy or reliability of the data without sufficient justification, often to cast doubt on the conclusions drawn from the data.

Key Characteristics:

- Questioning Data Accuracy: Raising doubts about the data without solid evidence.

- Suggesting Manipulation: Implying that data has been manipulated to fit a narrative.

- Ignoring Explanations: Overlooking valid reasons for data inconsistencies.

 \\
 \midrule
Failure to Account for Statistical Nuance 

\emph{\small Shorthand: Stat. Nu.} & Oversimplifying complex statistical data, and ignoring important details that are crucial for accurate interpretation.
    
Key Characteristics:

- Oversimplification: Ignoring complex statistical relationships and nuances.

- Lack of Comparison: Failing to compare with relevant control groups or baselines.

- Misinterpretation: Drawing conclusions without considering statistical significance or variability.

 \\
 \midrule
Misrepresentation of Scientific Studies 

\emph{\small Shorthand: Mis. Sci.}  & Selectively citing studies or exaggerating their findings to support a specific argument, often ignoring the broader scientific consensus. 

Key Characteristics:

- Selective Citation: Citing only studies that support the argument.

- Exaggeration: Overstating the significance or certainty of study findings.

- Ignoring Consensus: Overlooking the broader context or scientific consensus.

\\
\midrule
Incorrect Reading of the Chart 

\emph{\small Shorthand: Chart Read}  & Misinterpreting the data presented in a chart, often due to visual distortions or a lack of understanding of the chart's design.

Key Characteristics:

- Visual Distortion: Misinterpreting data due to design issues like truncated axes or misleading scales.

- Misreading Data: Incorrectly interpreting the data points or trends shown in the chart.

- Lack of Understanding: Failing to understand the chart's design or the data it represents.

 \\
\bottomrule
\end{tabular}
}
\caption{Descriptions of reasoning errors in captions that contribute to misinformation. The shorthand names are used in subsequent tables for compact reporting.}
\label{tab:reasoning_error_descriptions}
\end{table*}

\begin{table*}[!h]
\centering
\resizebox{0.9\linewidth}{!}{
\renewcommand{\arraystretch}{1}
\begin{tabular}{p{3cm} p{14.5cm}}
\toprule
\textbf{Error Type} & \textbf{Description} \\
\midrule
Truncated Axis 

\emph{\small Shorthand: Trunc. Axis}  & Shortening the axis scale in a chart to exaggerate the appearance of differences or trends in the data.

Key Characteristics:

- Exaggerated Trends: Small differences in data appear more significant due to axis truncation.

- Misleading Scales: The axis starts at a value other than zero without clear justification.

- Distorted Proportions: Viewers perceive larger changes than actually exist.

 \\
 \midrule
Dual Axis 

\emph{\small Shorthand: Dual Axis}  & Using two separate vertical axes to plot unrelated or loosely related variables, often creating misleading visual correlations.

Key Characteristics:

- Misaligned Scales: The scales of the two axes are unrelated, leading to false visual patterns.

- Forced Correlation: Unrelated datasets appear correlated due to shared chart space.

- Overloading Information: Multiple axes make the chart harder to interpret accurately.

 \\
 \midrule

Value as Area or Volume 

\emph{\small Shorthand: Area/Vol.} & Using shapes or 3D volumes to represent data although it is known that humans are poor at visually distinguishing differences in area or volume.

Key Characteristics:

- Exaggerated Perception: Changes in size appear larger than the actual proportional difference.

- Misleading Scaling: Areas or volumes do not accurately reflect the data values.

- Ineffective Comparison: Viewers struggle to interpret exact values or differences.

 \\
 \midrule
Inverted Axis 

\emph{\small Shorthand: Inv. Axis}  & Reversing the direction of an axis, which can confuse the audience and lead to misinterpretation of trends or comparisons. 

Key Characteristics:

- Reversed Direction: An axis increases in value downward or to the left instead of the standard upward or rightward directions.

- Misleading Trends: Data trends appear opposite to their actual direction.

- Lack of Clarity: The inversion is not clearly labeled or explained.

\\
\midrule
Uneven Binning 

\emph{\small Shorthand: Uneven Bin.}  & Grouping data into bins of unequal size or creating bins that do not span the data distribution, leading to biased or misleading visual distributions.

Key Characteristics:

- Inconsistent Intervals: Bin sizes vary without justification, skewing the data representation.

- Disproportionate Emphasis: Certain bins appear more significant due to size differences.

- Misleading Comparisons: Data is harder to compare accurately across bins.

 \\
 \midrule
Unclear Encoding  

\emph{\small Shorthand: Unclr. Enc.} & Using visual elements that are difficult to interpret or lack sufficient labeling, leading to confusion about what the chart represents.

Key Characteristics:

- Ambiguous Visuals: Symbols, colors, or patterns are non-standard or not clearly explained.

- Missing Labels: Key elements like axes, legends, or annotations are absent or unclear.

- Overloaded Design: Too many visual elements representing multiple data variables in a single chart, making interpretation difficult.

 \\
 \midrule
Inappropriate Encoding 

\emph{\small Shorthand: Inappr. Enc.}  & Choosing a visual representation that is unsuitable for the type of data, making interpretation inaccurate or misleading.

Key Characteristics:

- Misaligned Visuals: The chosen chart type or visual encoding does not match the data variable.

- Distorted Representation: Data relationships are inaccurately emphasized or diminished, resulting in ineffective comparisons. 

\\
\bottomrule
\end{tabular}
}
\caption{Descriptions of visualization errors that mislead viewers through visualization design. The shorthand names are used in subsequent tables for compact reporting.}
\label{tab:visualization_error_descriptions}
\end{table*}
\clearpage
\newpage

\onecolumn
\subsection{Examples from the Benchmark}
\label{app:examples_from_benchmark}

\begin{table*}[!h]
\centering
\resizebox{0.9\textwidth}{!}{
\renewcommand{\arraystretch}{1}
\begin{tabular}{
  m{0.4\textwidth}
  >{\RaggedRight\arraybackslash}m{0.3\textwidth}
  >{\centering\arraybackslash}m{0.3\textwidth}
}
\toprule
\textbf{Visualization} & \textbf{Caption} & \textbf{Reasoning Error} \\
\midrule

\includegraphics[scale=0.12]{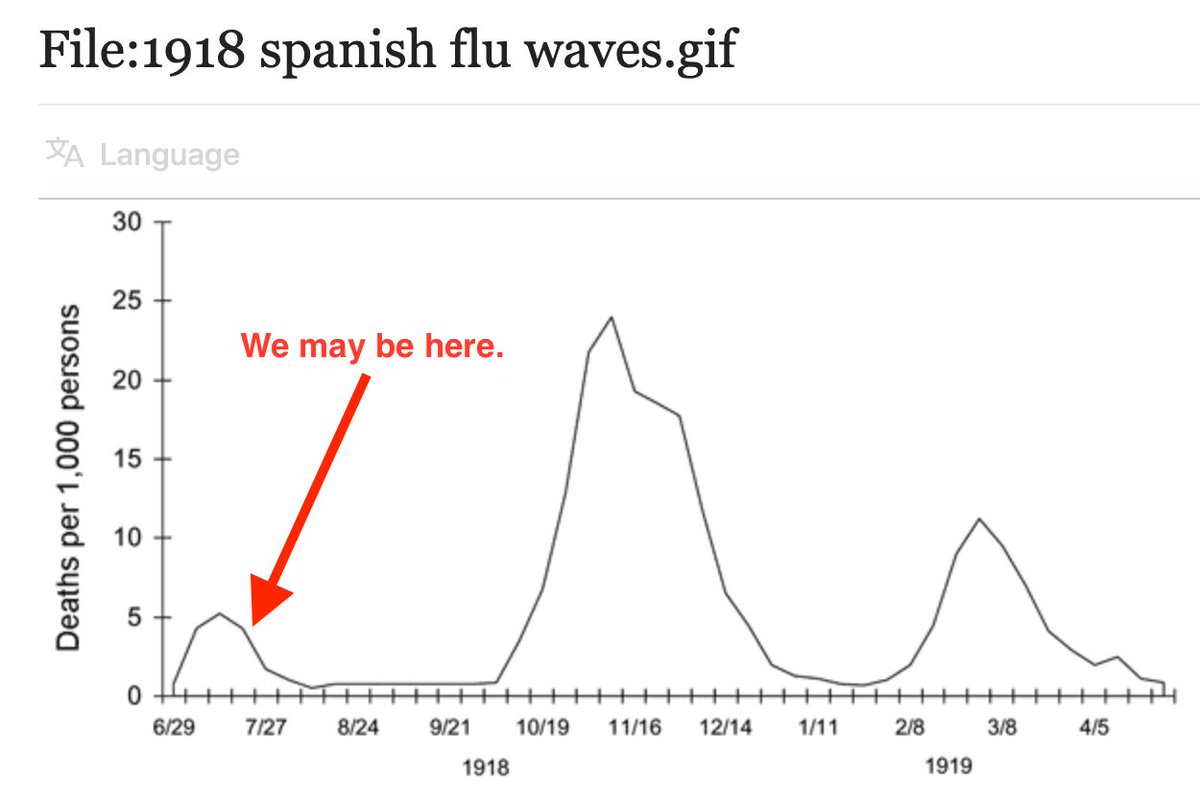} & Reminder: Just because we've hit a peak does not mean we've hit THE peak. & \textbf{Cherry-picking} \\

\hline

\includegraphics[scale=0.12]{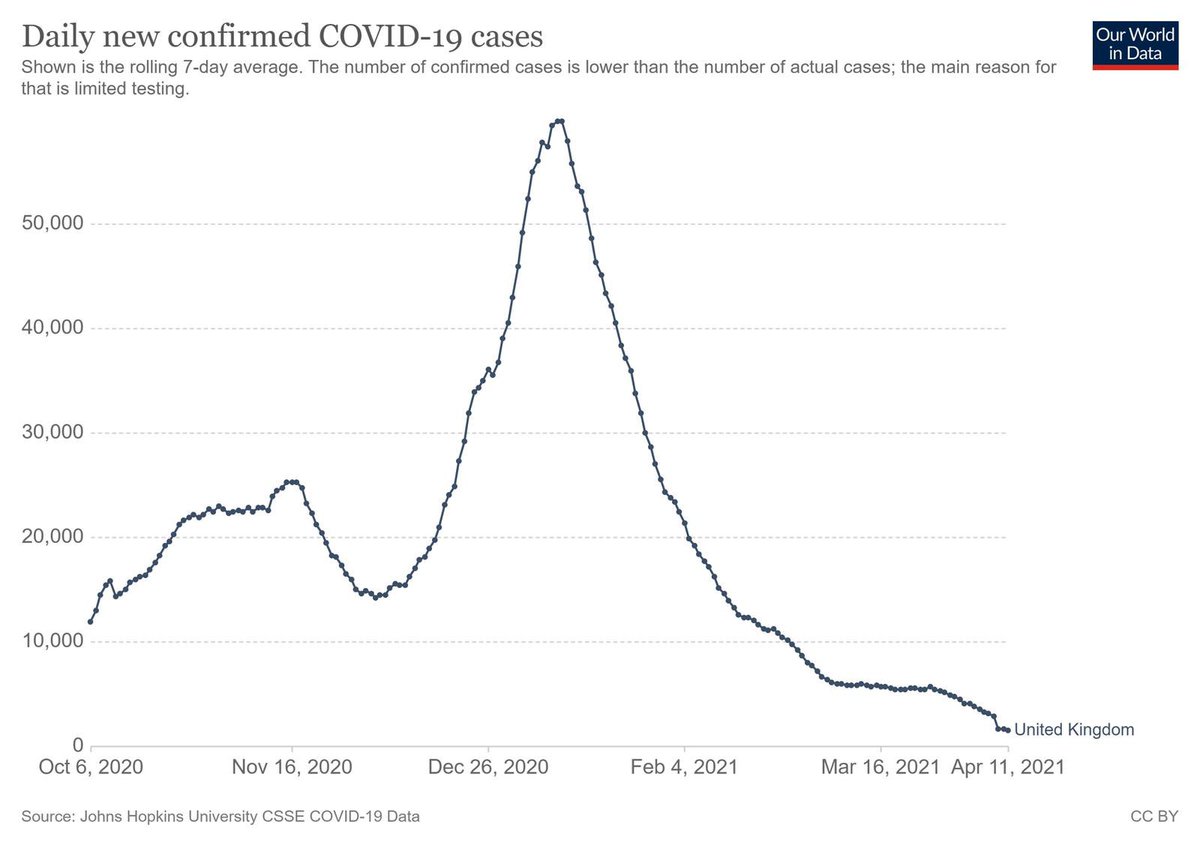} & The positive impact of the UK's vaccination efforts in one graph & \textbf{Causal inference} \\

\hline

\includegraphics[scale=0.12]{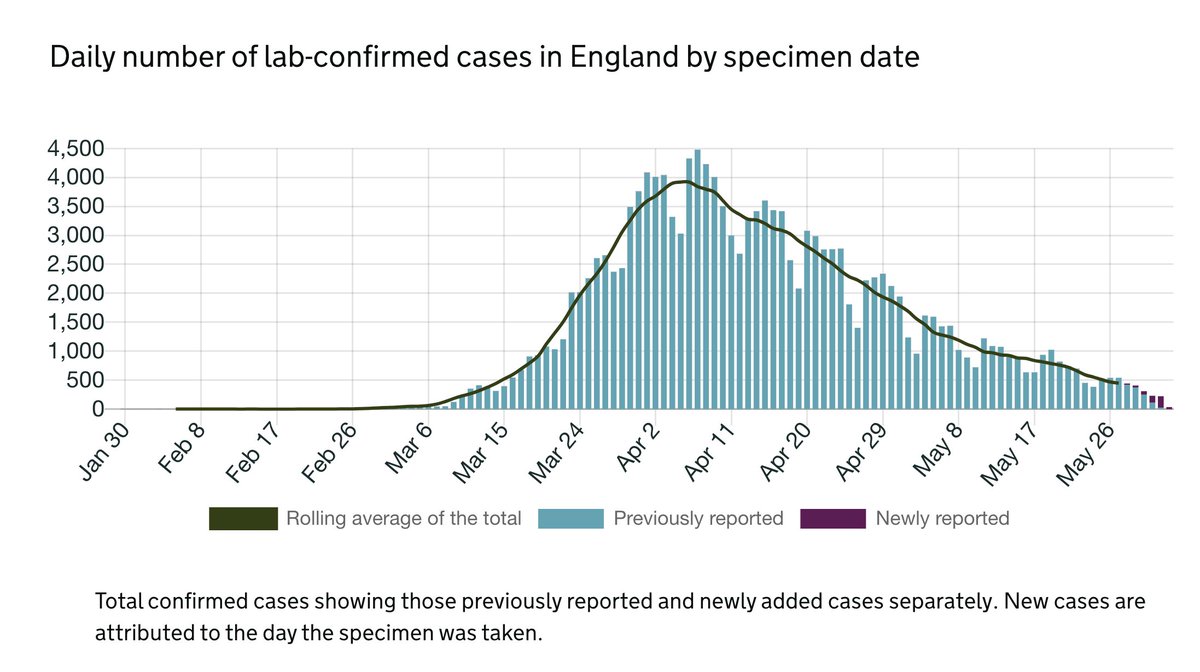} & This in a country of 56 million. Lift lockdown now, the virus is just gone. & \textbf{Setting an arbitrary threshold} \\

\hline

\includegraphics[scale=0.12]{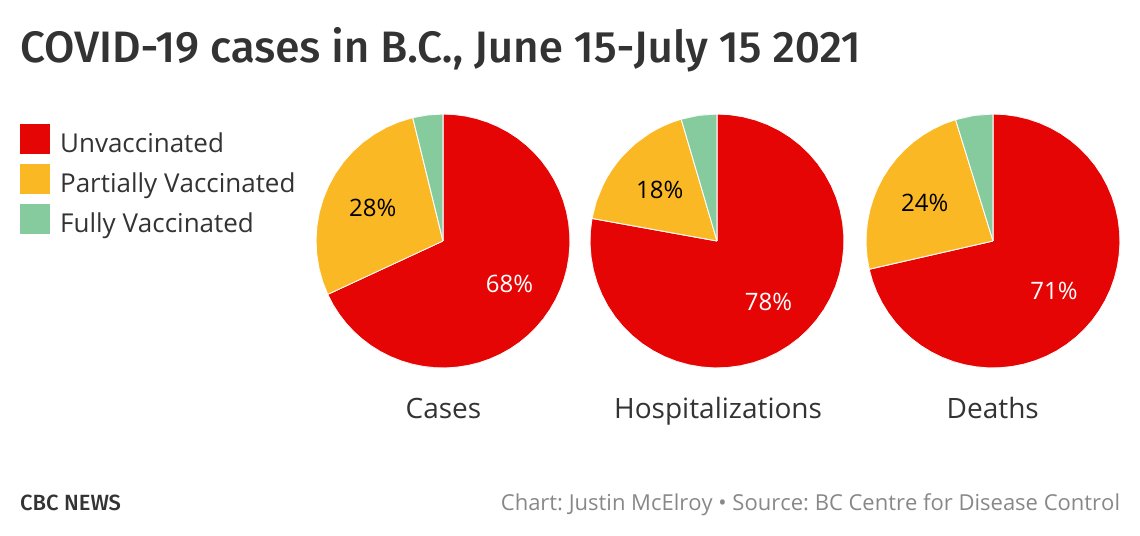} & The numbers absolutely speak for themselves. Get vaccinated! & \textbf{Failure to account for statistical nuance} \\

\hline

\includegraphics[scale=0.12]{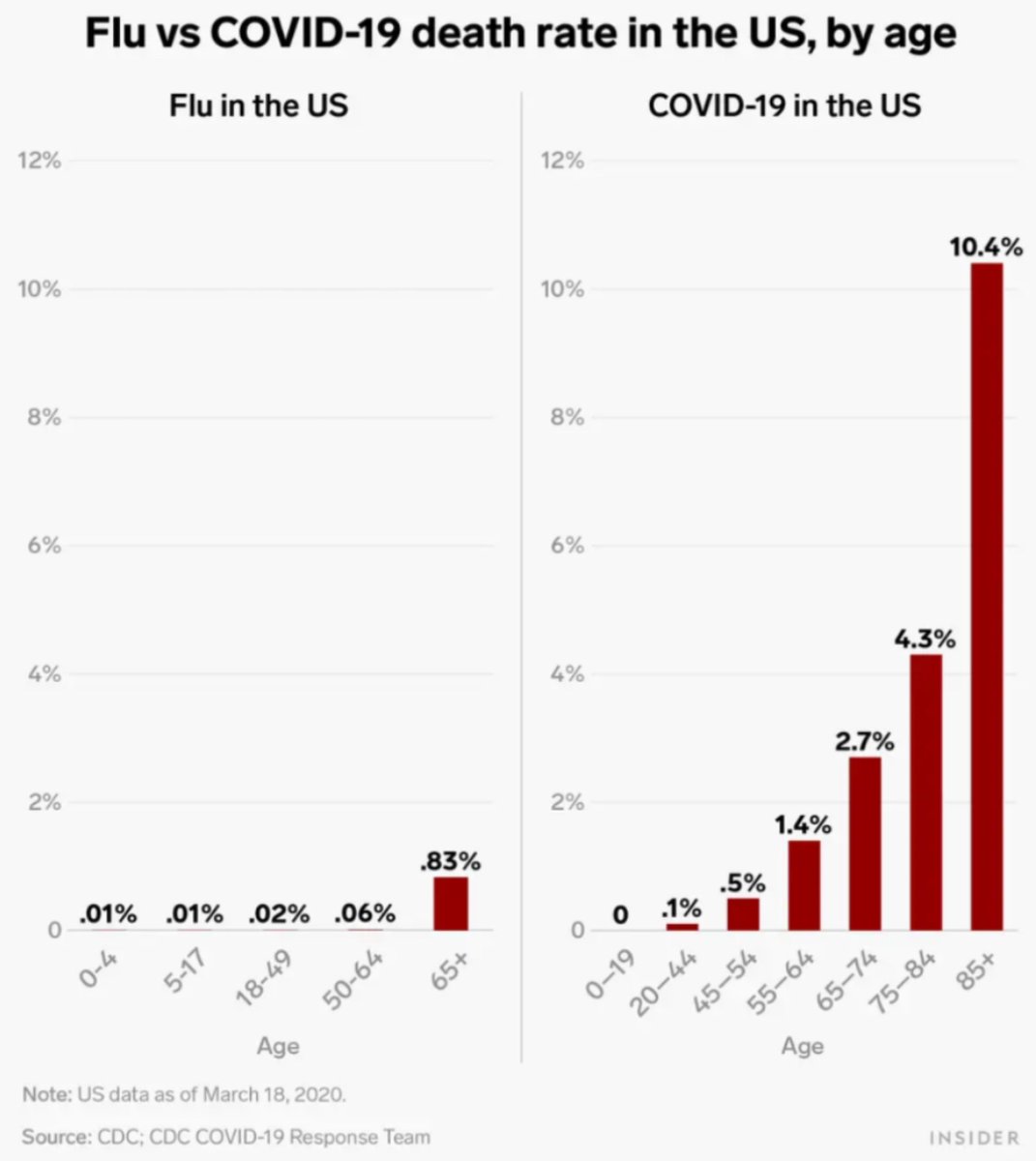} & The flu is 10 times less deadly - particularly for elderly - than Covid! & \textbf{Incorrect reading of chart} \\

\hline

\includegraphics[scale=0.12]{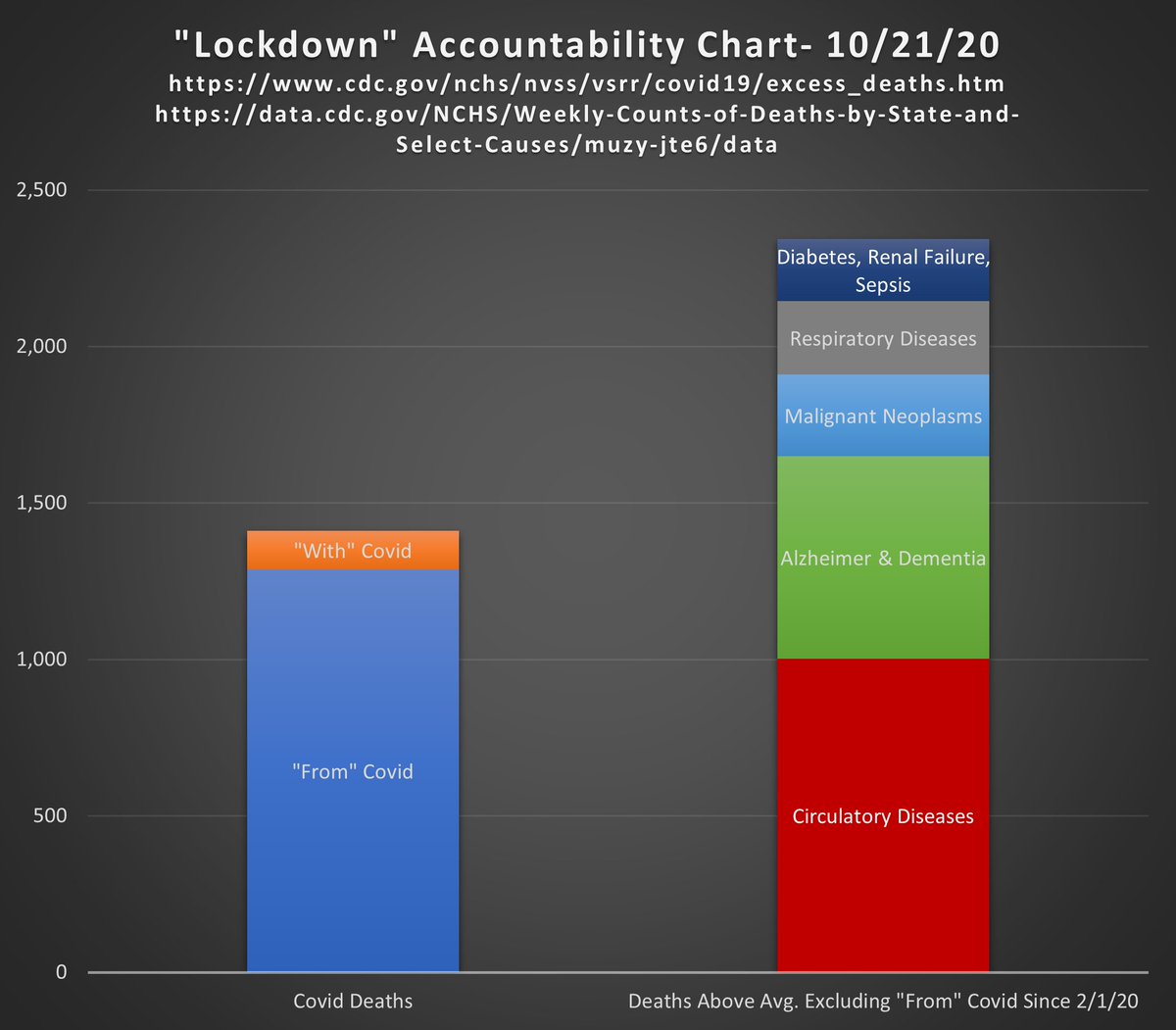} & This is a test of our humanity & \textbf{Issues with data validity} \\

\hline

\includegraphics[scale=0.12]{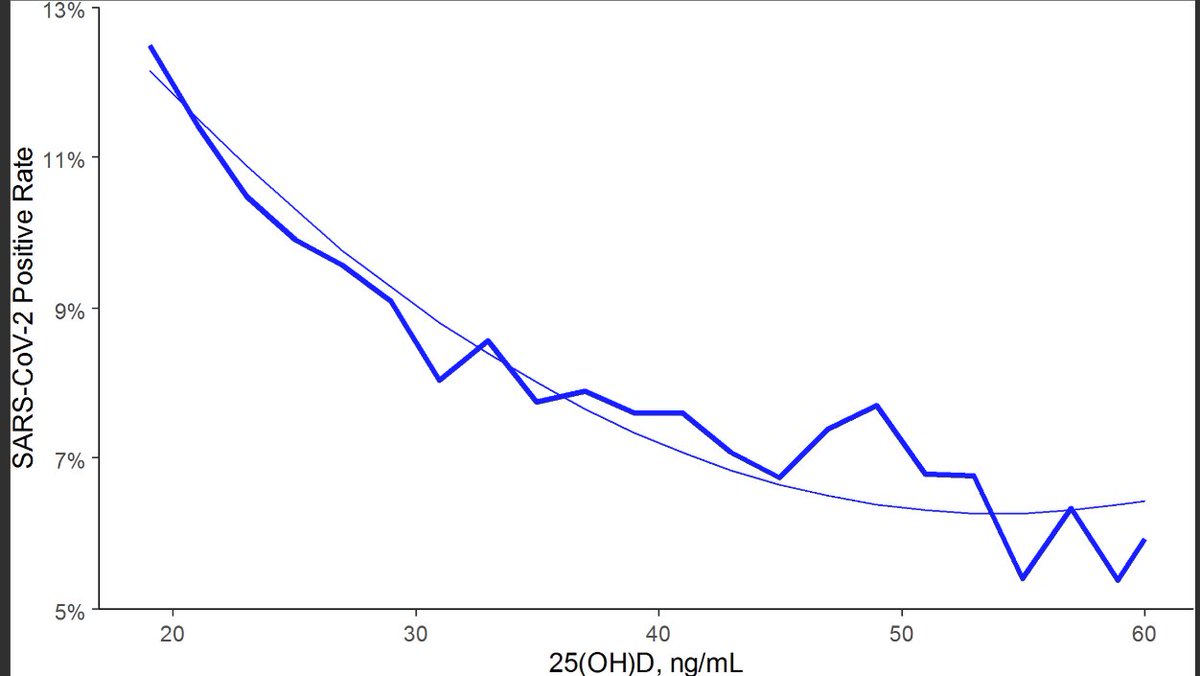} & SARS-CoV-2 positivity rates associated with circulating 25-hydroxyvitamin D levels (https://tinyurl.com/5n9xm536) & \textbf{Misrepresentation of scientific studies} \\

\bottomrule
\end{tabular}
}
\caption{Examples of misleading captions and their associated errors paired with visualizations.}
\label{tab:reasoning_err_eg}
\end{table*}

\begin{table*}[t]
\centering
\resizebox{0.9\textwidth}{!}{
\renewcommand{\arraystretch}{1}
\begin{tabular}{
  m{0.4\textwidth}
  >{\RaggedRight\arraybackslash}m{0.3\textwidth}
  >{\centering\arraybackslash}m{0.3\textwidth}
}
\toprule
\textbf{Visualization} & \textbf{Caption} & \textbf{Reasoning Error} \\
\midrule

\includegraphics[scale=0.12]{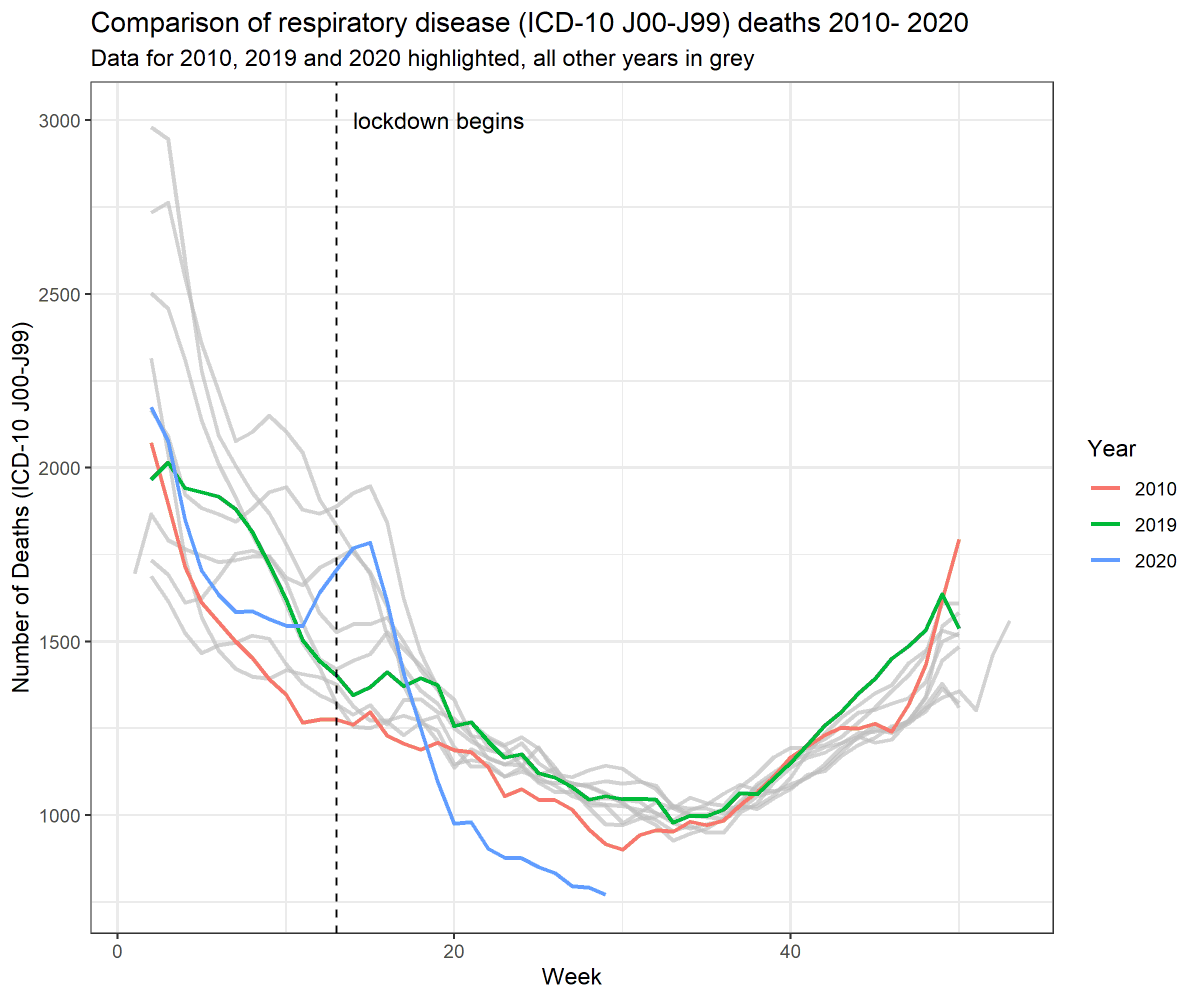} & Respiratory deaths at 10 year low! & \textbf{Truncated axis} \\

\hline

\includegraphics[scale=0.12]{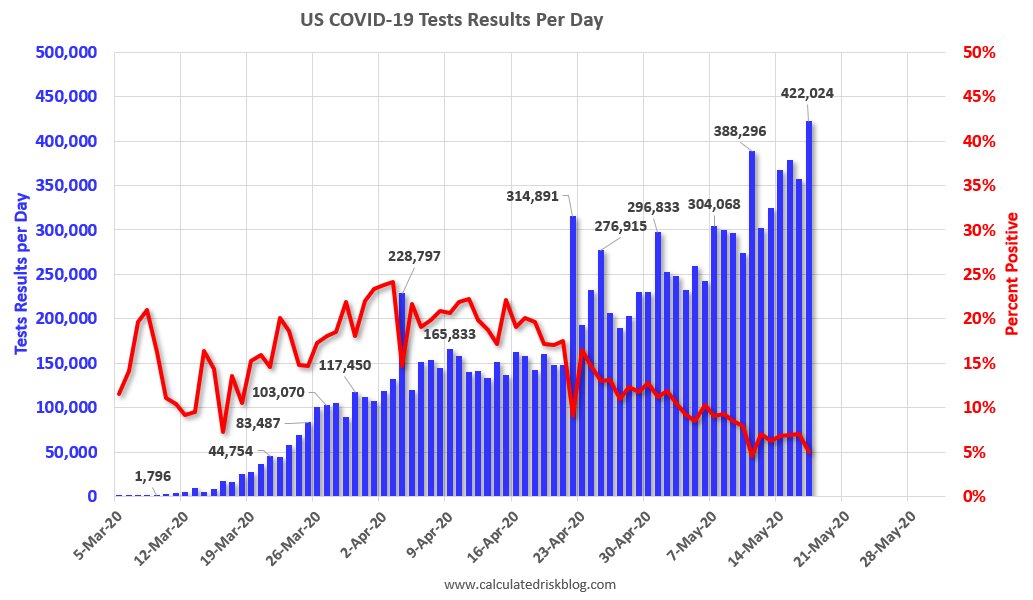} & May 17 Update: US COVID-19 Test Results: Test-and-Trace Success for Smallpox & \textbf{Dual Axis} \\

\hline

\includegraphics[scale=0.12]{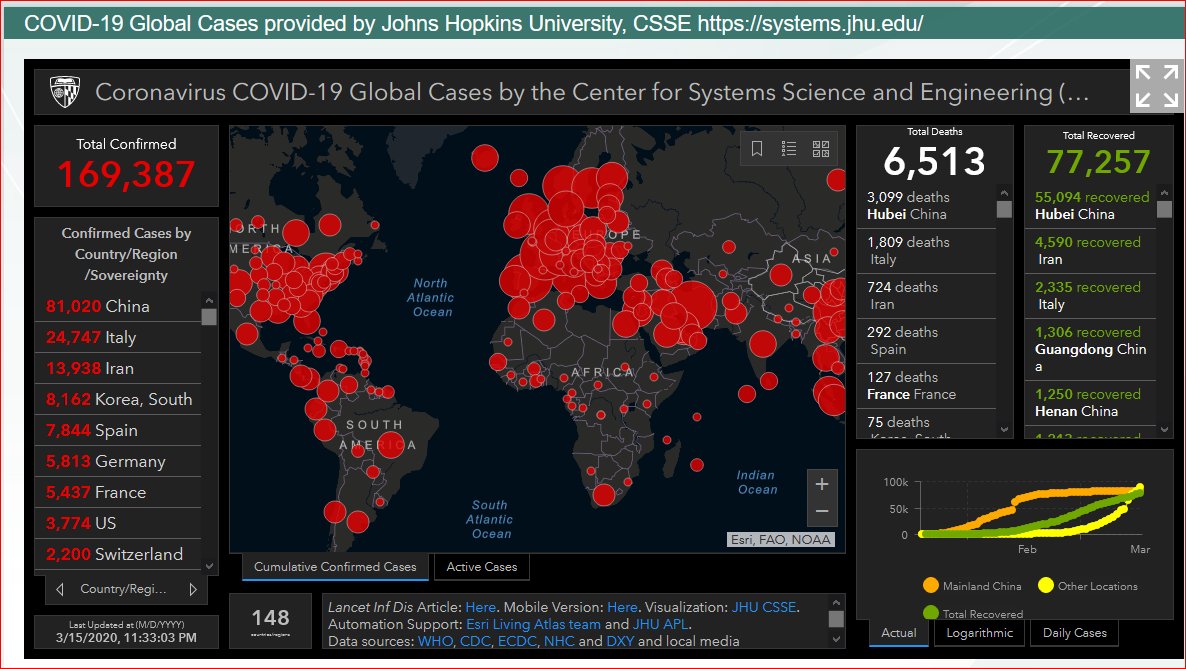} & Corona Virus Interactive Map. & \textbf{Value as area or volume} \\

\hline

\includegraphics[scale=0.12]{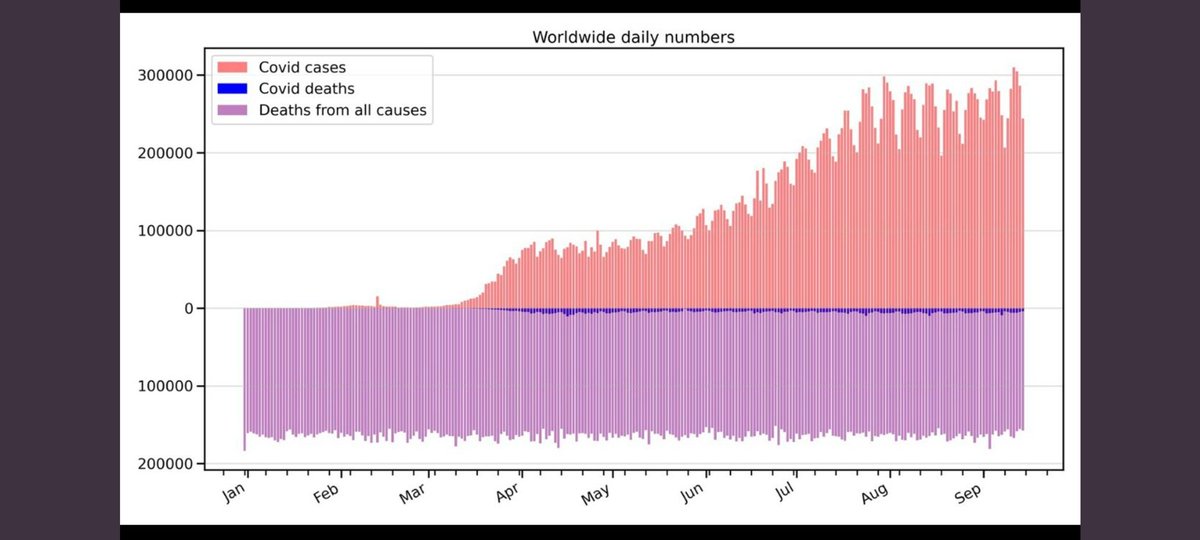} & Propaganda: RECORD NUMBER OF COVID POSITIVE CASES. Reality: & \textbf{Inverted axis} \\

\hline

\includegraphics[scale=0.12]{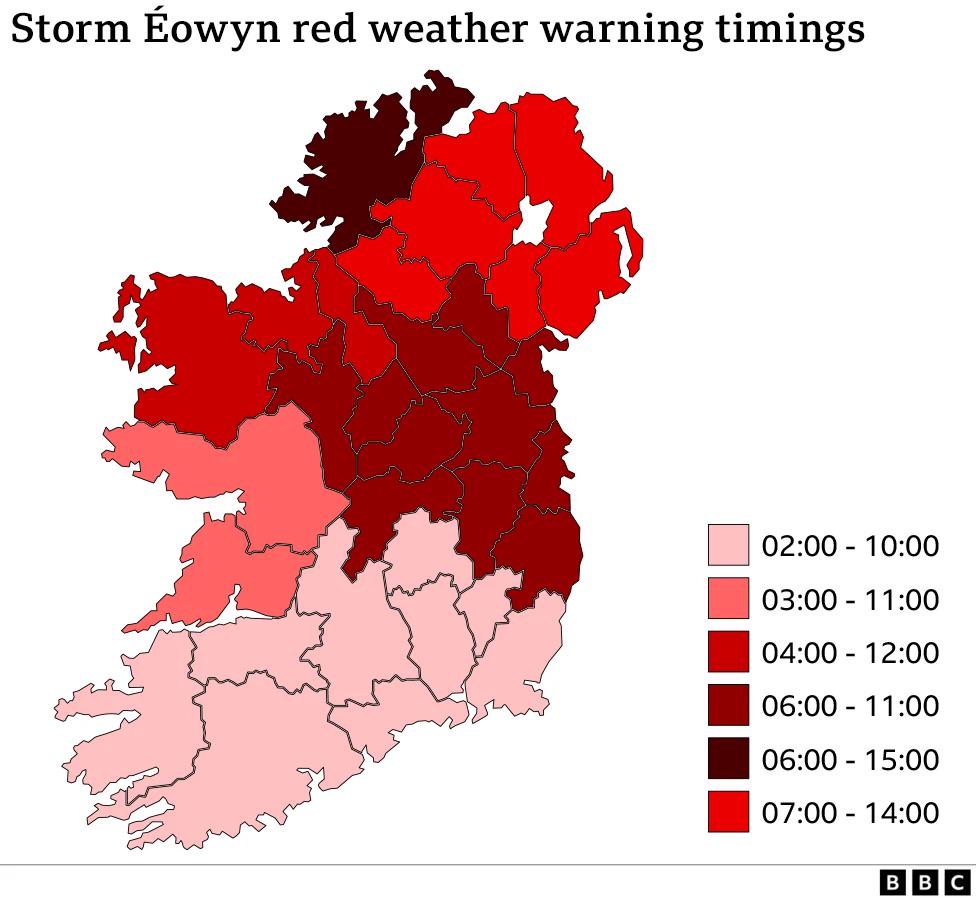} & Interesting colour coding from the BBC & \textbf{Uneven binning} \\

\hline

\includegraphics[scale=0.12]{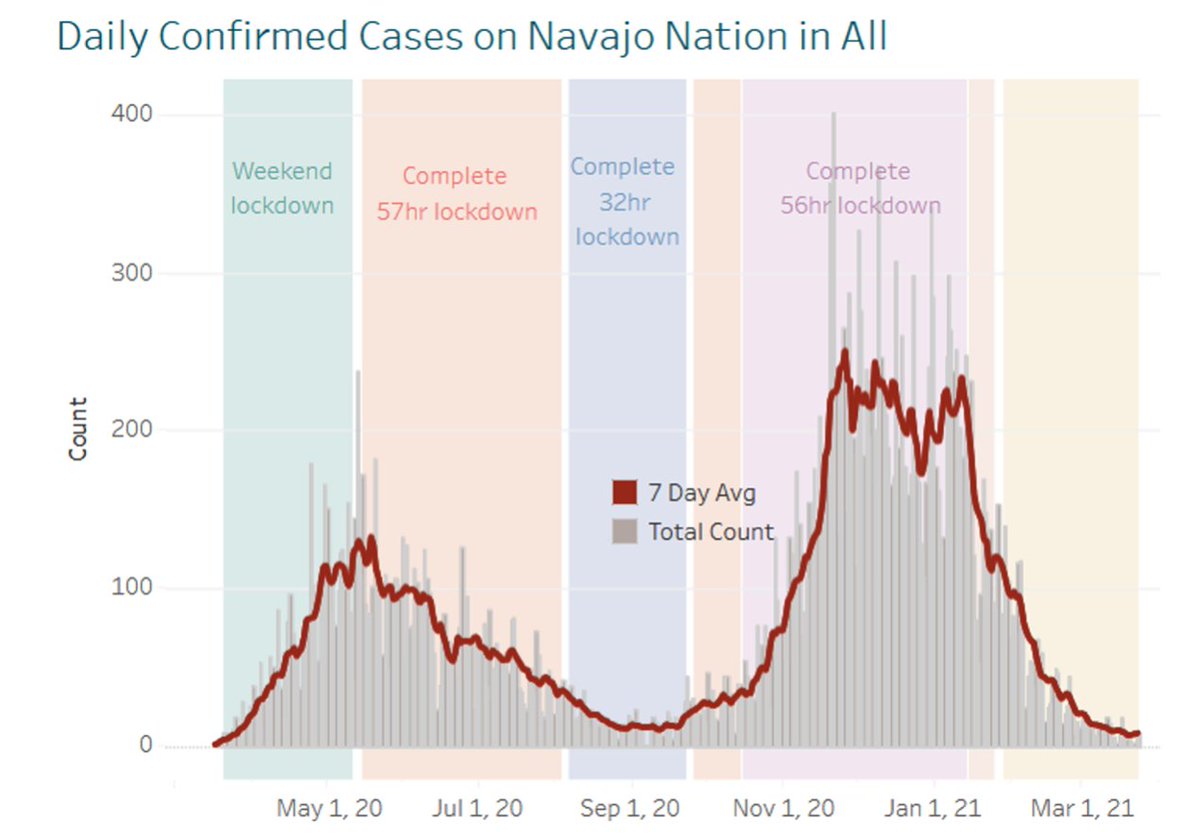} & The Navajo Nation crushed the Covid curve. Success is possible. & \textbf{Unclear encoding} \\

\hline

\includegraphics[scale=0.12]{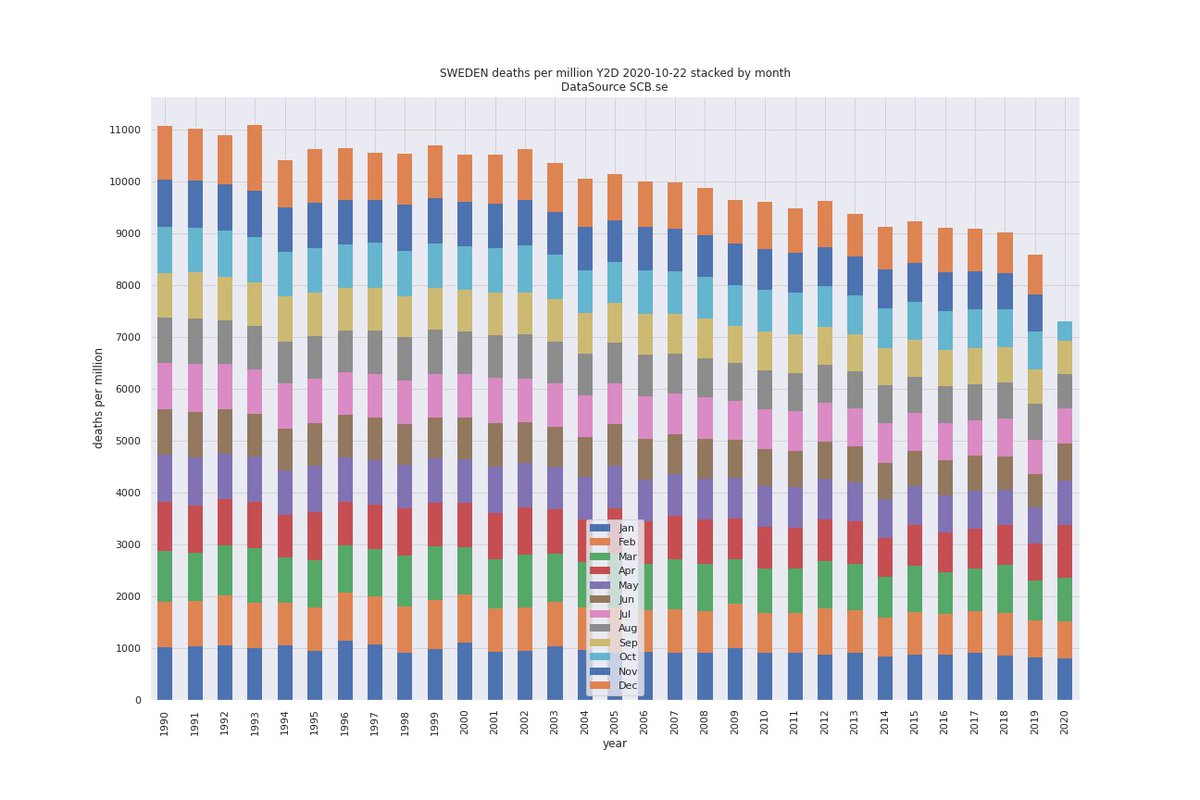} & The worst pandemic of the most contagious disease we have seen for 100 years. & \textbf{Inappropriate encoding} \\

\bottomrule
\end{tabular}
}
\caption{Examples of misleading visualizations and their associated errors paired with captions.}
\label{tab:visualization_error_eg}
\end{table*}

\clearpage
\newpage

\subsection{Reasoning and Error Compositions across the Benchmark}
\label{app:reasning_viz_composition}

\begin{figure*}[!h]
    \centering
        \hspace{-2cm} 
        \centering
        \includegraphics[scale=0.6]{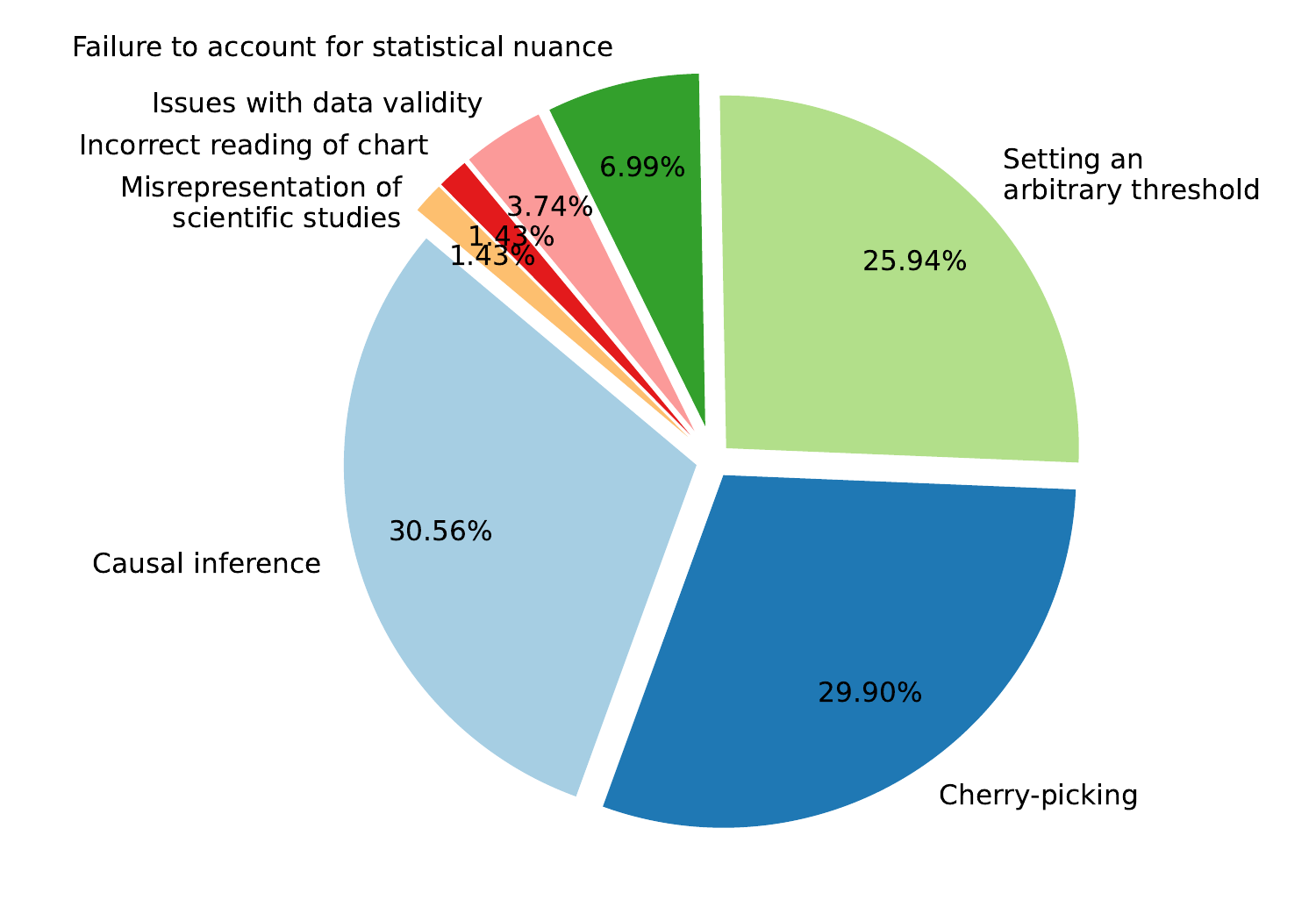}
        \vspace{-20pt}
        \caption{Reasoning Error Composition for the dataset.}
        \label{fig:reasoning_bar_chart}
        \vspace{-10pt}
\end{figure*}

\begin{figure*}[!h]
    \centering
        \centering
        \includegraphics[scale=0.6]{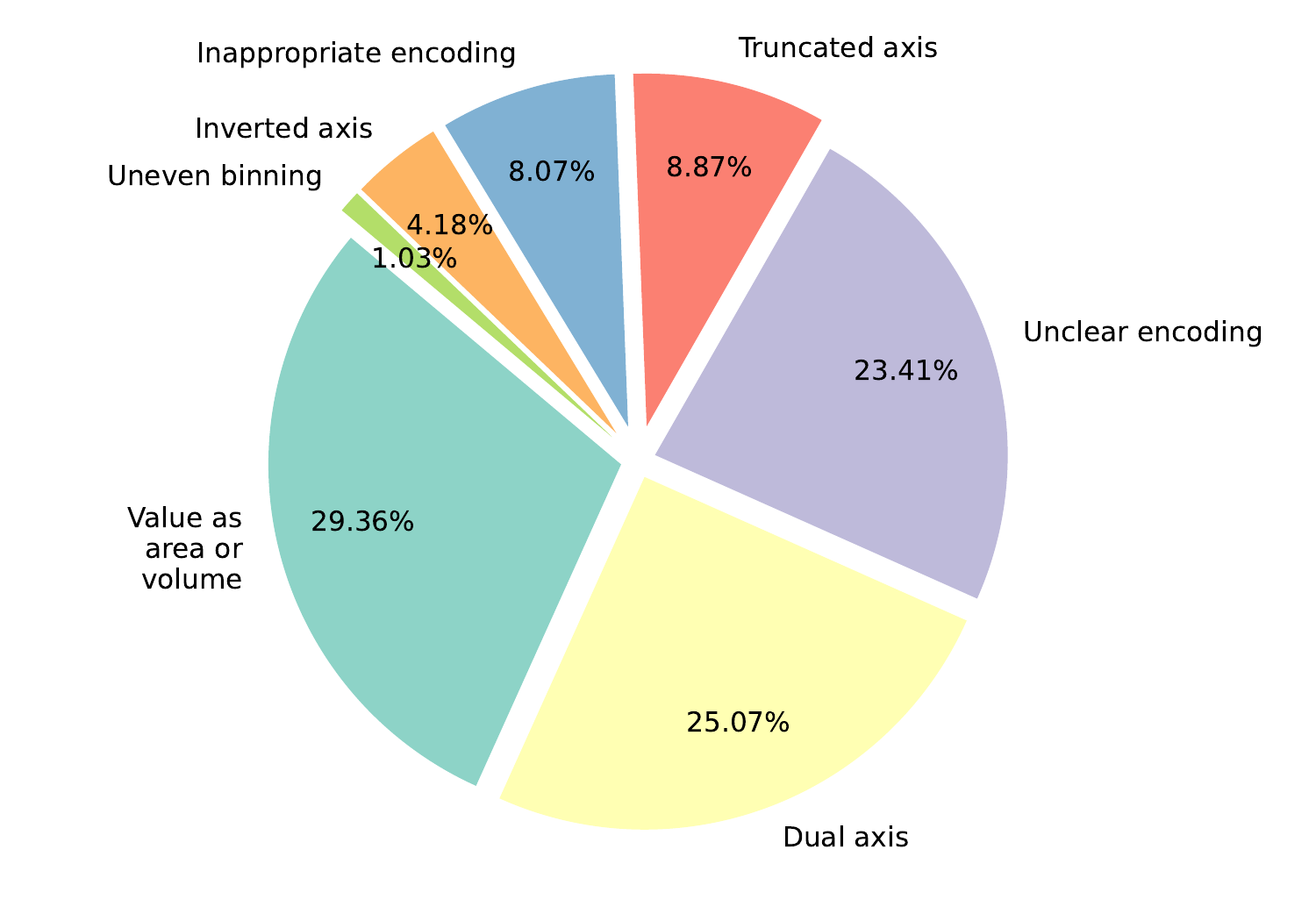}
        \vspace{-20pt}
        \caption{Visualization Error Composition for the dataset.}
        \label{fig:visualization_bar_chart}
        \vspace{20pt}
\end{figure*}

\newpage
\twocolumn

\subsection{Inference Configuration, Retry Policy, Total Number of Runs}
\label{app:inference_config}

We report the inference configurations used across all experiments for reproducibility and clarity.

\paragraph{Temperature.} For all models, we use the default temperature settings. 

\paragraph{Maximum Tokens.}
We set the maximum token limit to 10000 tokens for all models. This budget includes both visible output tokens and internal reasoning tokens (for applicable models; GPT family and Gemini family), ensuring that generations are not truncated during multi-label classification and justification generation.

\paragraph{Retry Policy and Malformed Output Handling.}
For each classification call, we allow up to 5 retry attempts if the model output does not conform to the required JSON schema or if the predicted labels do not exactly match the predefined set of valid error categories provided in the prompt. If all retry attempts fail, the sample is excluded from metric computation. Across all experiments, 3.7\% of samples were removed from evaluation due to invalid outputs. 

\paragraph{Number of Runs.}
Our analysis includes nine models evaluated on a dataset of 3{,}015 samples. For each sample, we make two independent model calls: one for reasoning error classification and one for visualization error classification. This results in a total of 54{,}270 (9$\times$3015$\times$2) model inference calls. \textit{Given typical academic resource constraints, this evaluation scale necessitates prioritization of experiments over exhaustive ablations.}

\subsection{Annotation and Verification}
\label{app:annotation_verification_metrics}

\textbf{Subset of our dataset taken from \citet{lisnic2023misleading}.} We reuse a subset of chart-caption pairs curated by \citet{lisnic2023misleading}. In their work, the authors collected tweets containing visualizations and manually filtered them to remove non-visualizations and unrelated content. They developed a codebook through an open-coding process and iteratively refined it through independent annotation and discussion among the authors, resulting in a taxonomy of visualization design violations and caption-level reasoning errors. The finalized codebook was then applied to annotate the full dataset. To verify these inherited annotations, we randomly sampled 50 instances and had two authors independently review the presence of the reported visualization and reasoning errors. Only two disagreements were observed and were resolved through discussion, confirming consistency with the original annotations.

\noindent \textbf{\emph{r/DataIsBeautiful}.} We randomly sampled 100 instances out of the 611 posts from \emph{r/DataIsBeautiful} used in our dataset for verification. Two authors independently reviewed these samples to verify the absence of visualization and reasoning errors. Disagreements were observed in 6 cases and were resolved through discussion to ensure a consistent interpretation of the original labels.

\noindent \textbf{\textit{Misleading Caption, Misleading Viz}.} For cases where both the visualization and caption are misleading (\textit{N=501}), we reuse charts exhibiting visualization design errors from \citet{lisnic2023misleading} and author new captions that introduce specific reasoning errors. Three authors independently wrote the new captions and annotated the reasoning errors present in the captions. To assess consistency, we randomly sampled 50 such instances and had the same three authors independently review the reasoning-error annotations. Disagreements were observed, yielding a Krippendorff's $\alpha$ of 0.84. These disagreements were subsequently discussed jointly and resolved through group discussion.

\subsection{Dataset statistics}
\begin{table}[!h]
\centering
\renewcommand{\arraystretch}{1.2}
\resizebox{0.5\textwidth}{!}{
\begin{tabular}{c cc cccccc}
\toprule
\multirow{1}{*}{\textbf{\# Errors (x)}} 
& \multirow{1}{*}{\textbf{Reasoning}} 
& \multirow{1}{*}{\textbf{Visualization}} 
& \multicolumn{6}{c}{\textbf{Reasoning + Visualization}} \\
\multirow{1}{*}{} 
& \multirow{1}{*}{\textbf{only}} 
& \multirow{1}{*}{\textbf{only}} 
& \multicolumn{6}{c}{\textbf{$(y,z)$}} \\
\cmidrule(lr){4-9}
& & 
& (1,1) 
& (1,2) 
& (2,1) 
& (2,2) 
& (3,1)
& (4,1) \\
\midrule
1 & 476 & 993 & -- & -- & -- & -- & -- & -- \\
2 & 292 & 115 & 344 & -- & -- & -- & -- & -- \\
3 & 24  & 2   & -- & 14 & 106 & -- & -- & -- \\
4 & 1   & 0   & -- & -- & -- & 3  & 32 & -- \\
5 & 0   & 0   & -- & -- & -- & -- & -- & 2  \\
\bottomrule
\end{tabular}
}
\caption{Distribution of samples by total number of errors ($x$). \textbf{Reasoning-only} and \textbf{Visualization-only} denote samples containing exclusively reasoning and visualization errors respectively. \textbf{Reasoning + Visualization} reports compositional error cases with $(y,z)$ indicating the number of reasoning and visualization errors, respectively, where $y+z=x$.}
\label{tab:error_count_distribution}
\vspace{-13pt}
\end{table}

\subsection{Image of our Annotation Interface}

\begin{figure*}[!t]
    \centering
        \centering
        \includegraphics[scale=0.4]{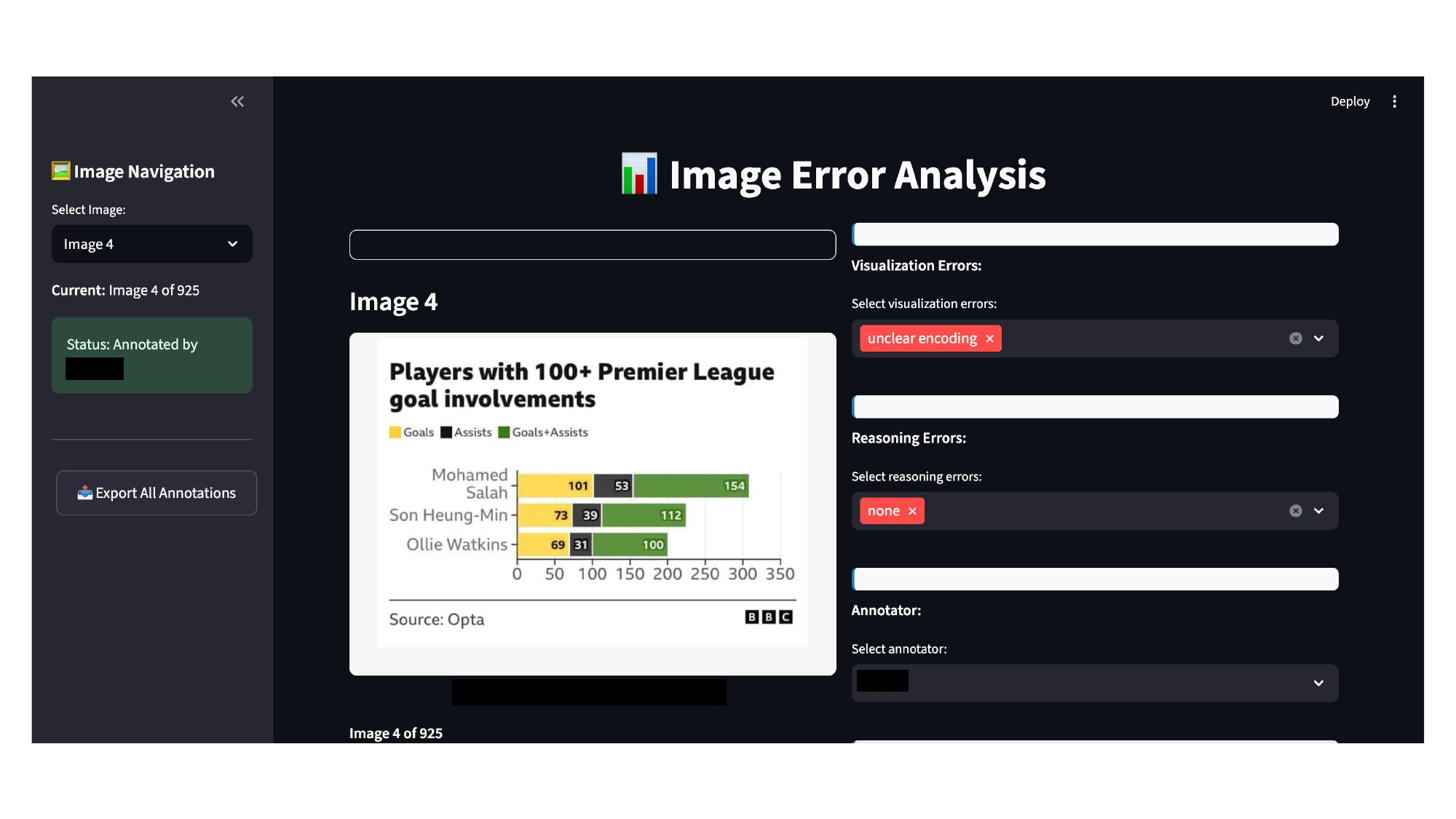}
        \vspace{-20pt}
        \caption{Example Picture of our Annotation Interface.}
        \label{fig:ui_picture_example}
        \vspace{-12pt}
\end{figure*}

Figure \ref{fig:ui_picture_example} shows a screenshot of our annotation interface used by annotators to label visualization and reasoning errors in charts and captions.

\subsection{Stability Across Multiple Runs}
\label{app:stability_multiple_runs}

Since VLMs use stochastic decoding during inference, model outputs can vary slightly across runs even when evaluated on the same dataset. To measure the stability of our results, we run \geminitwofivep \footnote{Note: \geminithreezerop was deprecated on March 26, 2026, and \geminithreeonep was newly introduced, with rate limits on our accounts.} three times with identical settings and report the mean and standard deviation of the evaluation metrics (F1, Partial Match, and Exact Match). Table~\ref{tab:gemini25pro_variance} summarizes the results across runs.

Overall, we observe that the standard deviations are relatively small across the reported metrics, indicating that the model's performance is stable across repeated evaluations. This suggests that the results reported in the main paper are not driven by a single favorable run but instead reflect consistent behavior of the model under stochastic decoding.

\begin{tcolorbox}[title=Reasoning Error Classification Prompt,
                  colback=gray!5,
                  colframe=black,
                  boxrule=0.5pt,
                  arc=2pt]
\small
You will be provided with a visualization, its accompanying caption,
and descriptions of reasoning errors. These reasoning errors represent
ways in which people use captions to spread misinformation. \\

Your task is to carefully examine the image and its accompanying caption.
Then, based on the information and the descriptions of reasoning errors,
you need to identify which of the following error categories, if any, apply to this chart-caption pair. \\

If none of the reasoning errors apply, classify the reasoning error as
"None." \\

Please classify which reasoning errors are present and explain your
reasoning. If more than one classification applies, include all applicable
classifications in a list. Even if only one classification applies, the
"classification" field must still be a list. \\

Only provide output in the following JSON format: \\

\{ \\

"reason": "[Explanation]", \\

"classification": ["Cherry-picking/Causal inference/Setting an arbitrary
threshold/Failure to account for statistical nuance/Incorrect reading of
chart/Issues with data validity/Misrepresentation of scientific studies/None"] \\

\} \\

Image: \{image\} \\

Accompanying Text: \{caption\} \\

Error Descriptions: \{reasoning\_error\_descriptions\}

\end{tcolorbox}

\begin{table}[!h]
\centering
\resizebox{\columnwidth}{!}{
\begin{tabular}{p{4cm}ccc}
\toprule
& F1 & PM & EM \\ 
\midrule
Combined & 0.59 ($\pm$ 0.04) & 0.84 ($\pm$ 0.03) & 0.06 ($\pm$ 0.01) \\
Reasoning Errors & 0.56 ($\pm$ 0.05) & 0.58 ($\pm$ 0.02) & 0.25 ($\pm$ 0.01) \\
Visualization Errors & 0.62 ($\pm$ 0.04) & 0.58 ($\pm$ 0.05) & 0.24 ($\pm$ 0.03) \\
\midrule
Cherry Picking & 0.57 ($\pm$ 0.08) &  &  \\
Causal Inference & 0.67 ($\pm$ 0.01) &  &  \\
Arbitrary Threshold & 0.62 ($\pm$ 0.06) &  &  \\
Failure to Account for Statistical Nuances & 0.12 ($\pm$ 0.00) & \textbf{N/A} & \textbf{N/A} \\
Incorrect Reading of the Chart & 0.10 ($\pm$ 0.03) &  &  \\
Data Validity Issues & 0.04 ($\pm$ 0.04) &  &  \\
Misrepresentation of Scientific Studies & 0.15 ($\pm$ 0.10) &  &  \\
\midrule
Truncated Axis & 0.66 ($\pm$ 0.04) &  &  \\
Dual Axis & 0.92 ($\pm$ 0.04) &  &  \\
Value as Area/Vol. & 0.66 ($\pm$ 0.04) &  &  \\
Inverted Axis & 0.44 ($\pm$ 0.01) & \textbf{N/A} & \textbf{N/A} \\
Uneven Binning & 0.13 ($\pm$ 0.03) &  &  \\
Unclear Encoding & 0.33 ($\pm$ 0.03) &  &  \\
Inappropriate Encoding & 0.16 ($\pm$ 0.01) &  &  \\
\bottomrule
\end{tabular}
}
\caption{Performance of \geminitwofivep across three independent runs. Each value corresponds to the mean and standard deviation ($\pm$).}
\vspace{-13pt}
\label{tab:gemini25pro_variance}
\end{table}

\subsection{Prompt Ablation}
\label{app:prompt_ablation}
\begin{tcolorbox}[title=Visualization Error Classification Prompt,
                  colback=gray!5,
                  colframe=black,
                  boxrule=0.5pt,
                  arc=2pt]

\small
You will be provided with a visualization, its accompanying caption, and descriptions of the visualization error. These visualization errors represent ways in which people use visualization to spread misinformation. \\

Your task is to carefully examine the image and its accompanying caption. Then, based on the information and the descriptions of visualization errors, you need to identify which of the following error categories, if any, apply to this chart-caption pair. \\

If none of the visualization errors apply, you may classify the visualization error as "None." \\

Please classify which visualization errors are present and explain your reasoning. If more than one classification applies, include all applicable classifications in a list. Even if there is only one classification, the "classification" field must still be a list. \\

Only provide output in the following JSON format: \\

\{ \\

"reason": "[Explanation]", \\

"classification": ["Truncated axis/Dual axis/Value as area or volume/Inverted axis/Uneven binning/Unclear encoding/Inappropriate encoding/None"] \\

\} \\

Image: \{image\} \\

Accompanying Text: \{caption\} \\

Error Descriptions: \{visualizaton\_error\_descriptions\}

\end{tcolorbox}

Prompt formulation can significantly influence the behavior of VLMs \cite{son2025advancing}. To assess the sensitivity of our results to prompt design, we conduct a prompt ablation using an alternative prompt variant. Specifically, we modify the task instructions to use more neutral wording and to slightly rephrase the classification objective. This change is intended to test whether the model's behavior is sensitive to minor variations in prompt phrasing and to evaluate the robustness of our results to alternative prompt formulations.

The results obtained for \geminitwofivep (Table \ref{tab:prompt_ablation_results}) with this prompt variant closely match the performance observed with the prompt used in our main experiments (Table \ref{tab:gemini25pro_variance}). This indicates that the overall results are largely insensitive to small changes in prompt wording, suggesting that the model's performance on our benchmark is robust to modest variations in prompt formulation.

\begin{table}[!h]
\centering
\renewcommand{\arraystretch}{1}
\resizebox{0.5\textwidth}{!}{
\begin{tabular}{lccc}
\toprule
& F1 & PM & EM \\ 
\midrule
Combined & 0.61 & 0.86 & 0.08 \\
Reasoning Errors & 0.59 & 0.61 & 0.29 \\
Visualization Errors & 0.64 & 0.59 & 0.24 \\
\midrule
Cherry Picking & 0.60 &  &  \\
Causal Inference & 0.68 &  &  \\
Setting an Arbitrary Threshold & 0.67 &  &  \\
Failure to Account for Statistical Nuances & 0.13 & \textbf{N/A} & \textbf{N/A} \\
Incorrect Reading of the Chart & 0.11 &  &  \\
Data Validity Issues & 0.01 &  &  \\
Misrepresentation of Scientific Studies & 0.09 &  &  \\
\midrule
Truncated Axis & 0.68 &  &  \\
Dual Axis & 0.94 &  &  \\
Value as Area/Vol. & 0.65 &  &  \\
Inverted Axis & 0.41 & \textbf{N/A} & \textbf{N/A} \\
Uneven Binning & 0.15 &  &  \\
Unclear Encoding  & 0.34 &  &  \\
Inappropriate Encoding & 0.16 &  &  \\
\bottomrule
\end{tabular}
}
\caption{Ablation results for the prompt variant. Results are reported from a single run with \geminitwofivep}
\vspace{-13pt}
\label{tab:prompt_ablation_results}
\end{table}

\subsection{Macro F1 scores}

We report the combined Macro F1 scores along with Macro F1 for (i) reasoning error classification, (ii) visualization error classification on the whole benchmark (Table \ref{tab:macro_f1_whole_benchmark}).

Because error categories are unevenly distributed and differ in intrinsic difficulty, this score provides a class-balanced perspective on performance. Unlike accuracy or weighted metrics, it shows whether models perform consistently across error types or disproportionately succeed on frequent or perceptually salient categories while failing on rarer ones. This metric, therefore, serves as a diagnostic indicator of robustness across the full error taxonomy.

\begin{table}[!h]
\centering
\resizebox{\columnwidth}{!}{
\begin{tabular}{lccc}
\toprule
Model & Combined & Reasoning Errors & Visualization Errors \\ 
\midrule
\geminithreeonep & 0.41 & 0.34 & 0.48 \\
\geminithreezerop & 0.42 & 0.34 & 0.50 \\
\geminitwofivep & 0.38 & 0.32 & 0.45 \\
\geminitwofivef          & 0.37 & 0.29 & 0.45 \\
\gptfive          & 0.38 & 0.33 & 0.42 \\
\gptfivemini       & 0.38 & 0.29  & 0.46 \\
\qwenthree        & 0.30 & 0.28 & 0.32 \\
\qwentwofive       & 0.18 & 0.16 & 0.19 \\
\qwentwofivechartqa & 0.14 & 0.12 & 0.17 \\
\bottomrule
\end{tabular}
}
\caption{Macro F1 scores of VLMs for reasoning and visualization errors separately and combined. Models consistently achieve higher scores on visualization error classification than reasoning errors, suggesting greater difficulty in identifying and reasoning about misinformation embedded in captions.}
\vspace{-14pt}
\label{tab:macro_f1_whole_benchmark}
\end{table}

\subsection{Grid-wise Results}
\label{app:grid-wise_results}

\begin{table}[!h]
\centering
\resizebox{1\linewidth}{!}{
\renewcommand{\arraystretch}{1}
\begin{tabular}{lcccc:cccc}
\toprule
\multicolumn{1}{c}{} & \multicolumn{4}{c:}{\textbf{$\triangle$}} & \multicolumn{4}{c}{\textbf{$\bigcirc$}} \\
\cmidrule(lr){2-5} \cmidrule(lr){6-9}
\textbf{Model} & \textbf{F1$_m$} & \textbf{F1$_w$} & \textbf{PM} & \textbf{EM}  & \textbf{F1$_m$} & \textbf{F1$_w$} & \textbf{PM} & \textbf{EM} \\
\midrule

\geminithreeonep  & 0.15 & 0.47 & 0.73 & 0.06 & 0.30 & 0.68 & 0.88 & 0.27 \\

\geminithreezerop & 0.16 & 0.50 & 0.79 & 0.08 & 0.30 & 0.69 & 0.91 & 0.28 \\

\geminitwofivep & 0.18 & 0.54 & 0.75 & 0.02 & 0.28 & 0.67 & 0.90 & 0.07 \\
\geminitwofivef & 0.16 & 0.49 & 0.75 & 0.00 & 0.28 & 0.68 & 0.84 & 0.02 \\
\gptfive & 0.17 & 0.53 & 0.80 & 0.01 & 0.26 & 0.67 & 0.90 & 0.16 \\
\gptfivemini & 0.16 & 0.50 & 0.81 & 0.00 & 0.28 & 0.68 & 0.88 & 0.06 \\
\qwenthree & 0.15 & 0.49 & 0.62 & 0.01 & 0.22 & 0.58 & 0.87 & 0.03 \\
\qwentwofive & 0.08 & 0.22 & 0.75 & 0.00 & 0.11 & 0.28 & 0.81 & 0.04 \\
\qwentwofivechartqa & 0.05 & 0.17 & 0.69 & 0.01 & 0.10 & 0.28 & 0.75 & 0.08 \\
\hline
\multicolumn{1}{c}{} & \multicolumn{4}{c:}{\textbf{$\blacksquare$}} & \multicolumn{4}{c}{\textbf{$\varnothing$}} \\
\cmidrule(lr){2-5} \cmidrule(lr){6-9}
& \textbf{F1$_m$} & \textbf{F1$_w$} & \textbf{PM} & \textbf{EM} & \textbf{F1$_m$} & \textbf{F1$_w$} & \textbf{PM} & \textbf{EM} \\
\midrule

\geminithreeonep & 0.45 & 0.71 & 0.93 & 0.14 & \multirow{9}{*}{\textbf{N/A}} & \multirow{9}{*}{\textbf{N/A}} & 0.87 & 0.34 \\

\geminithreezerop & 0.45 & 0.74 & 0.95 & 0.10 &  &  & 0.91 & 0.47 \\

\geminitwofivep & 0.43 & 0.74 & 0.98 & 0.01 &  & & 0.57 & 0.09 \\

\geminitwofivef & 0.44 & 0.74 & 0.98 & 0.02 &  & & 0.37 & 0.04 \\
\gptfive & 0.42 & 0.74 & 0.98 & 0.00 &  & & 0.75 & 0.30 \\
\gptfivemini & 0.42 & 0.74 & 0.98 & 0.01 & &  & 0.60 & 0.16 \\
\qwenthree & 0.34 & 0.67 & 0.99 & 0.00 & & & 0.65 & 0.08 \\
\qwentwofive & 0.22 & 0.42 & 0.75 & 0.01 &  & & 0.96 & 0.73 \\
\qwentwofivechartqa & 0.21 & 0.41 & 0.74 & 0.01 & & & 0.93 & 0.62 \\
\bottomrule
\end{tabular}
}
\caption{Performance of VLMs across the 2$\times$2 grid. We report combined macro F1 (F1$_m$), combined weighted F1 (F1$_w$), combined Partial Match (PM), and combined Exact Match (EM) for each model. Across the grids, models perform poorly on this task, with F1 scores staying below 0.75, highlighting the difficulty of reliably identifying and classifying multimodal misinformation. For the \textit{Non-Misleading Caption, Non-Misleading Viz} cell, we omit F1 scores as they are not applicable since there are no misinformation errors present, there are no true positives to evaluate per-error F1, and hence all per-error F1 scores are trivially zero and hence marked as \textbf{N/A}.}
\label{tab:2x2_grid_overall_combined}
\vspace{-13pt}
\end{table}

To better understand model behavior under different misleadingness conditions, we present a detailed breakdown across the four cells of the 2$\times$2 misleadingness grid.

We first report combined performance within each cell (Table \ref{tab:2x2_grid_overall_combined}), followed by disaggregated results for reasoning and visualization error classification (Table \ref{tab:overall_reasoning_visual_2x2_grid}). This grid-wise analysis clarifies whether performance varies systematically depending on whether misleadingness arises from the caption, the visualization, both, or neither.

To further characterize error-specific behavior, we report per-error F1 scores for each reasoning error type (Table \ref{tab:per_reasoning_error_f1_2x2}) and visualization error type (Table \ref{tab:per_viz_error_f1_2x2}) within each grid cell. These fine-grained results expose asymmetries in how models handle distinct categories of reasoning and visual distortions.

\paragraph{Precision, Recall, and False Positive Rates.}
Beyond F1, we provide per-error Precision and Recall scores across the full benchmark (Table \ref{tab:prec_recall_all_errors}) and across the $2\times2$ grid (Tables \ref{tab:reasoning_pr_rec_2x2}, \ref{tab:viz_pr_rec_2x2}) to disentangle over-prediction from under-detection. Given the deployment relevance of over-flagging benign content, we additionally report per-error False Positive Rates (FPR) on the whole benchmark (Table \ref{tab:per_error_fpr_whole_benchmark}) and within each grid cell (Tables \ref{tab:per_reasoning_error_fpr_2x2}, \ref{tab:per_visualization_error_fpr_2x2}). These metrics quantify calibration tendencies and help identify whether models systematically default toward predicting misleadingness in the absence of clear evidence.

\begin{table*}[!h]
\centering
\resizebox{\linewidth}{!}{
\renewcommand{\arraystretch}{1}
\begin{tabular}{lcccc:cccc:cccc:cccc}
\toprule
\multicolumn{1}{c}{} & \multicolumn{8}{c:}{\textbf{$\triangle$}} & \multicolumn{8}{c}{\textbf{$\bigcirc$}} \\
\cmidrule(lr){2-9} \cmidrule(lr){10-17}
\multicolumn{1}{c}{} & \multicolumn{4}{c:}{Reasoning Errors} & \multicolumn{4}{c:}{Visualization Errors} & \multicolumn{4}{c:}{Reasoning Errors} & \multicolumn{4}{c}{Visualization Errors} \\

\textbf{Model} & F1$_m$ & F1$_w$ & PM & EM & F1$_m$ & F1$_w$ & PM & EM & F1$_m$ & F1$_w$ & PM & EM & F1$_m$ & F1$_w$ & PM & EM \\
\midrule

\geminithreeonep & 0.31 & 0.47 & 0.53 & 0.13 & \multirow{9}{*}{\textbf{N/A}} & \multirow{9}{*}{\textbf{N/A}} & 0.47 & 0.47 & \multirow{9}{*}{\textbf{N/A}} & \multirow{9}{*}{\textbf{N/A}} & 0.55 & 0.55 & 0.60 & 0.68 & 0.75 & 0.48 \\

\geminithreezerop & 0.32 & 0.50 & 0.54 & 0.11 &  & & 0.61 & 0.61 &  &  & 0.61 & 0.61 & 0.59 & 0.69 & 0.76 & 0.43 \\
\geminitwofivep & 0.36 & 0.54 & 0.71 & 0.06 &  & & 0.20 & 0.20 &  & & 0.30 & 0.30 & 0.57 & 0.67 & 0.84 & 0.25 \\
\geminitwofivef & 0.32 & 0.49 & 0.68 & 0.02 &  & & 0.30 & 0.30 &  & & 0.07 & 0.07 & 0.57 & 0.68 & 0.82 & 0.28 \\
\gptfive & 0.33 & 0.53 & 0.66 & 0.01 &  & & 0.52 & 0.52 &  & & 0.47 & 0.47 & 0.53 & 0.67 & 0.81 & 0.31 \\
\gptfivemini & 0.33 & 0.50 & 0.68 & 0.00 &  & & 0.51 & 0.51 &  & & 0.20 & 0.20 & 0.57 & 0.68 & 0.85 & 0.30 \\
\qwenthree & 0.31 & 0.49 & 0.60 & 0.12 &  & & 0.04 & 0.04 &  & & 0.40 & 0.40 & 0.44 & 0.58 & 0.75 & 0.11 \\
\qwentwofive & 0.16 & 0.22 & 0.28 & 0.01 &  & & 0.64 & 0.64 &  & & 0.73 & 0.73 &0.22  & 0.28 & 0.26 & 0.05 \\
\qwentwofivechartqa & 0.10 & 0.17 & 0.24 & 0.02 &  & & 0.58 & 0.58 &  & & 0.65 & 0.65 & 0.20 & 0.28 & 0.27 & 0.12 \\

\midrule
\multicolumn{1}{c}{} & \multicolumn{8}{c:}{\textbf{$\blacksquare$}} & \multicolumn{8}{c}{\textbf{$\varnothing$}} \\
\cmidrule(lr){2-9} \cmidrule(lr){10-17}
\multicolumn{1}{c}{} & \multicolumn{4}{c:}{Reasoning Errors} & \multicolumn{4}{c:}{Visualization Errors} & \multicolumn{4}{c:}{Reasoning Errors} & \multicolumn{4}{c}{Visualization Errors} \\
 & F1$_m$ & F1$_w$ & PM & EM & F1$_m$ & F1$_w$ & PM & EM & F1$_m$ & F1$_w$  & PM & EM & F1$_m$ & F1$_w$  & PM & EM \\
\midrule

\geminithreeonep & 0.41 & 0.66 & 0.78 & 0.26 & 0.48 & 0.78 & 0.73 & 0.47  & \multirow{9}{*}{\textbf{N/A}} & \multirow{9}{*}{\textbf{N/A}} & 0.80 & 0.80 & \multirow{9}{*}{\textbf{N/A}} & \multirow{9}{*}{\textbf{N/A}} & 0.41 & 0.41 \\

\geminithreezerop & 0.39 & 0.66 & 0.80 & 0.21 & 0.51 & 0.84 & 0.82 & 0.48 & & & 0.83 & 0.83 &  &  & 0.55 & 0.55\\

\geminitwofivep & 0.39 & 0.68 & 0.91 & 0.08 & 0.48 & 0.81 & 0.79 & 0.19 &  & & 0.52 & 0.52 & &   & 0.15 & 0.15 \\
\geminitwofivef & 0.39 & 0.67 & 0.94 & 0.03 & 0.49 & 0.82 & 0.81 & 0.29 & & & 0.19 & 0.19 &  & & 0.22 & 0.22 \\
\gptfive & 0.43 & 0.68 & 0.95 & 0.03 & 0.41 & 0.81 & 0.79 & 0.31 &  & & 0.62 & 0.62 &  & & 0.42 & 0.42 \\
\gptfivemini & 0.36 & 0.67 & 0.94 & 0.01 & 0.48 & 0.84 & 0.83 & 0.30 &  & & 0.35 & 0.35 & &  & 0.42 & 0.42 \\
\qwenthree & 0.37 & 0.68 & 0.97 & 0.05 & 0.30 & 0.67 & 0.67 & 0.05 &  & & 0.62 & 0.62 &  & & 0.11 & 0.11 \\
\qwentwofive & 0.21 & 0.41 & 0.56 & 0.13 & 0.24 & 0.44 & 0.42 & 0.15 &  & & 0.87 & 0.87 & &  & 0.82 & 0.82 \\
\qwentwofivechartqa & 0.21 & 0.41 & 0.57 & 0.11 & 0.20 & 0.42 & 0.40 & 0.17 & &  & 0.77 & 0.77 &  & & 0.77 & 0.77 \\
\bottomrule
\end{tabular}
}
\caption{Performance breakdown across both reasoning and visualization errors, reported for the 2$\times$2 grid. Each section shows macro F1 (F1$_m$), weighted F1 (F1$_w$), Partial Match (PM), and Exact Match (EM) scores separately for reasoning and visualization classification tasks. In most cases, models achieve lower F1 scores on reasoning error classification than on visualization error classification, indicating that reasoning-based deception is more challenging for current VLMs to detect. F1 scores are marked \textbf{N/A} in grid cells where the relevant modality contains no positive instances due to the absence of the corresponding error types.}

\label{tab:overall_reasoning_visual_2x2_grid}
\end{table*}

\begin{table*}[!t]
\centering
\resizebox{\linewidth}{!}{
\renewcommand{\arraystretch}{1}
\begin{tabular}{lccccccc:ccccccc}
\toprule
\multicolumn{1}{c}{} & \multicolumn{7}{c:}{\textbf{$\triangle$}} & \multicolumn{7}{c}{\textbf{$\bigcirc$}} \\
\cmidrule(lr){2-8} \cmidrule(lr){9-15}
\textbf{Model}
& \shortstack{Cher.\\Pick}
 & \shortstack{Caus.\\Infer.}
 & \shortstack{Arb.\\Thr.}
 & \shortstack{Stat.\\Nu.} 
 & \shortstack{Chart\\Read}
 & \shortstack{Data\\Val.} 
 & \shortstack{Mis.\\Sci.}
& \shortstack{Cher.\\Pick}
 & \shortstack{Caus.\\Infer.}
 & \shortstack{Arb.\\Thr.}
 & \shortstack{Stat.\\Nu.} 
 & \shortstack{Chart\\Read}
 & \shortstack{Data\\Val.} 
 & \shortstack{Mis.\\Sci.} \\
\midrule

\geminithreeonep & 0.36 & 0.83 & 0.14 & 0.11 & 0.08 & 0.34 & 0.33 &  &  &  & \multirow{9}{*}{\textbf{N/A}} &  &  &  \\

\geminithreezerop & 0.48 & 0.80 & 0.16 & 0.08 & 0.00 & 0.37 & 0.38 &  &  &  &  & & & \\ 

\geminitwofivep & 0.49 & 0.85 & 0.27 & 0.17 & 0.06 & 0.25 & 0.44 &  &  &  &  &  &  &  \\
\geminitwofivef & 0.44 & 0.81 & 0.18 & 0.16 & 0.04 & 0.23 & 0.39 &  &  &  &  &  &  &  \\
\gptfive & 0.53 & 0.83 & 0.26 & 0.16 & 0.04 & 0.22 & 0.30 &  &  &  &  &  &  &  \\
\gptfivemini & 0.51 & 0.80 & 0.16 & 0.15 & 0.03 & 0.25 & 0.40 &  &  &  &  &  &  &  \\
\qwenthree & 0.51 & 0.73 & 0.24 & 0.36 & 0.03 & 0.06 & 0.22 &  &  &  &  &  &  &  \\
\qwentwofive & 0.47 & 0.13 & 0.14 & 0.18 & 0.03 & 0.08 & 0.06 &  &  &  &  &  &  &  \\
\qwentwofivechartqa & 0.45 & 0.09 & 0.08 & 0.02 & 0.00 & 0.00 & 0.06 &  &  &  &  &  &  &  \\
\midrule
\multicolumn{1}{c}{} & \multicolumn{7}{c:}{\textbf{$\blacksquare$}} & \multicolumn{7}{c}{\textbf{$\varnothing$}} \\
\cmidrule(lr){2-8} \cmidrule(lr){9-15}
& \shortstack{Cher.\\Pick}
 & \shortstack{Caus.\\Infer.}
 & \shortstack{Arb.\\Thr.}
 & \shortstack{Stat.\\Nu.} 
 & \shortstack{Chart\\Read}
 & \shortstack{Data\\Val.} 
 & \shortstack{Mis.\\Sci.}
 & \shortstack{Cher.\\Pick}
 & \shortstack{Caus.\\Infer.}
 & \shortstack{Arb.\\Thr.}
 & \shortstack{Stat.\\Nu.} 
 & \shortstack{Chart\\Read}
 & \shortstack{Data\\Val.} 
 & \shortstack{Mis.\\Sci.} \\
\midrule

\geminithreeonep & 0.59 & 0.74 & 0.90 & 0.26 & 0.15 & 0.25 & 0.00 &  &  &  & \multirow{9}{*}{\textbf{N/A}} & & & \\ 
 
\geminithreezerop & 0.64 & 0.71 & 0.86 & 0.28 & 0.09 & 0.14 & 0.00 &  &  &  &  &  &  &  \\

\geminitwofivep & 0.71 & 0.63 & 0.89 & 0.22 & 0.19 & 0.07 & 0.00 &  &  &  &  &  &  &  \\
\geminitwofivef & 0.72 & 0.53 & 0.91 & 0.23 & 0.13 & 0.09 & 0.14 &  &  &  &  &  &  &  \\
\gptfive & 0.72 & 0.62 & 0.87 & 0.22 & 0.12 & 0.33 & 0.00 &  &  &  &  &  &  &  \\
\gptfivemini & 0.72 & 0.52 & 0.92 & 0.22 & 0.12 & 0.07 & 0.00 &  &  &  &  &  &  &  \\
\qwenthree & 0.68 & 0.67 & 0.90 & 0.22 & 0.10 & 0.00 & 0.00 &  &  &  &  &  &  &  \\
\qwentwofive & 0.65 & 0.38 & 0.30 & 0.08 & 0.03 & 0.00 & 0.04 &  &  &  &  &  &  &  \\
\qwentwofivechartqa & 0.65 & 0.40 & 0.28 & 0.04 & 0.08 & 0.00 & 0.05 &  &  &  &  &  &  &  \\
\bottomrule
\end{tabular}
}
\caption{Per-error F1 scores for reasoning error classification across the 2$\times$2 grid. VLMs perform relatively well on certain reasoning error types such as \textit{Cherry-picking}, \textit{Setting an arbitrary threshold}, and \textit{Causal Inference}, but struggle on others. Cells corresponding to \textit{Non-Misleading Caption, Misleading Viz} and \textit{Non-Misleading Caption, Non-Misleading Viz} do not contain any reasoning errors, so per-error F1 scores are marked as \textbf{N/A}.}
\label{tab:per_reasoning_error_f1_2x2}
\end{table*}

\clearpage
\newpage

\begin{table*}[!h]
\centering
\resizebox{\linewidth}{!}{
\renewcommand{\arraystretch}{1}
\begin{tabular}{lccccccc:ccccccc}
\toprule
\multicolumn{1}{c}{} & \multicolumn{7}{c:}{\textbf{$\triangle$}} & \multicolumn{7}{c}{\textbf{$\bigcirc$}} \\
\cmidrule(lr){2-8} \cmidrule(lr){9-15}
\textbf{Model}
& \shortstack{Trunc.\\Axis}
 & \shortstack{Dual\\Axis}
 & \shortstack{Area/\\Vol.} 
 & \shortstack{Inv.\\Axis}
 & \shortstack{Uneven\\Bin.} 
 & \shortstack{Unclr.\\Enc.} 
 & \shortstack{Inappr.\\Enc.} 
& \shortstack{Trunc.\\Axis}
 & \shortstack{Dual\\Axis}
 & \shortstack{Area/\\Vol.} 
 & \shortstack{Inv.\\Axis}
 & \shortstack{Uneven\\Bin.} 
 & \shortstack{Unclr.\\Enc.} 
 & \shortstack{Inappr.\\Enc.}  \\
\midrule

\geminithreeonep &  &  &  & \multirow{9}{*}{\textbf{N/A}} &  &  &   0.68 & 0.80 & 0.98 & 0.70 & 0.57 & 0.20 & 0.59 & 0.38 \\

\geminithreezerop &  &  &  &  &  &  &  & 0.75 & 0.97 & 0.74 & 0.49 & 0.23 & 0.60 & 0.36 \\

\geminitwofivep &  &  &  &  &  &  &  & 0.71 & 0.97 & 0.75 & 0.47 & 0.20 & 0.58 & 0.32 \\
\geminitwofivef &  &  &  &  &  &  &  & 0.74 & 0.95 & 0.76 & 0.47 & 0.17 & 0.60 & 0.30 \\
\gptfive &  &  &  &  &  &  &  & 0.65 & 0.97 & 0.77 & 0.16 & 0.23 & 0.60 & 0.32 \\
\gptfivemini &  &  &  &  &  &  &  & 0.70 & 0.98 & 0.73 & 0.31 & 0.31 & 0.62 & 0.31 \\
\qwenthree &  &  &  &  &  &  &  & 0.44 & 0.94 & 0.62 & 0.15 & 0.17 & 0.52 & 0.25 \\
\qwentwofive &  &  &  &  &  &  &  & 0.27 & 0.80 & 0.11 & 0.07 & 0.00 & 0.18 & 0.10 \\
\qwentwofivechartqa &  &  &  &  &  &  &  & 0.22 & 0.77 & 0.17 & 0.02 & 0.00 & 0.15 & 0.11 \\
\midrule
\multicolumn{1}{c}{} & \multicolumn{7}{c:}{\textbf{$\blacksquare$}} & \multicolumn{7}{c}{\textbf{$\varnothing$}} \\
\cmidrule(lr){2-8} \cmidrule(lr){9-15}
& \shortstack{Trunc.\\Axis}
 & \shortstack{Dual\\Axis}
 & \shortstack{Area/\\Vol.} 
 & \shortstack{Inv.\\Axis}
 & \shortstack{Uneven\\Bin.} 
 & \shortstack{Unclr.\\Enc.} 
 & \shortstack{Inappr.\\Enc.} 
& \shortstack{Trunc.\\Axis}
 & \shortstack{Dual\\Axis}
 & \shortstack{Area/\\Vol.} 
 & \shortstack{Inv.\\Axis}
 & \shortstack{Uneven\\Bin.} 
 & \shortstack{Unclr.\\Enc.} 
 & \shortstack{Inappr.\\Enc.}  \\
\midrule

\geminithreeonep & 0.88 & 0.98 & 0.70 & 0.60 & 0.03 & 0.11 & 0.07 &  &  &  & \multirow{9}{*}{\textbf{N/A}} &  &  & \\

\geminithreezerop & 0.86 & 0.98 & 0.83 & 0.67 & 0.04 & 0.08 & 0.12 &  &  &  &  &  &  & \\
\geminitwofivep & 0.84 & 0.96 & 0.80 & 0.62 & 0.03 & 0.08 & 0.04 &  &  &  &  &  &  &  \\
\geminitwofivef & 0.83 & 0.95 & 0.83 & 0.65 & 0.02 & 0.09 & 0.05 &  &  &  &  &  &  &  \\
\gptfive & 0.69 & 0.97 & 0.86 & 0.19 & 0.03 & 0.08 & 0.06 &  &  &  &  &  &  &  \\
\gptfivemini & 0.78 & 0.98 & 0.86 & 0.55 & 0.04 & 0.08 & 0.03 &  &  &  &  &  &  &  \\
\qwenthree & 0.38 & 0.90 & 0.68 & 0.06 & 0.00 & 0.05 & 0.03 &  &  &  &  &  &  &  \\
\qwentwofive & 0.35 & 0.87 & 0.19 & 0.16 & 0.00 & 0.07 & 0.04 &  &  &  &  &  &  &  \\
\qwentwofivechartqa & 0.25 & 0.87 & 0.18 & 0.14 & 0.00 & 0.00 & 0.00 &  &  &  &  &  &  \\
\bottomrule
\end{tabular}
}
\caption{Per-error F1 scores for visualization error classification across the 2$\times$2 grid. VLMs perform relatively well on certain visualization error types such as \textit{Dual Axis}, \textit{Truncated Axis}, and \textit{Value as Area or Volume}, but struggle on others. Cells corresponding to \textit{Misleading Caption, Non-Misleading Viz} and \textit{Non-Misleading Caption, Non-Misleading Viz} do not contain any visualization errors, so per-error F1 scores are marked as \textbf{N/A}.}
\label{tab:per_viz_error_f1_2x2}
\end{table*}

\begin{table*}[!t]
\centering
\renewcommand{\arraystretch}{1}
\resizebox{\linewidth}{!}{
\begin{tabular}{l cc: cc: cc: cc: cc: cc: cc: cc: cc: cc: cc: cc: cc: cc}
\toprule
& \multicolumn{14}{c:}{\textbf{Reasoning Errors}} & \multicolumn{14}{c}{\textbf{Visualization Errors}} \\
\cmidrule(lr){2-15} \cmidrule(lr){16-29}
\textbf{Model} 
& \multicolumn{2}{c:}{\shortstack{Cher.\\Pick}} 
& \multicolumn{2}{c:}{\shortstack{Caus.\\Infer.}} 
& \multicolumn{2}{c:}{\shortstack{Arb.\\Thr.}} 
& \multicolumn{2}{c:}{\shortstack{Stat.\\Nu.}} 
& \multicolumn{2}{c:}{\shortstack{Chart\\Read}} 
& \multicolumn{2}{c:}{\shortstack{Data\\Val.}} 
& \multicolumn{2}{c:}{\shortstack{Mis.\\Sci.}} 

& \multicolumn{2}{c:}{\shortstack{Trunc.\\Axis}} 
& \multicolumn{2}{c:}{\shortstack{Dual\\Axis}} 
& \multicolumn{2}{c:}{\shortstack{Area/\\Vol.}} 
& \multicolumn{2}{c:}{\shortstack{Inv.\\Axis}}
& \multicolumn{2}{c:}{\shortstack{Uneven\\Bin.}} 
& \multicolumn{2}{c:}{\shortstack{Unclr.\\Enc.}} 
& \multicolumn{2}{c}{\shortstack{Inappr.\\Enc.}}  \\
\cmidrule(lr){2-29}
& P & R & P & R & P & R & P & R & P & R & P & R & P & R 
& P & R & P & R & P & R & P & R & P & R & P & R & P & R \\
\midrule

\geminithreeonep & 0.50 & 0.38 & 0.73 & 0.75 & 0.80 & 0.42 & 0.08 & 0.49 & 0.03 & 0.62 & 0.33 & 0.25 & 0.26 & 0.19 & 0.53 & 0.90 & 0.82 & 0.98 & 0.76 & 0.56 & 0.42 & 0.55 & 0.04 & 0.72 & 0.27 & 0.63 & 0.15 & 0.56 \\

\geminithreezerop & 0.44 & 0.54 & 0.69 & 0.73 & 0.73 & 0.40 & 0.06 & 0.50 & 0.02 & 0.29 & 0.25 & 0.28 & 0.17 & 0.45 & 0.54 & 0.85 & 0.81 & 0.99 & 0.74 & 0.68 & 0.48 & 0.50 & 0.07 & 0.67 & 0.29 & 0.68 & 0.15 & 0.59 \\

\geminitwofivep & 0.36 & 0.70 & 0.52 & 0.91 & 0.63 & 0.50 & 0.06 & 0.86 & 0.03 & 0.50 & 0.04 & 0.25 & 0.18 & 0.58 
        & 0.50 & 0.78 & 0.78 & 0.98 & 0.71 & 0.68 & 0.43 & 0.43 & 0.05 & 0.83 & 0.18 & 0.89 & 0.09 & 0.76 \\
\geminitwofivef & 0.31 & 0.68 & 0.44 & 0.90 & 0.57 & 0.46 & 0.05 & 0.99 & 0.02 & 0.42 & 0.04 & 0.25 & 0.19 & 0.50 
        & 0.44 & 0.86 & 0.79 & 0.95 & 0.62 & 0.77 & 0.43 & 0.45 & 0.04 & 0.83 & 0.22 & 0.77 & 0.09 & 0.68 \\
\gptfive & 0.39 & 0.80 & 0.56 & 0.85 & 0.64 & 0.50 & 0.07 & 0.98 & 0.02 & 0.62 & 0.10 & 0.18 & 0.17 & 0.31 
        & 0.49 & 0.62 & 0.82 & 0.97 & 0.70 & 0.77 & 0.23 & 0.11 & 0.06 & 0.78 & 0.23 & 0.83 & 0.12 & 0.63 \\
\gptfivemini & 0.35 & 0.80 & 0.44 & 0.90 & 0.54 & 0.46 & 0.05 & 0.98 & 0.02 & 0.69 & 0.04 & 0.32 & 0.20 & 0.38 
        & 0.51 & 0.72 & 0.81 & 0.98 & 0.56 & 0.82 & 0.48 & 0.29 & 0.09 & 0.89 & 0.23 & 0.88 & 0.12 & 0.58 \\
\qwenthree & 0.31 & 0.84 & 0.51 & 0.72 & 0.48 & 0.51 & 0.10 & 0.77 & 0.01 & 0.65 & 0.33 & 0.03 & 0.23 & 0.12 
        & 0.10 & 0.71 & 0.75 & 0.88 & 0.64 & 0.55 & 0.24 & 0.07 & 0.14 & 0.11 & 0.17 & 0.87 & 0.06 & 0.76 \\
\qwentwofive & 0.33 & 0.73 & 0.58 & 0.12 & 0.34 & 0.13 & 0.12 & 0.12 & 0.01 & 0.15 & 0.25 & 0.04 & 0.02 & 0.23 
        & 0.11 & 0.33 & 0.73 & 0.77 & 0.51 & 0.08 & 0.10 & 0.08 & 0.00 & 0.00 & 0.16 & 0.12 & 0.05 & 0.10 \\
\qwentwofivechartqa & 0.29 & 0.72 & 0.51 & 0.10 & 0.40 & 0.10 & 0.04 & 0.02 & 0.02 & 0.12 & 0.00 & 0.00 & 0.02 & 0.31 
        & 0.09 & 0.21 & 0.57 & 0.83 & 0.38 & 0.10 & 0.04 & 0.05 & 0.00 & 0.00 & 0.22 & 0.08 & 0.05 & 0.10 \\

\bottomrule
\end{tabular}
}
\caption{Per-error Precision (P) and Recall (R) scores for reasoning and visualization error classification across the whole benchmark. Models show relatively high recall but low precision for certain error types such as \textit{Failure to Account for Statistical Nuance} and \textit{Unclear Encoding}, indicating a tendency to over-predict these categories even when not applicable. This behavior highlights challenges in accurately distinguishing subtle or context-dependent misinformation patterns.}

\label{tab:prec_recall_all_errors}
\end{table*}

\begin{table*}[t]
\centering
\renewcommand{\arraystretch}{1.15}
\resizebox{\textwidth}{!}{%
\begin{tabular}{l c:c: c:c: c:c: c:c: c:c: c:c: c:c: cc cc cc cc cc cc cc}
\toprule
\multicolumn{1}{c}{} & \multicolumn{14}{c:}{\textbf{$\triangle$}} & \multicolumn{14}{c}{\textbf{$\bigcirc$}} \\
 \cmidrule(lr){2-15} \cmidrule(lr){16-29}

\textbf{Model}
& \multicolumn{2}{c}{\shortstack{Cher.\\Pick}} 
& \multicolumn{2}{c}{\shortstack{Caus.\\Infer.}} 
& \multicolumn{2}{c}{\shortstack{Arb.\\Thr.}} 
& \multicolumn{2}{c}{\shortstack{Stat.\\Nu.}} 
& \multicolumn{2}{c}{\shortstack{Chart\\Read}} 
& \multicolumn{2}{c}{\shortstack{Data\\Val.}} 
& \multicolumn{2}{c}{\shortstack{Mis.\\Sci.}} 
& \multicolumn{2}{c}{\shortstack{Cher.\\Pick}} 
& \multicolumn{2}{c}{\shortstack{Caus.\\Infer.}} 
& \multicolumn{2}{c}{\shortstack{Arb.\\Thr.}} 
& \multicolumn{2}{c}{\shortstack{Stat.\\Nu.}} 
& \multicolumn{2}{c}{\shortstack{Chart\\Read}} 
& \multicolumn{2}{c}{\shortstack{Data\\Val.}} 
& \multicolumn{2}{c}{\shortstack{Mis.\\Sci.}}  \\
\cline{2-29}
& P & R & P & R & P & R & P & R & P & R & P & R & P & R
& P & R & P & R & P & R & P & R & P & R & P & R & P & R \\
\hline

\geminithreeonep & 0.45 & 0.31 & 0.93 & 0.74 & 0.77 & 0.08 & 0.06 & 0.40 & 0.04 & 0.57 & 0.55 & 0.25 & 0.45 & 0.26 & \multicolumn{14}{c}{\textbf{}} \\

\geminithreezerop & 0.45 & 0.52 & 0.91 & 0.71 & 0.64 & 0.09 & 0.04 & 0.33 & 0.00 & 0.00 & 0.53 & 0.28 & 0.27 & 0.67 & \multicolumn{14}{c}{ } \\

\geminitwofivep
& 0.40 & 0.62 & 0.81 & 0.90 & 0.76 & 0.16 & 0.09 & 0.94 & 0.03 & 0.28 & 0.25 & 0.25 & 0.31 & 0.79
& \multicolumn{14}{c}{ } \\

\geminitwofivef
& 0.36 & 0.57 & 0.76 & 0.87 & 0.76 & 0.10 & 0.08 & 1.00 & 0.02 & 0.28 & 0.22 & 0.25 & 0.28 & 0.63
& \multicolumn{14}{c}{ } \\

\gptfive
& 0.40 & 0.75 & 0.82 & 0.83 & 0.74 & 0.16 & 0.08 & 0.95 & 0.02 & 0.43 & 0.34 & 0.16 & 0.26 & 0.37
& \multicolumn{14}{c}{\textbf{N/A}} \\

\gptfivemini
& 0.39 & 0.76 & 0.73 & 0.88 & 0.56 & 0.10 & 0.08 & 0.98 & 0.02 & 0.71 & 0.20 & 0.33 & 0.33 & 0.53
& \multicolumn{14}{c}{\textbf{}} \\

\qwenthree
& 0.38 & 0.74 & 0.82 & 0.66 & 0.45 & 0.17 & 0.22 & 0.88 & 0.02 & 1.00 & 0.50 & 0.03 & 0.37 & 0.16
& \multicolumn{14}{c}{ } \\

\qwentwofive
& 0.39 & 0.59 & 0.86 & 0.07 & 0.31 & 0.09 & 0.20 & 0.17 & 0.01 & 0.43 & 0.50 & 0.05 & 0.04 & 0.16
& \multicolumn{14}{c}{ } \\

\qwentwofivechartqa
& 0.37 & 0.59 & 0.66 & 0.05 & 0.50 & 0.04 & 0.04 & 0.02 & 0.00 & 0.00 & 0.00 & 0.00 & 0.03 & 0.21
& \multicolumn{14}{c}{ } \\

\hline

\multicolumn{1}{c}{} & \multicolumn{14}{c:}{\textbf{$\blacksquare$}} & \multicolumn{14}{c}{\textbf{$\varnothing$}} \\
 \cmidrule(lr){2-15} \cmidrule(lr){16-29}
 
& \multicolumn{2}{c}{\shortstack{Cher.\\Pick}} 
& \multicolumn{2}{c}{\shortstack{Caus.\\Infer.}} 
& \multicolumn{2}{c}{\shortstack{Arb.\\Thr.}} 
& \multicolumn{2}{c}{\shortstack{Stat.\\Nu.}} 
& \multicolumn{2}{c}{\shortstack{Chart\\Read}} 
& \multicolumn{2}{c}{\shortstack{Data\\Val.}} 
& \multicolumn{2}{c}{\shortstack{Mis.\\Sci.}} 
& \multicolumn{2}{c}{\shortstack{Cher.\\Pick}} 
& \multicolumn{2}{c}{\shortstack{Caus.\\Infer.}} 
& \multicolumn{2}{c}{\shortstack{Arb.\\Thr.}} 
& \multicolumn{2}{c}{\shortstack{Stat.\\Nu.}} 
& \multicolumn{2}{c}{\shortstack{Chart\\Read}} 
& \multicolumn{2}{c}{\shortstack{Data\\Val.}} 
& \multicolumn{2}{c}{\shortstack{Mis.\\Sci.}}  \\
\cline{2-29}
& P & R & P & R & P & R & P & R & P & R & P & R & P & R
& P & R & P & R & P & R & P & R & P & R & P & R & P & R \\
\hline

\geminithreeonep & 0.83 & 0.46 & 0.70 & 0.78 & 0.97 & 0.83 & 0.17 & 0.58 & 0.84 & 0.63 & 0.25 & 0.25 & \multicolumn{14}{c}{ } \\

\geminithreezerop & 0.74 & 0.56 & 0.64 & 0.79 & 0.95 & 0.79 & 0.18 & 0.66 & 0.05 & 0.37 & 0.10 & 0.25 & 0.00 & 0.00 & \multicolumn{14}{c}{ } \\

\geminitwofivep
& 0.64 & 0.79 & 0.47 & 0.93 & 0.86 & 0.91 & 0.13 & 0.77 & 0.11 & 0.58 & 0.04 & 0.25 & 0.00 & 0.00
& \multicolumn{14}{c}{ } \\

\geminitwofivef
& 0.64 & 0.82 & 0.36 & 0.98 & 0.91 & 0.91 & 0.13 & 0.98 & 0.08 & 0.47 & 0.05 & 0.25 & 0.14 & 0.14
& \multicolumn{14}{c}{ } \\

\gptfive
& 0.63 & 0.85 & 0.47 & 0.90 & 0.83 & 0.93 & 0.12 & 1.00 & 0.06 & 0.68 & 0.25 & 0.50 & 0.14 & 0.14
& \multicolumn{14}{c}{\textbf{N/A}} \\

\gptfivemini
& 0.61 & 0.86 & 0.36 & 0.95 & 0.93 & 0.91 & 0.12 & 0.98 & 0.06 & 0.68 & 0.04 & 0.25 & 0.00 & 0.00
& \multicolumn{14}{c}{\textbf{}} \\

\qwenthree
& 0.53 & 0.96 & 0.53 & 0.91 & 0.88 & 0.93 & 0.13 & 0.66 & 0.06 & 0.53 & 0.00 & 0.00 & 0.00 & 0.00
& \multicolumn{14}{c}{ } \\

\qwentwofive
& 0.52 & 0.88 & 0.62 & 0.28 & 0.81 & 0.18 & 0.11 & 0.06 & 0.02 & 0.05 & 0.00 & 0.00 & 0.02 & 0.43
& \multicolumn{14}{c}{ } \\

\qwentwofivechartqa
& 0.52 & 0.88 & 0.65 & 0.28 & 0.66 & 0.18 & 0.07 & 0.03 & 0.06 & 0.16 & 0.00 & 0.00 & 0.03 & 0.57
& \multicolumn{14}{c}{ } \\

\bottomrule
\end{tabular}%
}
\caption{Per-error Precision (P) and Recall (R) for reasoning error classification across the 2$\times$2 grid. Errors such as \textit{Failure to Account for Statistical Nuance} exhibit high recall but very low precision, highlighting a tendency among models to over-predict this label even when it is not present. Cells corresponding to \textit{Non-Misleading Caption, Misleading Viz} and \textit{Non-Misleading Caption, Non-Misleading Viz} do not contain any reasoning errors, so values are marked as \textbf{N/A}.}
\label{tab:reasoning_pr_rec_2x2}
\end{table*}

\begin{table*}[t]
\centering
\renewcommand{\arraystretch}{1.15}
\resizebox{\textwidth}{!}{%
\begin{tabular}{l c:c: c:c: c:c: c:c: c:c: c:c: c:c: c:c: c:c: c:c: c:c: c:c c:c c:c}
\toprule
\multicolumn{1}{c}{} & \multicolumn{14}{c:}{\textbf{$\triangle$}} & \multicolumn{14}{c}{\textbf{$\bigcirc$}} \\
\cmidrule(lr){2-15} \cmidrule(lr){16-29}

\textbf{Model}
& \multicolumn{2}{c}{\shortstack{Trunc.\\Axis}} 
& \multicolumn{2}{c}{\shortstack{Dual\\Axis}} 
& \multicolumn{2}{c}{\shortstack{Area/\\Vol.}} 
& \multicolumn{2}{c}{\shortstack{Inv.\\Axis}}
& \multicolumn{2}{c}{\shortstack{Uneven\\Bin.}} 
& \multicolumn{2}{c}{\shortstack{Unclr.\\Enc.}} 
& \multicolumn{2}{c}{\shortstack{Inappr.\\Enc.}}
& \multicolumn{2}{c}{\shortstack{Trunc.\\Axis}} 
& \multicolumn{2}{c:}{\shortstack{Dual\\Axis}} 
& \multicolumn{2}{c:}{\shortstack{Area/\\Vol.}} 
& \multicolumn{2}{c:}{\shortstack{Inv.\\Axis}}
& \multicolumn{2}{c:}{\shortstack{Uneven\\Bin.}} 
& \multicolumn{2}{c:}{\shortstack{Unclr.\\Enc.}} 
& \multicolumn{2}{c}{\shortstack{Inappr.\\Enc.}} \\
\cline{2-29}
& \multicolumn{1}{c}{P} & \multicolumn{1}{c}{R}
& \multicolumn{1}{c}{P} & \multicolumn{1}{c}{R}
& \multicolumn{1}{c}{P} & \multicolumn{1}{c}{R}
& \multicolumn{1}{c}{P} & \multicolumn{1}{c}{R}
& \multicolumn{1}{c}{P} & \multicolumn{1}{c}{R}
& \multicolumn{1}{c}{P} & \multicolumn{1}{c}{R}
& \multicolumn{1}{c}{P} & \multicolumn{1}{c}{R}

& \multicolumn{1}{c}{P} & \multicolumn{1}{c}{R}
& \multicolumn{1}{c}{P} & \multicolumn{1}{c}{R}
& \multicolumn{1}{c}{P} & \multicolumn{1}{c}{R}
& \multicolumn{1}{c}{P} & \multicolumn{1}{c}{R}
& \multicolumn{1}{c}{P} & \multicolumn{1}{c}{R}
& \multicolumn{1}{c}{P} & \multicolumn{1}{c}{R}
& \multicolumn{1}{c}{P} & \multicolumn{1}{c}{R} \\
\hline

\geminithreeonep & \multicolumn{14}{c:}{ }
& 0.73 & 0.88 & 0.96 & 0.99 & 0.88 & 0.58 & 0.57 & 0.57 & 0.11 & 0.75 & 0.56 & 0.64 & 0.29 & 0.57 \\

\geminithreezerop & \multicolumn{14}{c:}{ }
& 0.68 & 0.83 & 0.96 & 0.99 & 0.83 & 0.66 & 0.51 & 0.48 & 0.14 & 0.69 & 0.54 & 0.69 & 0.26 & 0.58 \\

\geminitwofivep
& \multicolumn{14}{c:}{ }
& 0.66 & 0.77 & 0.95 & 0.98 & 0.81 & 0.69 & 0.54 & 0.41 & 0.11 & 0.87 & 0.43 & 0.89 & 0.20 & 0.78 \\

\geminitwofivef
& \multicolumn{14}{c:}{ }
& 0.67 & 0.82 & 0.95 & 0.95 & 0.73 & 0.79 & 0.53 & 0.43 & 0.10 & 0.87 & 0.49 & 0.77 & 0.19 & 0.68 \\

\gptfive
& \multicolumn{14}{c:}{\textbf{N/A}}
& 0.68 & 0.62 & 0.97 & 0.98 & 0.78 & 0.77 & 0.29 & 0.11 & 0.13 & 0.81 & 0.47 & 0.84 & 0.21 & 0.64 \\

\gptfivemini
& \multicolumn{14}{c:}{\textbf{}}
& 0.70 & 0.70 & 0.97 & 0.98 & 0.65 & 0.83 & 0.48 & 0.23 & 0.19 & 0.94 & 0.48 & 0.89 & 0.21 & 0.60 \\

\qwenthree
& \multicolumn{14}{c:}{ }
& 0.32 & 0.69 & 0.96 & 0.92 & 0.72 & 0.55 & 0.50 & 0.09 & 0.25 & 0.12 & 0.37 & 0.87 & 0.15 & 0.77 \\

\qwentwofive
& \multicolumn{14}{c:}{ }
& 0.24 & 0.32 & 0.92 & 0.70 & 0.56 & 0.06 & 0.13 & 0.04 & 0.00 & 0.00 & 0.30 & 0.12 & 0.12 & 0.09 \\

\qwentwofivechartqa
& \multicolumn{14}{c:}{ }
& 0.24 & 0.20 & 0.74 & 0.80 & 0.48 & 0.10 & 0.02 & 0.02 & 0.00 & 0.00 & 0.52 & 0.09 & 0.11 & 0.11 \\

\hline

\multicolumn{1}{c}{} & \multicolumn{14}{c:}{\textbf{$\blacksquare$}} & \multicolumn{14}{c}{\textbf{$\varnothing$}} \\
\cmidrule(lr){2-15} \cmidrule(lr){16-29}

& \multicolumn{2}{c:}{\shortstack{Trunc.\\Axis}} 
& \multicolumn{2}{c:}{\shortstack{Dual\\Axis}} 
& \multicolumn{2}{c:}{\shortstack{Area/\\Vol.}} 
& \multicolumn{2}{c:}{\shortstack{Inv.\\Axis}}
& \multicolumn{2}{c:}{\shortstack{Uneven\\Bin.}} 
& \multicolumn{2}{c:}{\shortstack{Unclr.\\Enc.}} 
& \multicolumn{2}{c:}{\shortstack{Inappr.\\Enc.}}
& \multicolumn{2}{c}{\shortstack{Trunc.\\Axis}} 
& \multicolumn{2}{c}{\shortstack{Dual\\Axis}} 
& \multicolumn{2}{c}{\shortstack{Area/\\Vol.}} 
& \multicolumn{2}{c}{\shortstack{Inv.\\Axis}}
& \multicolumn{2}{c}{\shortstack{Uneven\\Bin.}} 
& \multicolumn{2}{c}{\shortstack{Unclr.\\Enc.}} 
& \multicolumn{2}{c}{\shortstack{Inappr.\\Enc.}} \\
\cline{2-29}
& \multicolumn{1}{c}{P} & \multicolumn{1}{c}{R}
& \multicolumn{1}{c}{P} & \multicolumn{1}{c}{R}
& \multicolumn{1}{c}{P} & \multicolumn{1}{c}{R}
& \multicolumn{1}{c}{P} & \multicolumn{1}{c}{R}
& \multicolumn{1}{c}{P} & \multicolumn{1}{c}{R}
& \multicolumn{1}{c}{P} & \multicolumn{1}{c}{R}
& \multicolumn{1}{c}{P} & \multicolumn{1}{c}{R}

& \multicolumn{1}{c}{P} & \multicolumn{1}{c}{R}
& \multicolumn{1}{c}{P} & \multicolumn{1}{c}{R}
& \multicolumn{1}{c}{P} & \multicolumn{1}{c}{R}
& \multicolumn{1}{c}{P} & \multicolumn{1}{c}{R}
& \multicolumn{1}{c}{P} & \multicolumn{1}{c}{R}
& \multicolumn{1}{c}{P} & \multicolumn{1}{c}{R}
& \multicolumn{1}{c}{P} & \multicolumn{1}{c}{R} \\

\hline

\geminithreeonep & 0.85 & 0.92 & 0.98 & 0.97 & 1.00 & 0.54 & 0.71 & 0.52 & 0.02 & 0.50 & 0.06 & 0.56 & 0.04 & 0.37 & \multicolumn{14}{c}{ } \\

\geminithreezerop & 0.83 & 0.90 & 0.97 & 0.98 & 1.00 & 0.71 & 0.88 & 0.54 & 0.02 & 0.50 & 0.04 & 0.50 & 0.07 & 0.62 & \multicolumn{14}{c}{ } \\

\geminitwofivep
& 0.85 & 0.82 & 0.96 & 0.97 & 0.98 & 0.67 & 1.00 & 0.45 & 0.02 & 0.50 & 0.04 & 0.81 & 0.02 & 0.50
& \multicolumn{14}{c}{ } \\

\geminitwofivef
& 0.76 & 0.92 & 0.95 & 0.95 & 0.95 & 0.74 & 1.00 & 0.48 & 0.01 & 0.50 & 0.05 & 0.62 & 0.03 & 0.62
& \multicolumn{14}{c}{ } \\

\gptfive
& 0.77 & 0.62 & 0.97 & 0.97 & 0.99 & 0.76 & 1.00 & 0.10 & 0.02 & 0.50 & 0.04 & 0.69 & 0.03 & 0.50
& \multicolumn{14}{c}{\textbf{N/A}} \\

\gptfivemini
& 0.81 & 0.76 & 0.98 & 0.98 & 0.94 & 0.80 & 1.00 & 0.38 & 0.02 & 0.50 & 0.04 & 0.69 & 0.02 & 0.25
& \multicolumn{14}{c}{\textbf{}} \\

\qwenthree
& 0.25 & 0.74 & 0.97 & 0.84 & 0.93 & 0.54 & 0.25 & 0.03 & 0.00 & 0.00 & 0.03 & 0.75 & 0.01 & 0.62
& \multicolumn{14}{c}{ } \\

\qwentwofive
& 0.35 & 0.36 & 0.88 & 0.86 & 0.77 & 0.11 & 0.18 & 0.14 & 0.00 & 0.00 & 0.04 & 0.12 & 0.02 & 0.25
& \multicolumn{14}{c}{ } \\

\qwentwofivechartqa
& 0.29 & 0.22 & 0.88 & 0.86 & 0.62 & 0.10 & 0.21 & 0.10 & 0.00 & 0.00 & 0.00 & 0.00 & 0.00 & 0.00
& \multicolumn{14}{c}{ } \\

\bottomrule
\end{tabular}%
}
\caption{Per-error Precision (P) and Recall (R) for visualization error classification across the 2$\times$2 grid. Models often exhibit high recall but low precision for certain categories such as \textit{Unclear Encoding}, suggesting over-prediction of these error types even when not warranted. Cells corresponding to \textit{Misleading Text, Non-Misleading Viz} and \textit{Non-Misleading Text, Non-Misleading Viz} do not contain visualization errors, so values are marked as \textbf{N/A}.}
\label{tab:viz_pr_rec_2x2}
\end{table*}

\begin{table*}[!h]
\centering
\resizebox{\linewidth}{!}{
\begin{tabular}{lccccccc:ccccccc}
\toprule
\multicolumn{1}{c}{} & \multicolumn{7}{c:}{\textbf{$\triangle$}} & \multicolumn{7}{c}{\textbf{$\bigcirc$}} \\
\cmidrule(lr){2-8} \cmidrule(lr){9-15}
\textbf{Model}
& \shortstack{Cher.\\Pick}
 & \shortstack{Caus.\\Infer.}
 & \shortstack{Arb.\\Thr.}
 & \shortstack{Stat.\\Nu.} 
 & \shortstack{Chart\\Read}
 & \shortstack{Data\\Val.} 
 & \shortstack{Mis.\\Sci.}
& \shortstack{Cher.\\Pick}
 & \shortstack{Caus.\\Infer.}
 & \shortstack{Arb.\\Thr.}
 & \shortstack{Stat.\\Nu.} 
 & \shortstack{Chart\\Read}
 & \shortstack{Data\\Val.} 
 & \shortstack{Mis.\\Sci.} \\
\midrule

\geminithreeonep & 0.22 & 0.07 & 0.01 & 0.52 & 0.12 & 0.02 & 0.01 & 0.04 & 0.06 & 0.02 & 0.15 & 0.24 & 0.01 & 0.00 \\

\geminithreezerop & 0.38 & 0.09 & 0.02 & 0.62 & 0.07 & 0.02 & 0.04 & 0.09 & 0.06 & 0.03 & 0.20 & 0.09 & 0.02 & 0.02 \\ 

\geminitwofivep & 0.53 & 0.25 & 0.02 & 0.83 & 0.07 & 0.06 & 0.04 & 0.21 & 0.15  & 0.06 & 0.46 & 0.18 & 0.15 & 0.02 \\ 
\geminitwofivef & 0.58 & 0.32 & 0.02 & 0.93 & 0.12 & 0.08 & 0.04 & 0.25 & 0.18 & 0.08 & 0.70 & 0.32 & 0.16 & 0.01 \\ 
\gptfive & 0.64 & 0.21 & 0.03 & 0.91 & 0.20 & 0.03 & 0.02 & 0.15 & 0.10 & 0.04 & 0.38 & 0.30 & 0.05 & 0.01 \\ 
\gptfivemini & 0.69 & 0.38 & 0.04 & 0.96 & 0.38 & 0.12 & 0.03 & 0.21 & 0.18 & 0.07 & 0.68 & 0.50 & 0.21 & 0.01  \\ 
\qwenthree & 0.70 & 0.18 & 0.10 & 0.27 & 0.58 & 0.00 & 0.01 & 0.28 & 0.14 & 0.10 & 0.22 & 0.49 & 0.00 & 0.00 \\ 
\qwentwofive & 0.54 & 0.01 & 0.10 & 0.06  & 0.26 & 0.00 & 0.09 & 0.22 & 0.01 & 0.05 & 0.02 & 0.16 & 0.00 & 0.02 \\ 
\qwentwofivechartqa & 0.59 & 0.03 & 0.02 & 0.03 & 0.07 & 0.01 & 0.15 & 0.31 & 0.02 & 0.02 & 0.02 & 0.05 & 0.01 & 0.07 \\ 
\midrule
\multicolumn{1}{c}{} & \multicolumn{7}{c:}{\textbf{$\blacksquare$}} & \multicolumn{7}{c}{\textbf{$\varnothing$}} \\
\cmidrule(lr){2-8} \cmidrule(lr){9-15}
& \shortstack{Cher.\\Pick}
 & \shortstack{Caus.\\Infer.}
 & \shortstack{Arb.\\Thr.}
 & \shortstack{Stat.\\Nu.} 
 & \shortstack{Chart\\Read}
 & \shortstack{Data\\Val.} 
 & \shortstack{Mis.\\Sci.}
& \shortstack{Cher.\\Pick}
 & \shortstack{Caus.\\Infer.}
 & \shortstack{Arb.\\Thr.}
 & \shortstack{Stat.\\Nu.} 
 & \shortstack{Chart\\Read}
 & \shortstack{Data\\Val.} 
 & \shortstack{Mis.\\Sci.} \\
\midrule

\geminithreeonep & 0.10 & 0.11 & 0.02 & 0.40 & 0.27 & 0.01 & 0.01 & 0.04 & 0.05 & 0.02 & 0.06 & 0.07 & 0.01 & 0.00 \\

\geminithreezerop & 0.19 & 0.16 & 0.03 & 0.46 & 0.28 & 0.02 & 0.01 & 0.05 & 0.04 & 0.03 & 0.08 & 0.02 & 0.01 & 0.00 \\ 

\geminitwofivep & 0.44 & 0.35 & 0.10 & 0.75 & 0.18 & 0.05 & 0.01 & 0.11 & 0.12 & 0.05 & 0.31 & 0.03 & 0.19 & 0.01 \\ 
\geminitwofivef & 0.46 & 0.58 & 0.06 & 0.95 & 0.23 & 0.04 & 0.01 & 0.19 & 0.17 & 0.08 & 0.58 & 0.10 & 0.18 & 0.00 \\ 
\gptfive & 0.51 & 0.34 & 0.14 & 0.98 & 0.39 & 0.01 & 0.01 & 0.08 & 0.08 & 0.06 & 0.31 & 0.11 & 0.04 & 0.00 \\ 
\gptfivemini & 0.55 & 0.58 & 0.05 & 0.99 & 0.39 & 0.05 & 0.01 & 0.16 & 0.15 & 0.10 & 0.62 & 0.17 & 0.21 & 0.00 \\ 
\qwenthree & 0.84 & 0.27 & 0.10 & 0.62 & 0.34 & 0.00 & 0.00 & 0.24 & 0.11 & 0.10 & 0.20 & 0.25 & 0.00 & 0.00 \\ 
\qwentwofive & 0.83 & 0.06 & 0.03 & 0.07 & 0.09 & 0.01 & 0.25 & 0.10 & 0.01 & 0.00 & 0.01 & 0.06 & 0.00 & 0.02 \\ 
\qwentwofivechartqa & 0.83 & 0.05 & 0.06 & 0.06 & 0.10 & 0.03 & 0.30 & 0.21 & 0.01 & 0.02 & 0.01 & 0.03 & 0.00 & 0.03 \\ 
\bottomrule
\end{tabular}
}
\caption{Per-error False Positive Rates (FPR) for reasoning error classification across the 2$\times$2 grid. Models show relatively high FPR for \textit{Failure to Account for Statistical Nuance}, indicating a tendency to over-predict this category even when not applicable.}
\label{tab:per_reasoning_error_fpr_2x2}
\end{table*}

\begin{table*}[!t]
\centering
\renewcommand{\arraystretch}{1}
\resizebox{1\linewidth}{!}{
\begin{tabular}{p{41pt}ccccccc:ccccccc}
\toprule
 & \multicolumn{7}{c:}{\textbf{Reasoning Errors}} 
 & \multicolumn{7}{c}{\textbf{Visualization Errors}} \\
\cmidrule(lr){2-8} \cmidrule(lr){9-15}

\small
\textbf{Model}
& \shortstack{Cher.\\Pick}
 & \shortstack{Caus.\\Infer.}
 & \shortstack{Arb.\\Thr.}
 & \shortstack{Stat.\\Nu.} 
 & \shortstack{Chart\\Read}
 & \shortstack{Data\\Val.} 
 & \shortstack{Mis.\\Sci.}
 & \shortstack{Trunc.\\Axis}
 & \shortstack{Dual\\Axis}
 & \shortstack{Area/\\Vol.} 
 & \shortstack{Inv.\\Axis}
 & \shortstack{Uneven\\Bin..} 
 & \shortstack{Unclr.\\Enc.} 
 & \shortstack{Inappr.\\Enc.} \\

\midrule
\normalsize

\geminithreeonep & 0.08 & 0.06 & 0.02 & 0.26 & 0.18 & 0.01 & 0.00 & 0.04 & 0.04 & 0.04 & 0.02 & 0.09 & 0.26 & 0.15 \\

\geminithreezerop & 0.15 & 0.07 & 0.03 & 0.32 & 0.10 & 0.02 & 0.02 & 0.04 & 0.04 & 0.04 & 0.01 & 0.06 & 0.27 & 0.17 \\

\geminitwofivep
& 0.28 & 0.19 & 0.05 & 0.57 & 0.12 & 0.12 & 0.02 & 0.04 & 0.04 & 0.06 & 0.01 & 0.10 & 0.62 & 0.38 \\

\geminitwofivef
& 0.32 & 0.26 & 0.06 & 0.77 & 0.20 & 0.12 & 0.02 & 0.06 & 0.04 & 0.10 & 0.02 & 0.11 & 0.42 & 0.34 \\

\gptfive
& 0.27 & 0.15 & 0.05 & 0.59 & 0.25 & 0.04 & 0.01 & 0.03 & 0.04 & 0.07 & 0.01 & 0.07 & 0.44 & 0.24 \\

\gptfivemini
& 0.33 & 0.26 & 0.07 & 0.79 & 0.38 & 0.16 & 0.01 & 0.04 & 0.04 & 0.13 & 0.01 & 0.05 & 0.45 & 0.24 \\

\qwenthree
& 0.41 & 0.16 & 0.10 & 0.29 & 0.44 & 0.00 & 0.00 & 0.34 & 0.05 & 0.06 & 0.01 & 0.00 & 0.67 & 0.54 \\

\qwentwofive
& 0.32 & 0.02 & 0.05 & 0.04 & 0.16 & 0.00 & 0.08 & 0.15 & 0.05  & 0.02 & 0.02 & 0.00 & 0.10 & 0.08 \\

\qwentwofivechartqa
& 0.40 & 0.02 & 0.03 & 0.03 & 0.06 & 0.01 & 0.12 & 0.12 & 0.11 & 0.03 & 0.03 & 0.00 & 0.04 & 0.09 \\
\bottomrule
\end{tabular}
}
\caption{Per-error False Postiive Rates (FPR) for reasoning and visualization error classification on the whole dataset. Models show relatively high FPR for certain error types such as \textit{Failure to Account for Statistical Nuance} and \textit{Unclear Encoding}, indicating a tendency to over-predict these categories even when not applicable.}
\vspace{-10pt}
\label{tab:per_error_fpr_whole_benchmark}
\end{table*}

\begin{table*}[!t]
\centering
\resizebox{\linewidth}{!}{
\renewcommand{\arraystretch}{1}
\begin{tabular}{lccccccc:ccccccc}
\toprule
\multicolumn{1}{c}{} & \multicolumn{7}{c:}{\textbf{$\triangle$}} & \multicolumn{7}{c}{\textbf{$\bigcirc$}} \\
\cmidrule(lr){2-8} \cmidrule(lr){9-15}
\textbf{Model}
& \shortstack{Trunc.\\Axis}
 & \shortstack{Dual\\Axis}
 & \shortstack{Area/\\Vol.} 
 & \shortstack{Inv.\\Axis}
 & \shortstack{Uneven\\Bin.} 
 & \shortstack{Unclr.\\Enc.} 
 & \shortstack{Inappr.\\Enc.}
& \shortstack{Trunc.\\Axis}
 & \shortstack{Dual\\Axis}
 & \shortstack{Area/\\Vol.} 
 & \shortstack{Inv.\\Axis}
 & \shortstack{Uneven\\Bin.} 
 & \shortstack{Unclr.\\Enc.} 
 & \shortstack{Inappr.\\Enc.} \\
\midrule

\geminithreeonep & 0.03 & 0.08 & 0.04 & 0.02 & 0.08 & 0.23 & 0.13 & 0.03 & 0.01 & 0.03 & 0.02 & 0.09 & 0.27 & 0.19 \\

\geminithreezerop & 0.03 & 0.09 & 0.04 & 0.01 & 0.03 & 0.16 & 0.14 & 0.04 & 0.01 & 0.04 & 0.02 & 0.06 & 0.33 & 0.23 \\ 

\geminitwofivep & 0.04 & 0.10 & 0.05 & 0.01 & 0.07 & 0.57 & 0.40 & 0.04 & 0.01 & 0.06 & 0.01 & 0.10 & 0.63 & 0.42 \\ 
\geminitwofivef & 0.06 & 0.09 & 0.07 & 0.01 & 0.07 & 0.38 & 0.33 & 0.04 & 0.01 & 0.10 & 0.02 & 0.12 & 0.43 & 0.40 \\ 
\gptfive & 0.04 & 0.08 & 0.06 & 0.01 & 0.04 & 0.33 & 0.18 & 0.03 & 0.01 & 0.08 & 0.01 & 0.08 & 0.52 & 0.32 \\ 
\gptfivemini & 0.03 & 0.09 & 0.11 & 0.01 & 0.04 & 0.35 & 0.21 & 0.03 & 0.01 & 0.16 & 0.01 & 0.06 & 0.52 & 0.30 \\ 
\qwenthree & 0.64 & 0.11 & 0.06 & 0.00 & 0.00 & 0.47 & 0.37 & 0.15 & 0.01 & 0.08 & 0.00 & 0.01 & 0.79 & 0.58 \\ 
\qwentwofive & 0.28 & 0.08 & 0.02 & 0.02 & 0.00 & 0.10 & 0.06 & 0.10 & 0.02 & 0.02 & 0.01 & 0.00 & 0.16 & 0.09 \\ 
\qwentwofivechartqa & 0.22 & 0.16 & 0.03 & 0.04 & 0.01 & 0.04 & 0.05 & 0.07 & 0.08 & 0.04 & 0.03 & 0.00 & 0.04 & 0.12 \\ 
\midrule
\multicolumn{1}{c}{} & \multicolumn{7}{c:}{\textbf{$\blacksquare$}} & \multicolumn{7}{c}{\textbf{$\varnothing$}} \\
\cmidrule(lr){2-8} \cmidrule(lr){9-15}
& \shortstack{Trunc.\\Axis}
 & \shortstack{Dual\\Axis}
 & \shortstack{Area/\\Vol.} 
 & \shortstack{Inv.\\Axis}
 & \shortstack{Uneven\\Bin.} 
 & \shortstack{Unclr.\\Enc.} 
 & \shortstack{Inappr.\\Enc.}
& \shortstack{Trunc.\\Axis}
 & \shortstack{Dual\\Axis}
 & \shortstack{Area/\\Vol.} 
 & \shortstack{Inv.\\Axis}
 & \shortstack{Uneven\\Bin.} 
 & \shortstack{Unclr.\\Enc.} 
 & \shortstack{Inappr.\\Enc.} \\
\midrule

\geminithreeonep & 0.02 & 0.01 & 0.00 & 0.01 & 0.11 & 0.29 & 0.14 & 0.08 & 0.02 & 0.06 & 0.03 & 0.10 & 0.28 & 0.13 \\

\geminithreezerop & 0.02 & 0.02 & 0.00 & 0.00 & 0.09 & 0.33 & 0.15 & 0.07 & 0.02 & 0.07 & 0.01 & 0.06 & 0.30 & 0.11 \\ 

\geminitwofivep & 0.02 & 0.02 & 0.01 & 0.00 & 0.13 & 0.60 & 0.38 & 0.07 & 0.02 & 0.09 & 0.02 & 0.11 & 0.70 & 0.30\\ 
\geminitwofivef & 0.03 & 0.03 & 0.03 & 0.00  & 0.16 & 0.41 & 0.36 & 0.10 & 0.03 & 0.15 & 0.03 & 0.11 & 0.47 & 0.27\\ 
\gptfive & 0.02 & 0.02 & 0.00 & 0.00 & 0.12 & 0.50 & 0.24 & 0.05 & 0.02 & 0.10 & 0.02 & 0.07 & 0.43 & 0.17 \\
\gptfivemini & 0.02 & 0.01 & 0.04 & 0.00 & 0.08 & 0.51 & 0.27 & 0.07 & 0.02 & 0.16 & 0.01 & 0.05 & 0.44 & 0.17 \\ 
\qwenthree & 0.24 & 0.02 & 0.03 & 0.01 & 0.00 & 0.84 & 0.68 & 0.35 & 0.04 & 0.06 & 0.01 & 0.00 & 0.66 & 0.59\\ 
\qwentwofive & 0.08 & 0.07 & 0.02 & 0.04 & 0.00 & 0.09 & 0.17 & 0.11 & 0.04 & 0.01 & 0.01 & 0.00 & 0.06 & 0.03 \\ 
\qwentwofivechartqa & 0.06 & 0.07 & 0.05  & 0.02 & 0.00 & 0.07 & 0.15 & 0.10 & 0.09 & 0.02 & 0.03 & 0.00 & 0.03 & 0.04\\ 
\bottomrule
\end{tabular}
}
\caption{Per-error False Positive Rates (FPR) for visualization error classification across the 2$\times$2 grid. Models show relatively high FPR for \textit{Unclear Encoding}, indicating a tendency to over-predict this category even when not applicable.}
\label{tab:per_visualization_error_fpr_2x2}
\end{table*}

\clearpage
\newpage
\onecolumn
\subsection{Comparison to prior-work}
\label{app:prior_work_comparison}
\begin{table}[!h]
\centering
\small
\setlength{\tabcolsep}{4pt}
\resizebox{0.6\linewidth}{!}{

\rotatebox{90}{%
\begin{tabular}{p{2.6cm} p{2.6cm} p{2.8cm} p{2.2cm} p{1.8cm} p{1.8cm} p{2.4cm} p{2.8cm}}
\toprule
\textbf{} &
\textbf{Task} &
\textbf{Source of deception} &
\textbf{Error Granularity} &
\textbf{Chart-Caption Interaction} &
\textbf{Error Attribution} &
\textbf{Data Source} &
\textbf{Evaluation Focus} \\
\midrule

\cite{kahou2017figureqa,kafle2018dvqa,methani2020plotqa,masry2022chartqa} &
Chart question answering &
Assumed truthful charts &
None &
\xmark &
\xmark &
Synthetic/curated charts &
Chart comprehension accuracy \\

\cite{huang-etal-2024-lvlms} &
Caption factuality checking and correction &
Caption-chart factual mismatch &
Low (value, label, trend) &
\cmark &
\xmark &
Generated captions &
Caption factual consistency \\

\cite{akhtar2024chartcheck} &
Fact-checking claims against charts &
Explicit factual claims &
Moderate (supported/refuted + explanation) &
\cmark &
\xmark &
Real-world charts &
Claim verification accuracy \\

\cite{chen2025unmasking} &
MCQ-based misleading chart reasoning &
Visualization design manipulation &
Fine-grained (21 misleaders) &
\bcircle &
\bcircle &
Synthetic + standardized charts &
Answer correctness and reasoning \\

\cite{lo2024good, alexander2024can}
 &
Misleader detection via prompting &
Visualization design errors &
Medium (design issues) &
\xmark &
\bcircle &
Social media charts &
Prompt sensitivity and detection AUC \\

\cite{tonglet2025chart} &
Misleader classification &
Visualization design violations &
Fine-grained (12+ misleaders) &
\xmark &
\cmark (design only) &
Real + synthetic charts &
Multi-label misleader detection \\

\cite{mahbub2025perils} &
Susceptibility analysis of VLMs &
Visualization design distortions &
Design-level &
\xmark &
\xmark &
Synthetic chart pairs &
Behavioral degradation analysis \\

\midrule
\textbf{Ours} &
\textbf{Misleading chart-caption detection} &
\textbf{Caption-based reasoning errors \& visualization errors} &
\textbf{Fine-grained (reasoning + design taxonomy)} &
\textbf{\cmark} &
\textbf{\cmark (caption vs.\ visualization)} &
\textbf{Real-world charts + human-authored captions} &
\textbf{Diagnostic error attribution and over-flagging analysis} \\

\bottomrule
\end{tabular}
}
}
\caption{Comparison of our benchmark with prior work on misleading visualizations and chart understanding. Unlike existing benchmarks, our work explicitly disentangles misleadingness arising from caption-level reasoning errors versus visualization design errors and evaluates fine-grained error attribution by Vision-Language Models.}
\label{tab:benchmark-comparison}
\end{table}

\end{document}

%% file: related_works_raj.tex
\section{Related Works}
\subsection{Visualization Design and Misleading Communication}
Visualization research has long documented how charts can mislead audiences even without manipulating the underlying data, with many early studies focused on categorizing and describing visual distortion techniques, such as truncated or inverted axes, misleading aspect ratios, inappropriate legends, etc. \cite{pandey2015deceptive,correll2017black, lo2022misinformed}.
These detailed, descriptive error taxonomies enable researchers to better discuss the extent and impact of visualization misinformation \cite{correll2017black}.
They also help researchers explain why misinterpretation occurs (e.g., the graphical literacy of designers and audiences \cite{lo2022misinformed, lan2024came}) and develop tools to automatically detect and mitigate the effect of different errors \cite{chen2021vizlinter}.

However, recent work has shown that misleading visualizations are not limited to graphical distortions. \citet{lisnic2023misleading} conducted a large-scale analysis of COVID-19 charts shared on X (formerly Twitter) and found that the majority of misleading cases stemmed not from visual design flaws, but from \textit{reasoning errors in the accompanying captions}.
Similar work by \citet{lan2024came} also found reasoning errors in an online gallery of misleading visualizations curated by the public. Taken together, these findings suggest that \textit{focusing solely on visual distortions significantly underestimates how visualizations are used to misinform in real-world settings.}

Our work builds on and extends these prior studies by examining a combined taxonomy of \emph{visual design and reasoning errors}.
Unlike prior findings that describe how humans produce and interpret misleading visualizations, we examine the potential of VLMs as scalable, automated detectors of visual and reasoning errors, grounded in taxonomies derived from human misinformation practices.

\begin{figure}[t]
\centering
\includegraphics[trim={5pt 5pt 5pt 5pt}, width=\linewidth]{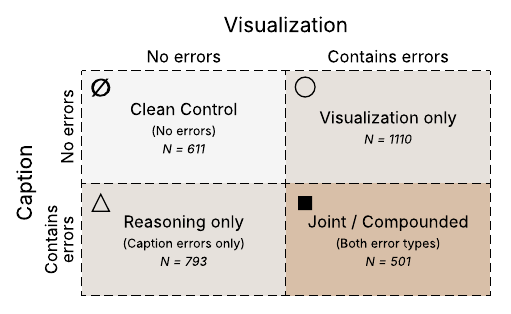}
\caption{Structure of our dataset organized as a 2$\times$2 grid based on the presence or absence of misleading content in captions and visualizations. Counts denote the number of chart-caption pairs in each cell. Symbols denote error composition: $\varnothing$ no errors, $\triangle$ caption-only errors, $\bigcirc$ visualization-only errors, $\blacksquare$ joint errors.}
\label{fig:2by2_grid}
\vspace{-12pt}
\end{figure}

\subsection{Misinformation Detection Capabilities of VLMs}
Most prior work has focused on chart understanding: evaluating whether a VLM can interpret visual elements, extract values, and answer questions about data visualizations \cite{guo2024understanding}. Early benchmarks such as FigureQA \cite{kahou2017figureqa}, DVQA \cite{kafle2018dvqa}, PlotQA \cite{methani2020plotqa}, ChartQA \cite{masry2022chartqa}, and LEAF-QA \cite{chaudhry2020leaf} primarily assessed a model's ability to interpret and answer questions about charts under the assumption that the charts themselves are truthful. Other works extend this line of work to chart summarization, chart-to-table conversion, and chart-based fact verification \cite{masry2023unichart, islam2024large, lo2024understanding}.

More recently, research has focused on multimodal misinformation detection in visualizations, examining visual cues of deception, though typically without reasoning about accompanying text \cite{alexander2024can, chen2025unmasking, wu2025seeing}. 
Some efforts focus on visual distortions in charts, including detecting design-principle violations (e.g., truncated axes or misleading scales; \citet{tonglet2025chart}) and evaluating VLMs' vulnerability to such distortions using inference-time correction strategies (e.g., table extraction and redrawing; \citet{tonglet2025protecting}).
Other studies address image-text inconsistencies, like identifying out-of-context image captions in which a real image is paired with an incorrect description \cite{kalla2024covlm}. However, such approaches are \emph{restricted to visual distortions in charts and do not evaluate whether models can detect reasoning-based deception in which the chart's caption draws false conclusions from the data (see Appendix Table~\ref{tab:benchmark-comparison}).} Motivated by this, we focus on detecting misinformation in visualization-caption pairs that explicitly differentiate visualization design errors from reasoning errors in the caption. 

%% file: Problem_new_version.tex
\section{Problem Setup and Methodology}

We structure our problem along two orthogonal dimensions: whether the visualization is misleading and whether the caption is misleading, producing a 2$\times$2 decomposition that isolates different modes of misinformation (Figure~\ref{fig:2by2_grid}). 

\subsection{Error Taxonomy}

We adopt the taxonomy of visualization design errors and caption-level reasoning errors introduced by \citet{lisnic2023misleading}. Table \ref{tab:error_taxonomy_short} provides the abbreviated descriptions of each error category. 
The reasoning taxonomy comprises seven caption-level errors: cherry-picking, setting an arbitrary threshold, causal inference, failure to account for statistical nuance, incorrect interpretation of the chart, issues with data validity, and misrepresentation of scientific studies. The visualization taxonomy similarly includes seven error types: truncated axis, dual axis, value encoded as area or volume, inverted axis, uneven binning, unclear encoding, and inappropriate encoding.  
Each chart-caption pair may contain zero, one, or multiple errors, reflecting the fact that misleading communication often combines several forms of distortion. Detailed definitions, distributions, and examples for all error categories are in Appendix~\ref{app:error_descriptions}, ~\ref{app:examples_from_benchmark}, and ~\ref{app:reasning_viz_composition}.

\begin{table*}[t]
\centering

\resizebox{\linewidth}{!}{
\renewcommand{\arraystretch}{1}
\begin{tabular}{p{5cm} p{12.5cm}}
\toprule

 \multicolumn{2}{c}{\textbf{Reasoning Errors}} \\
 \hline

 Cherry-picking & 
Selectively highlighting data subsets that support a claim while ignoring context. \\

 Causal Inference & 
Claiming causation based solely on correlation or temporal association. \\

 Setting an Arbitrary Threshold & 
Introducing an unjustified cutoff to frame comparisons as meaningful. \\

 Statistical Nuance & 
Ignoring uncertainty, baselines, or statistical significance when interpreting data. \\

 Incorrect Reading of the Chart & 
Misinterpreting trends or values shown in the visualization. \\

 Issues with Data Validity & 
Questioning data reliability or integrity without substantiated evidence. \\

 Misrepresentation of Studies & 
Exaggerating or selectively citing scientific findings to support a claim. \\

\midrule

\multicolumn{2}{c}{\textbf{Visualization Errors}} \\
\hline

Truncated Axis & 
Manipulating axis ranges to exaggerate visual differences or trends. \\

 Dual Axis & 
Using multiple axes with unrelated scales to suggest misleading associations. \\

 Values as Area / Volume & 
Encoding values via area or volume, leading to perceptual distortion. \\

 Inverted Axis & 
Reversing axis direction in a way that obscures or flips trends. \\

 Uneven Binning & 
Using non-uniform bins to distort distributions or comparisons. \\

 Unclear Encoding & 
Using ambiguous or insufficiently labeled visual elements. \\

 Inappropriate Encoding & 
Applying a chart type unsuitable for the data semantics. \\

\bottomrule
\end{tabular}
}

\caption{Abbreviated definition of the error types used in our dataset. Full descriptions and examples are provided in Appendix Tables~\ref{tab:reasoning_error_descriptions}, \ref{tab:visualization_error_descriptions}, \ref{tab:reasoning_err_eg}, and \ref{tab:visualization_error_eg}.}
\vspace{-12pt}
\label{tab:error_taxonomy_short}
\end{table*}

\subsection{Dataset Construction}
To support a controlled analysis of these error sources, we construct a dataset organized to isolate different modes of misleadingness\footnote{We provide the whole dataset at \url{https://huggingface.co/datasets/HarvardVCG/MisVisBench} as an artifact under the CC BY-NC-SA 4.0 license.}. Samples are drawn from multiple sources of \emph{real-world charts}, such as X or Reddit, and then used to populate the 2$\times$2 grid. 

\paragraph{Populating the 2$\times$2 grid.}
Each chart-caption pair is assigned to one of four conditions depending on whether the caption and/or the visualization contains misleading content (Figure~\ref{fig:2by2_grid}). 
$\triangle$ Chart-caption pairs with misleading captions and non-misleading visualizations are drawn from \citet{lisnic2023misleading} (CC BY 4.0). $\bigcirc$ For samples with misleading visualizations and non-misleading captions, we combine examples from \citet{lisnic2023misleading} with additional visualizations obtained from the \emph{r/DataIsUgly subreddit}. $\blacksquare$ For cases where both the visualization and caption are misleading, we reuse charts exhibiting visualization design errors from \citet{lisnic2023misleading} and author new captions that introduce specific reasoning errors. $\varnothing$ Finally, non-misleading chart-caption pairs are collected from the \emph{r/DataIsBeautiful} subreddit and manually verified to ensure that neither visualization design errors nor reasoning errors are present.

\noindent \textbf{Note.} For samples collected from \emph{r/DataIsUgly} ($\bigcirc$ condition), the authors manually annotated (see annotation interface in Appendix Figure~\ref{fig:ui_picture_example}) the visualization error types in the charts. Initially, annotation guidelines were refined through pilot labeling and discussion among the authors to clarify category boundaries and resolve ambiguities. To assess annotation reliability, a subset of 50 samples was independently annotated by multiple authors, yielding a Krippendorff's $\alpha$ of 0.81 across visualization error categories. Following this validation step, the remaining samples were individually annotated using the finalized guidelines.
Annotation and verification details for all samples in our dataset, including those manually annotated by the authors as well as those inherited or curated from external sources, are provided in Appendix~\ref{app:annotation_verification_metrics}.

\noindent \textbf{Dataset statistics.} A sample in our dataset can contain one or more reasoning and visualization errors. While most samples contain a single error, a considerable subset includes multiple errors. We provide the exact statistics in Appendix Table~\ref{tab:error_count_distribution}.

\subsection{Task Definition}

We study whether VLMs can identify and attribute misleadingness in chart-caption pairs by formulating two related multi-label classification tasks: reasoning-error and visualization-error classification. For each sample, the model is provided with the visualization, the accompanying caption (if any), and natural language descriptions of the relevant error categories. The two tasks are evaluated independently to isolate model behavior on caption-level reasoning versus visual design errors. Models are asked to predict the set of applicable error categories and provide a brief justification for each prediction; if no error applies, the model outputs \texttt{[``None"]}. Details on the prompts, their construction, and ablations are in Appendix~\ref{app:prompts}, ~\ref{app:prompt_ablation}. 



\subsection{Models Studied}

\begin{table}[t]
\centering
\resizebox{\columnwidth}{!}{
\begin{tabular}{p{2.2cm} l l}
\toprule
\textbf{Family} & \textbf{Symbol} & \textbf{Model} \\
\midrule
\multirow{4}{2.2cm}{\parbox{2.2cm}{\raggedright Gemini \cite{comanici2025gemini}}}
  & \geminithreeonep & Gemini-3.1-Pro-Preview \\
  & \geminithreezerop & Gemini-3.0-Pro-Preview \\
  & \geminitwofivep & Gemini-2.5-Pro \\
  & \geminitwofivef & Gemini-2.5-Flash \\
\midrule
\multirow{2}{2.2cm}{\parbox{2.2cm}{\raggedright GPT \cite{openai_gpt5_2025}}}
  & \gptfive & GPT-5 (25-08-07)\\
  & \gptfivemini & GPT-5-mini (25-08-07)\\
\midrule

\multirow{3}{2.2cm}{\parbox{2.2cm}{\raggedright Qwen \cite{Qwen3-VL,bai2025qwen2, askanything-charts-qwen2.5}}}
  & \qwenthree & Qwen3-VL-30B-A3B \\
  & \qwentwofive & Qwen2.5-VL-7B \\
  & \qwentwofivechartqa & Qwen2.5-VL-7B-ChartQA \\
\bottomrule
\end{tabular}
}
\caption{\normalsize Vision-language models studied in our analysis. Models are grouped by family, with symbol identifiers used throughout the paper.}
\vspace{-14pt}
\label{tab:models}
\end{table}

We explore a set of widely used proprietary and open-source vision-language models that report strong performance on existing multimodal benchmarks \cite{chiang2024chatbot}. The models span multiple families and architectural choices, including general-purpose frontier models and a chart-specialized variant \cite{askanything-charts-qwen2.5}. Table~\ref{tab:models} summarizes the models included in our study. All models are evaluated using a consistent inference configuration. To account for stochastic decoding, we run each model multiple times and observe consistent performance across runs, indicating that our conclusions are not driven by a single favorable generation. Additional details on inference configuration, retry policies, and run stability are provided in Appendix~\ref{app:inference_config} and \ref{app:stability_multiple_runs}.



\subsection{Evaluation Measures}

We use multiple evaluation measures to characterize how models identify and attribute misleadingness in chart-caption pairs. As each sample may contain multiple reasoning and visualization errors, we adopt metrics that capture both partial detection and complete attribution. Following prior work \cite{tonglet2025chart}, we report weighted F1, Partial Match, and Exact Match scores, computed separately for reasoning errors, visualization errors, and their combination. The F1 score provides fine-grained per-error performance; PM captures the model's ability to identify at least some of the misinformation present (useful in multi-label settings); and EM sets a strict bar for complete and accurate error detection across modalities.

\noindent \textbf{F1 Score} We calculate per-error-type F1 scores for each reasoning and visualization error category. We also compute weighted F1 scores separately for (i) \textit{reasoning error classification}, where the score is the weighted average over the 7 reasoning error categories, and (ii) \textit{visualization error classification}, where the score is the weighted average over the 7 visualization error categories. We also calculate a \textit{combined weighted} F1 score computed as a weighted average over all 14 categories (7 reasoning + 7 visualization). We additionally report macro-averaged F1 scores for reasoning error classification, visualization error classification, and the combined setting in Appendix Table \ref{tab:macro_f1_whole_benchmark}.

\noindent \textbf{Partial Match (PM)} is computed at three levels: (i) \textit{reasoning-only}, where a sample is counted as a match if the predicted reasoning error set overlaps with the ground-truth reasoning error set; (ii) \textit{visualization-only}, defined analogously for visualization errors; and (iii) \textit{combined}, where a sample is counted as a partial match if there is a partial match with \textit{any subset} of the reasoning or the visualization errors.

\noindent \textbf{Exact Match (EM)} is also computed at three levels: (i) \textit{reasoning-only}, where the predicted reasoning error set must exactly equal the ground-truth reasoning error set; (ii) \textit{visualization-only}, defined analogously for visualization errors; and (iii) \textit{combined}, where a sample is counted as an exact match only if the model achieves an exact match on \textit{both} the reasoning and the visualization errors.




%% file: arxiv_new_version.bbl
\begin{thebibliography}{41}
\providecommand{\natexlab}[1]{#1}

\bibitem[{Akhtar et~al.(2024)Akhtar, Subedi, Gupta, Tahmasebi, Cocarascu, and Simperl}]{akhtar2024chartcheck}
Mubashara Akhtar, Nikesh Subedi, Vivek Gupta, Sahar Tahmasebi, Oana Cocarascu, and Elena Simperl. 2024.
\newblock Chartcheck: Explainable fact-checking over real-world chart images.
\newblock In \emph{Findings of the Association for Computational Linguistics: ACL 2024}, pages 13921--13937.

\bibitem[{Alexander et~al.(2024)Alexander, Nanda, Yang, and Sarvghad}]{alexander2024can}
Jason Alexander, Priyal Nanda, Kai-Cheng Yang, and Ali Sarvghad. 2024.
\newblock Can gpt-4 models detect misleading visualizations?
\newblock In \emph{2024 IEEE Visualization and Visual Analytics (VIS)}, pages 106--110. IEEE.

\bibitem[{Bai et~al.(2025{\natexlab{a}})Bai, Cai, Chen, Chen, Chen, Cheng, Deng, Ding, Gao, Ge, Ge, Guo, Huang, Huang, Huang, Hui, Jiang, Li, Li, Li, Li, Lin, Lin, Liu, Liu, Liu, Liu, Liu, Liu, Lu, Luo, Lv, Men, Meng, Ren, Ren, Song, Sun, Tang, Tu, Wan, Wang, Wang, Wang, Wang, Xie, Xu, Xu, Xu, Yang, Yang, Yang, Yang, Yu, Zhang, Zhang, Zhang, Zheng, Zhong, Zhou, Zhou, Zhou, Zhu, and Zhu}]{Qwen3-VL}
Shuai Bai, Yuxuan Cai, Ruizhe Chen, Keqin Chen, Xionghui Chen, Zesen Cheng, Lianghao Deng, Wei Ding, Chang Gao, Chunjiang Ge, Wenbin Ge, Zhifang Guo, Qidong Huang, Jie Huang, Fei Huang, Binyuan Hui, Shutong Jiang, Zhaohai Li, Mingsheng Li, and 45 others. 2025{\natexlab{a}}.
\newblock Qwen3-vl technical report.
\newblock \emph{arXiv preprint arXiv:2511.21631}.

\bibitem[{Bai et~al.(2025{\natexlab{b}})Bai, Chen, Liu, Wang, Ge, Song, Dang, Wang, Wang, Tang et~al.}]{bai2025qwen2}
Shuai Bai, Keqin Chen, Xuejing Liu, Jialin Wang, Wenbin Ge, Sibo Song, Kai Dang, Peng Wang, Shijie Wang, Jun Tang, and 1 others. 2025{\natexlab{b}}.
\newblock Qwen2. 5-vl technical report.
\newblock \emph{arXiv preprint arXiv:2502.13923}.

\bibitem[{Chaudhry et~al.(2020)Chaudhry, Shekhar, Gupta, Maneriker, Bansal, and Joshi}]{chaudhry2020leaf}
Ritwick Chaudhry, Sumit Shekhar, Utkarsh Gupta, Pranav Maneriker, Prann Bansal, and Ajay Joshi. 2020.
\newblock Leaf-qa: Locate, encode \& attend for figure question answering.
\newblock In \emph{Proceedings of the IEEE/CVF winter conference on applications of computer vision}, pages 3512--3521.

\bibitem[{Chen et~al.(2021)Chen, Sun, Xu, Chen, Wang, and Cao}]{chen2021vizlinter}
Qing Chen, Fuling Sun, Xinyue Xu, Zui Chen, Jiazhe Wang, and Nan Cao. 2021.
\newblock Vizlinter: A linter and fixer framework for data visualization.
\newblock \emph{IEEE transactions on visualization and computer graphics}, 28(1):206--216.

\bibitem[{Chen et~al.(2025)Chen, Song, Shum, Lin, Sheng, Wang, and Qu}]{chen2025unmasking}
Zixin Chen, Sicheng Song, Kashun Shum, Yanna Lin, Rui Sheng, Weiqi Wang, and Huamin Qu. 2025.
\newblock Unmasking deceptive visuals: Benchmarking multimodal large language models on misleading chart question answering.
\newblock In \emph{Proceedings of the 2025 Conference on Empirical Methods in Natural Language Processing}, pages 13767--13800.

\bibitem[{Chhipa(2025)}]{askanything-charts-qwen2.5}
Prakash~Chandra Chhipa. 2025.
\newblock \href {https://huggingface.co/prakashchhipa/Qwen2.5-VL-7B-ChartQA-LoRA} {Askanythingincharts-qwen2.5-7b: Fine-tuned qwen2.5-vl for chart understanding}.

\bibitem[{Chiang et~al.(2024)Chiang, Zheng, Sheng, Angelopoulos, Li, Li, Zhu, Zhang, Jordan, Gonzalez et~al.}]{chiang2024chatbot}
Wei-Lin Chiang, Lianmin Zheng, Ying Sheng, Anastasios~Nikolas Angelopoulos, Tianle Li, Dacheng Li, Banghua Zhu, Hao Zhang, Michael Jordan, Joseph~E Gonzalez, and 1 others. 2024.
\newblock Chatbot arena: An open platform for evaluating llms by human preference.
\newblock In \emph{Forty-first International Conference on Machine Learning}.

\bibitem[{Comanici et~al.(2025)Comanici, Bieber, Schaekermann, Pasupat, Sachdeva, Dhillon, Blistein, Ram, Zhang, Rosen et~al.}]{comanici2025gemini}
Gheorghe Comanici, Eric Bieber, Mike Schaekermann, Ice Pasupat, Noveen Sachdeva, Inderjit Dhillon, Marcel Blistein, Ori Ram, Dan Zhang, Evan Rosen, and 1 others. 2025.
\newblock Gemini 2.5: Pushing the frontier with advanced reasoning, multimodality, long context, and next generation agentic capabilities.
\newblock \emph{arXiv preprint arXiv:2507.06261}.

\bibitem[{Correll and Heer(2017)}]{correll2017black}
Michael Correll and Jeffrey Heer. 2017.
\newblock Black hat visualization.
\newblock In \emph{Workshop on Dealing with Cognitive Biases in Visualisations (DECISIVe), IEEE VIS}, volume~1, page~10.

\bibitem[{Duarte et~al.(2022)Duarte, Carvalhais, and Amado}]{duarte2022role}
Ana Duarte, Miguel Carvalhais, and Pedro Amado. 2022.
\newblock The role of data visualization in science communication: Principles, encoding, and design patterns.
\newblock In \emph{International Conference on Design and Digital Communication}, pages 753--764. Springer.

\bibitem[{Fu and Stasko(2023)}]{fu2023more}
Yu~Fu and John Stasko. 2023.
\newblock More than data stories: Broadening the role of visualization in contemporary journalism.
\newblock \emph{IEEE Transactions on Visualization and Computer Graphics}.

\bibitem[{Guo et~al.(2024)Guo, Kang, Shah, Pfister, and Varma}]{guo2024understanding}
Grace Guo, Jenna~Jiayi Kang, Raj~Sanjay Shah, Hanspeter Pfister, and Sashank Varma. 2024.
\newblock Understanding graphical perception in data visualization through zero-shot prompting of vision-language models.
\newblock \emph{arXiv preprint arXiv:2411.00257}.

\bibitem[{Huang et~al.(2024)Huang, Zhou, Chan, Fung, Wang, Zhang, Chang, and Ji}]{huang-etal-2024-lvlms}
Kung-Hsiang Huang, Mingyang Zhou, Hou~Pong Chan, Yi~Fung, Zhenhailong Wang, Lingyu Zhang, Shih-Fu Chang, and Heng Ji. 2024.
\newblock \href {https://doi.org/10.18653/v1/2024.findings-acl.41} {Do {LVLM}s understand charts? analyzing and correcting factual errors in chart captioning}.
\newblock In \emph{Findings of the Association for Computational Linguistics: ACL 2024}, pages 730--749, Bangkok, Thailand. Association for Computational Linguistics.

\bibitem[{Islam et~al.(2024)Islam, Rahman, Masry, Laskar, Nayeem, and Hoque}]{islam2024large}
Mohammed~Saidul Islam, Raian Rahman, Ahmed Masry, Md~Tahmid~Rahman Laskar, Mir~Tafseer Nayeem, and Enamul Hoque. 2024.
\newblock Are large vision language models up to the challenge of chart comprehension and reasoning? an extensive investigation into the capabilities and limitations of lvlms.
\newblock \emph{arXiv preprint arXiv:2406.00257}.

\bibitem[{Kafle et~al.(2018)Kafle, Price, Cohen, and Kanan}]{kafle2018dvqa}
Kushal Kafle, Brian Price, Scott Cohen, and Christopher Kanan. 2018.
\newblock Dvqa: Understanding data visualizations via question answering.
\newblock In \emph{Proceedings of the IEEE conference on computer vision and pattern recognition}, pages 5648--5656.

\bibitem[{Kahou et~al.(2017)Kahou, Michalski, Atkinson, K{\'a}d{\'a}r, Trischler, and Bengio}]{kahou2017figureqa}
Samira~Ebrahimi Kahou, Vincent Michalski, Adam Atkinson, {\'A}kos K{\'a}d{\'a}r, Adam Trischler, and Yoshua Bengio. 2017.
\newblock Figureqa: An annotated figure dataset for visual reasoning.
\newblock \emph{arXiv preprint arXiv:1710.07300}.

\bibitem[{Kalla et~al.(2024)Kalla, Biswas et~al.}]{kalla2024covlm}
Jayateja Kalla, Soma Biswas, and 1 others. 2024.
\newblock Covlm: Leveraging consensus from vision-language models for semi-supervised multimodal fake news detection.
\newblock In \emph{Proceedings of the Asian Conference on Computer Vision}, pages 1197--1214.

\bibitem[{Lan and Liu(2024)}]{lan2024came}
Xingyu Lan and Yu~Liu. 2024.
\newblock “i came across a junk”: Understanding design flaws of data visualization from the public's perspective.
\newblock \emph{IEEE Transactions on Visualization and Computer Graphics}.

\bibitem[{Lisnic et~al.(2023)Lisnic, Polychronis, Lex, and Kogan}]{lisnic2023misleading}
Maxim Lisnic, Cole Polychronis, Alexander Lex, and Marina Kogan. 2023.
\newblock Misleading beyond visual tricks: How people actually lie with charts.
\newblock In \emph{Proceedings of the 2023 CHI conference on human factors in computing systems}, pages 1--21.

\bibitem[{Lo et~al.(2022)Lo, Gupta, Shigyo, Wu, Bertini, and Qu}]{lo2022misinformed}
Leo Yu-Ho Lo, Ayush Gupta, Kento Shigyo, Aoyu Wu, Enrico Bertini, and Huamin Qu. 2022.
\newblock Misinformed by visualization: What do we learn from misinformative visualizations?
\newblock In \emph{Computer Graphics Forum}, volume~41, pages 515--525. Wiley Online Library.

\bibitem[{Lo and Qu(2024)}]{lo2024good}
Leo Yu-Ho Lo and Huamin Qu. 2024.
\newblock How good (or bad) are llms at detecting misleading visualizations?
\newblock \emph{IEEE Transactions on Visualization and Computer Graphics}.

\bibitem[{Lo(2024)}]{lo2024understanding}
Yu~Ho Lo. 2024.
\newblock \emph{On Understanding Misleading Visualizations, Automatic Detection, and Prevention}.
\newblock Hong Kong University of Science and Technology (Hong Kong).

\bibitem[{Mahbub et~al.(2025)Mahbub, Islam, Laskar, Rahman, Nayeem, and Hoque}]{mahbub2025perils}
Ridwan Mahbub, Mohammed~Saidul Islam, Md~Tahmid~Rahman Laskar, Mizanur Rahman, Mir~Tafseer Nayeem, and Enamul Hoque. 2025.
\newblock The perils of chart deception: How misleading visualizations affect vision-language models.
\newblock \emph{arXiv preprint arXiv:2508.09716}.

\bibitem[{Masry et~al.(2022)Masry, Do, Tan, Joty, and Hoque}]{masry2022chartqa}
Ahmed Masry, Xuan~Long Do, Jia~Qing Tan, Shafiq Joty, and Enamul Hoque. 2022.
\newblock Chartqa: A benchmark for question answering about charts with visual and logical reasoning.
\newblock In \emph{Findings of the association for computational linguistics: ACL 2022}, pages 2263--2279.

\bibitem[{Masry et~al.(2023)Masry, Kavehzadeh, Do, Hoque, and Joty}]{masry2023unichart}
Ahmed Masry, Parsa Kavehzadeh, Xuan~Long Do, Enamul Hoque, and Shafiq Joty. 2023.
\newblock Unichart: A universal vision-language pretrained model for chart comprehension and reasoning.
\newblock \emph{arXiv preprint arXiv:2305.14761}.

\bibitem[{Methani et~al.(2020)Methani, Ganguly, Khapra, and Kumar}]{methani2020plotqa}
Nitesh Methani, Pritha Ganguly, Mitesh~M Khapra, and Pratyush Kumar. 2020.
\newblock Plotqa: Reasoning over scientific plots.
\newblock In \emph{Proceedings of the ieee/cvf winter conference on applications of computer vision}, pages 1527--1536.

\bibitem[{Mogull and Stanfield(2015)}]{mogull2015current}
Scott~A Mogull and Candice~T Stanfield. 2015.
\newblock Current use of visuals in scientific communication.
\newblock In \emph{2015 IEEE international professional communication conference (IPCC)}, pages 1--6. IEEE.

\bibitem[{OpenAI(2025)}]{openai_gpt5_2025}
OpenAI. 2025.
\newblock \href {https://openai.com/gpt-5/} {Gpt-5}.
\newblock Large language model.

\bibitem[{Pandey et~al.(2015)Pandey, Rall, Satterthwaite, Nov, and Bertini}]{pandey2015deceptive}
Anshul~Vikram Pandey, Katharina Rall, Margaret~L Satterthwaite, Oded Nov, and Enrico Bertini. 2015.
\newblock How deceptive are deceptive visualizations? an empirical analysis of common distortion techniques.
\newblock In \emph{Proceedings of the 33rd annual acm conference on human factors in computing systems}, pages 1469--1478.

\bibitem[{Park et~al.(2025)Park, Panigrahi, Cheng, Yu, Goyal, and Arora}]{park2025generalizing}
Simon Park, Abhishek Panigrahi, Yun Cheng, Dingli Yu, Anirudh Goyal, and Sanjeev Arora. 2025.
\newblock Generalizing from simple to hard visual reasoning: Can we mitigate modality imbalance in vlms?
\newblock \emph{arXiv preprint arXiv:2501.02669}.

\bibitem[{Parks and Yeh(2021)}]{parks2021lie}
Jonathan Parks and D~Dante Yeh. 2021.
\newblock How to lie with statistics and figures.
\newblock \emph{Surgical infections}, 22(6):611--619.

\bibitem[{Richards(2013)}]{richards2013deceptive}
Jef Richards. 2013.
\newblock \emph{Deceptive advertising: Behavioral study of a legal concept}.
\newblock Routledge.

\bibitem[{Sim et~al.(2025)Sim, Zhang, Dai, and Fang}]{sim2025can}
Mong~Yuan Sim, Wei~Emma Zhang, Xiang Dai, and Biaoyan Fang. 2025.
\newblock Can vlms actually see and read? a survey on modality collapse in vision-language models.
\newblock In \emph{Findings of the Association for Computational Linguistics: ACL 2025}, pages 24452--24470.

\bibitem[{Son and Lee(2025)}]{son2025advancing}
Minjun Son and Sungjin Lee. 2025.
\newblock Advancing multimodal large language models: optimizing prompt engineering strategies for enhanced performance.
\newblock \emph{Applied Sciences}, 15(7):3992.

\bibitem[{Tonglet et~al.(2025{\natexlab{a}})Tonglet, Tuytelaars, Moens, and Gurevych}]{tonglet2025protecting}
Jonathan Tonglet, Tinne Tuytelaars, Marie-Francine Moens, and Iryna Gurevych. 2025{\natexlab{a}}.
\newblock Protecting multimodal large language models against misleading visualizations.
\newblock \emph{arXiv preprint arXiv:2502.20503}.

\bibitem[{Tonglet et~al.(2025{\natexlab{b}})Tonglet, Zimny, Tuytelaars, and Gurevych}]{tonglet2025chart}
Jonathan Tonglet, Jan Zimny, Tinne Tuytelaars, and Iryna Gurevych. 2025{\natexlab{b}}.
\newblock Is this chart lying to me? automating the detection of misleading visualizations.
\newblock \emph{arXiv preprint arXiv:2508.21675}.

\bibitem[{Tufte and Graves-Morris(1983)}]{tufte1983visual}
Edward~R Tufte and Peter~R Graves-Morris. 1983.
\newblock \emph{The visual display of quantitative information}, volume~2.
\newblock Graphics press Cheshire, CT.

\bibitem[{Weber and Rall(2012)}]{weber2012data}
Wibke Weber and Hannes Rall. 2012.
\newblock Data visualization in online journalism and its implications for the production process.
\newblock In \emph{2012 16th International Conference on Information Visualisation}, pages 349--356. IEEE.

\bibitem[{Wu et~al.(2025)Wu, Li, Fu, Kan, and Hooi}]{wu2025seeing}
Jiaying Wu, Fanxiao Li, Zihang Fu, Min-Yen Kan, and Bryan Hooi. 2025.
\newblock Seeing through deception: Uncovering misleading creator intent in multimodal news with vision-language models.
\newblock \emph{arXiv preprint arXiv:2505.15489}.

\end{thebibliography}
